\documentclass[11pt]{article}
\usepackage{PRIMEarxiv}
\usepackage{tabularx}
\usepackage[utf8]{inputenc} %
\usepackage[T1]{fontenc}    %
\usepackage{url}            %
\usepackage{booktabs}       %
\usepackage{amsfonts}       %
\usepackage{nicefrac}       %
\usepackage{microtype}      %
\usepackage{lipsum}
\usepackage{fancyhdr}       %
\usepackage{graphicx}       %
\graphicspath{{media/}}     %
\usepackage{etoolbox}     
\usepackage{makecell} 

\usepackage{soul}
\usepackage{bbm}
\usepackage{url}
\usepackage{booktabs}
\usepackage[utf8]{inputenc}
\usepackage[small]{caption}
\usepackage{amsmath}
\usepackage{amssymb}
\usepackage{amsthm}
\usepackage{wasysym}
\usepackage{algorithm}
\usepackage{algorithmic}
\usepackage[switch]{lineno}
\usepackage{listings}
\usepackage{xparse}
\usepackage{enumitem}
\usepackage{soul}
\usepackage{CJKutf8}
\usepackage{tocloft}
\usepackage{float}
\usepackage{subfig}
\usepackage{soul}
\usepackage{xcolor}
\usepackage{tikz}
\usepackage[edges]{forest}
\definecolor{hidden-draw}{RGB}{20,68,106}
\definecolor{hidden-pink}{RGB}{255,245,247}
\usepackage{bbm}
\newcommand{\etal}{\textit{et~al.}}

\cftsetindents{subsection}{2em}{3em}

\definecolor{mygreen}{RGB}{30, 128, 20}
\colorlet{shadecolor}{gray!20}

\tikzset{chatstyle/.style={text width=2.8in,rounded corners=2pt}}

\definecolor{mygreen}{HTML}{88EABB}
\definecolor{OliveGreen}{HTML}{00693E}
\usepackage{setspace}
\usepackage{wrapfig}
\usepackage{adjustbox}
\usepackage{xspace}

\usepackage[most]{tcolorbox}

\usepackage{fontawesome5}
\usepackage{pifont}

\usepackage{multicol}
\usepackage{multirow}
\usepackage{bbding}
\usepackage{circledsteps}
\usepackage{makecell}

\usepackage{fancyhdr}

\usetikzlibrary{mindmap,shadows}
\usepackage[numbers]{natbib}
\usepackage{hyperref}
\usepackage{cleveref}
\definecolor{LightCyan}{RGB}{232,241,255}
\definecolor{LightRed}{RGB}{255,235,235}
\definecolor{LightPink}{RGB}{255,235,255}
\definecolor{LightGreen}{RGB}{218,255,234}
\definecolor{LightYellow}{RGB}{255,255,235}
\definecolor{LightGray}{RGB}{242,242,242}
\definecolor{Red}{RGB}{253, 239, 242}
\definecolor{Yellow}{RGB}{255, 255, 204}
\definecolor{Pink}{RGB}{255, 243, 254}
\definecolor{Gray}{RGB}{249, 249, 249}
\definecolor{Green}{RGB}{230, 255, 241}
\definecolor{Blue1}{RGB}{218, 232, 245}
\definecolor{Blue2}{RGB}{239, 248, 253}
\definecolor{Blue3}{RGB}{136, 190, 220}
\definecolor{Blue4}{RGB}{83, 157, 204}
\definecolor{Blue5}{RGB}{42, 122, 185}
\definecolor{Blue6}{RGB}{11, 85, 159}
\definecolor{GreenCheck}{RGB}{0, 102, 51}
\definecolor{LightBack}{RGB}{247,249,251}

\ifodd 1

\fi

\NewDocumentCommand{\heng}
{ mO{} }{\textcolor{red}{\textsuperscript{\textit{Heng}}\textsf{\textbf{\small[#1]}}}}

\newcommand{\squishlist}{
\begin{list}{{{\small{$\bullet$}}}}
{\setlength{\itemsep}{1pt}      \setlength{\parsep}{5pt}
\setlength{\topsep}{-2pt}       \setlength{\partopsep}{0pt}
\setlength{\leftmargin}{2.5em} \setlength{\labelwidth}{1em}
\setlength{\labelsep}{1em} } }

\newcommand{\squishend}{  \end{list}  }
\definecolor{aigold}{RGB}{244,210, 1} 
\definecolor{aigreen}{RGB}{210,244,211} 

\sethlcolor{aigreen}
\definecolor{aired}{RGB}{255,180,181} 
\definecolor{lighterseafoam}{RGB}{194,218,184}

\newtcolorbox{boxJ}{
    sharpish corners, 
    colback = sub, 
    colframe = main, 
    boxrule = 0pt, 
    toprule = 4.5pt, 
    enhanced,
    fuzzy shadow = {0pt}{-2pt}{-0.5pt}{0.5pt}{black!35} 
}

\definecolor{CASIABlue}{HTML}{0159E7}

\usepackage{datetime}
\usepackage{CJKutf8}
\usepackage{libertinus}
\usepackage{manyfoot}
\DeclareNewFootnote{A}[fnsymbol]

\title{A Comprehensive Survey on Trustworthiness in Reasoning with Large Language Models}

\begin{document}
\author[1,2]{Yanbo Wang}
\author[1,2]{Yongcan Yu}
\author[1,2,‡]{Jian Liang}
\author[1,2]{Ran He}

\affil[1]{School of Artificial Intelligence, University of Chinese Academy of Sciences}
\affil[2]{NLPR \& MAIS, Institute of Automation, Chinese Academy of Sciences}

\maketitle
\footnotetextA[3]{Corresponding author: Jian Liang \textit{(liangjian92@gmail.com).}}
\footnotetextA[4]{This survey considers papers published up to June 30, 2025. Work in progress.}


\begin{abstract}


\begin{spacing}{1.2}
\textbf{Abstract:} The development of Long-CoT reasoning has advanced LLM performance across various tasks, including language understanding, complex problem solving, and code generation. This paradigm enables models to generate intermediate reasoning steps, thereby improving both accuracy and interpretability. However, despite these advancements, a comprehensive understanding of how CoT-based reasoning affects the trustworthiness of language models remains underdeveloped. In this paper, we survey recent work on reasoning models and CoT techniques, focusing on five core dimensions of trustworthy reasoning: truthfulness, safety, robustness, fairness, and privacy. For each aspect, we provide a clear and structured overview of recent studies in chronological order, along with detailed analyses of their methodologies, findings, and limitations. Future research directions are also appended at the end for reference and discussion. Overall, while reasoning techniques hold promise for enhancing model trustworthiness through hallucination mitigation, harmful content detection, and robustness improvement, cutting-edge reasoning models themselves often suffer from comparable or even greater vulnerabilities in safety, robustness, and privacy. By synthesizing these insights, we hope this work serves as a valuable and timely resource for the AI safety community to stay informed on the latest progress in reasoning trustworthiness. A full list of related papers can be found at \href{https://github.com/ybwang119/Awesome-reasoning-safety}{https://github.com/ybwang119/Awesome-reasoning-safety}.

\end{spacing}


\end{abstract}

\newpage
\newpage
\tableofcontents
\newpage
\begin{CJK*}{UTF8}{gbsn}
\section{Introduction}
With the advancement of large language models (LLMs), Chain-of-Thought (CoT) techniques have become an important way to improve model performance on various downstream tasks, especially in math and code generation. After the release of OpenAI's o1 series models as well as the DeepSeek-R1, developing reasoning models with system-2 thinking also attracted significant interest from researchers around the world, followed by innovations in reinforcement learning algorithms, training data generation, and adaptation methods for other tasks. 

Despite these improvements, the trustworthiness of CoT techniques as well as reasoning models remains underexplored. Intuitively, it may be reasonable that the thinking capability could be generalized to the trustworthiness domain, resulting in a safer and more reliable model. However, recent works~\cite{jiang2025safechain,lu2025does,ying2025towards} did not support such an ideal hypothesis. Furthermore, prior surveys on LLM safety~\cite{wang2025comprehensive,dong2024attacks,shi2024large} provide little discussion of reasoning as a factor in model trustworthiness. This gap motivates the central question:
\textbf{What does the reasoning capability bring to the language model trustworthiness?}

To answer this question, we propose the first comprehensive survey to thoroughly review recent advancements in trustworthy reasoning. We unfold our survey through five main components: truthfulness, safety, robustness, fairness, and privacy. In the truthfulness section, with a focus on model reliability, we include hallucination and reasoning faithfulness, encompassing hallucination detection and mitigation methods with CoT techniques, hallucination analysis in reasoning models, reasoning faithfulness measurement, faithfulness understanding, as well as methods to improve reasoning faithfulness. In the safety section, we aim to understand the harmlessness of the generation content, and mainly take vulnerability assessment, jailbreak, alignment, and backdoor into consideration. For better readability, we specifically distinguish between jailbreak attacks targeting reasoning models and the use of reasoning techniques in attack and defense, forming different paragraphs to structure the literature. In the robustness section, we mainly focus on adversarial input noises that elicit false answers at inference time. The overthinking and underthinking problems are highlighted as a special case when language models are equipped with reasoning capability. After that, in the fairness section, we mainly cover the latest evaluations and methods for bias detection. As for the privacy section, we split the related works into model-related privacy and prompt-related privacy, with topics containing model unlearning, IP protection, watermarking, and privacy inference.

While existing surveys have explored reasoning techniques~\cite{chen2025towards,xu2025towards} and reasoning efficiency~\cite{qu2025survey,sui2025stop,feng2025efficient}, relatively little attention has been paid to the trustworthiness of reasoning in large language models. A related survey~\cite{wang2025safety} provided valuable discussions on safety-related aspects. In contrast, our work offers a more comprehensive perspective on trustworthiness.
In general, we provide a clear taxonomy for model trustworthiness in reasoning, which includes both early CoT techniques and end-to-end reasoning models. Through our review of existing work, we suggest that reasoning techniques not only facilitate the development of more interpretable and trustworthy models but also introduce new vulnerabilities. As models acquire more advanced reasoning capabilities, the attack surface correspondingly expands, enabling more complex and targeted adversarial strategies. We hope that both the surveyed literature and our proposed taxonomy will serve as a timely reference for the AI safety community, supporting ongoing efforts to understand and improve the trustworthiness of reasoning in language models.

\begin{table*}[ht]
\centering
\caption{List of Abbreviations and Acronyms}
\resizebox{.95\textwidth}{!}{
\begin{tabular}{cc|cc}
\hline
\textbf{Abbreviation} & \textbf{Full Term} & \textbf{Abbreviation} & \textbf{Full Term} \\
\hline
AOC & Area Over Curve & MCTS & Monte-Carlo Tree Search \\
ASR & Attack Success Rate & MLLM & Multimodal Large Language Model \\
CNN & Convolutional Neural Network & MLRM & Multimodal Large Reasoning Model \\
CoT & Chain-of-Thought & ORM & Outcome Reward Model \\
DFS & Depth-First Search & PRM & Process Reward Model \\
DPO & Direct Preference Optimization & QA & Question-Answering \\
GRPO & Group Relative Policy Optimization & RL & Reinforcement Learning \\
ICL & In-Context Learning & RLHF & Reinforcement Learning from Human Feedback \\
KL & Kullback-Leibler Divergence & RLVR & Reinforcement Learning with Verifiable Reward \\
LAS & Leakage-Adjusted Simulatability & RAG & Retrieval-Augmented Generation \\
LLM & Large Language Model & SCM & Structural Causal Model \\
LRM & Large Reasoning Model & SoTA & State-of-the-Art \\
LoRA & Low-Rank Adapter & SFT & Supervised Fine-Tuning \\
MoE & Mixture-of-Experts & VR & Verifiable Reward \\
\hline
\end{tabular}
}
\end{table*}

\section{Background}
In this section, we provide an overview of fundamental concepts related to reasoning in language models, including discussions of the general definition of reasoning, an introduction to CoT as a widely adopted technique, and key considerations in model training that influence the reasoning abilities.
\subsection{Large Language Model Reasoning}
LLM reasoning is a novel paradigm that leverages the knowledge embedded within models like GPT-4~\cite{achiam2023gpt}, Claude~\cite{TheC3}, and DeepSeek-R1~\cite{guo2025deepseek} to solve complex tasks—such as math, coding, and logical reasoning—by mimicking human cognitive processes.
Typically, LLM reasoning involves generating both the final answer and the intermediate steps, often referred to as “thoughts”, which guide the model from the question to the answer.
Formally, given a prompt $x$ and context $C$, the reasoning of an LLM $\mathcal{M}$ can be represented as follows:
\begin{equation}
    T, A = \mathcal{M}(x,C),
\end{equation}
where $T$ refers to the intermediate reasoning process and $A$ is the answer.
By enabling the AI system to generate interpretable reasoning steps alongside the solution, LLM reasoning not only solves complex tasks but also improves human understanding of the problem-solving process, thereby enhancing its utility and reliability.
Currently, the two main paradigms for implementing large language model reasoning are CoT prompting and large reasoning model training.

\subsection{Chain-of-Thought Prompting}
\begin{figure*}[h]
    \centering
    \includegraphics[width=\linewidth]{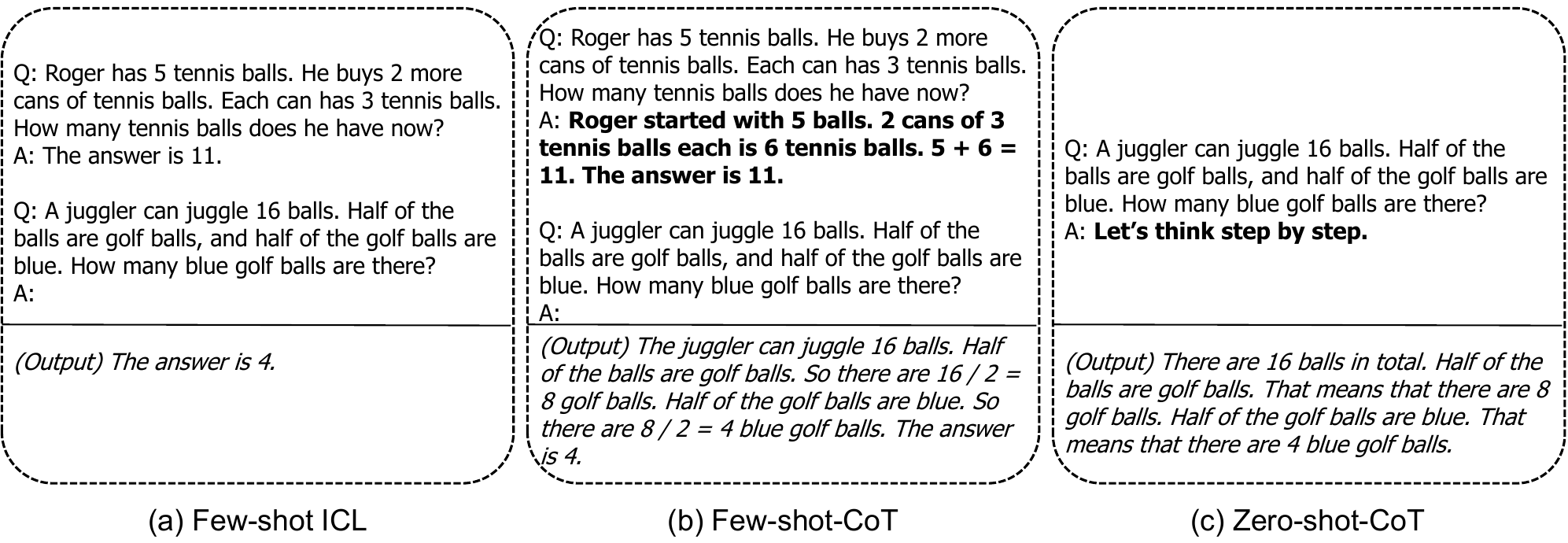}
    \centering
    \caption{Illustration of typical CoT prompting. Few-shot-CoT uses several examples with the reasoning process to elicit CoT, and zero-shot-CoT uses a prefix prompt to induce the reasoning process.}
    \label{fig:cot}
\end{figure*}

CoT prompting~\cite{wei2022chain,kojima2022large} is a prompt engineering technique designed to elicit a sequence of intermediate reasoning steps referred to as the thought, before providing the final answer.
There are various methods for implementing CoT, with two of the most common being few-shot-CoT~\cite{wei2022chain} and zero-shot-CoT~\cite{kojima2022large}.
As illustrated in \Cref{fig:cot}, few-shot-CoT mirrors the approach of few-shot in-context learning (ICL)~\cite{brown2020language}, utilizing a small number of examples to guide the model in answering questions.
Unlike traditional ICL, few-shot-CoT~\cite{li2024chain} not only shows the answer in the demonstrations, but also gives the specific reasoning steps before the answer. Therefore, the model will also give CoT before answering the question.
While few-shot-CoT demonstrates strong performance on complex tasks such as math and symbolic reasoning, it requires human-annotated, task-specific examples with intricate reasoning paths, limiting its applicability.
In contrast, zero-shot-CoT~\cite{wei2022chain} offers a more flexible, task-agnostic method for eliciting CoT by simply adding the prefix ``Let's think step by step'' before generating the answer.
\subsection{Large Reasoning Models}
Large reasoning models (LRMs), represented by OpenAI o1~\cite{jaech2024openai} and DeepSeek-R1~\cite{guo2025deepseek}, refer to a series of large language models that explicitly generate their thinking process before filling the final answers~\cite{xu2025towards}. Instead of prompting models to ``think step by step'', reasoning models could automatically create the thinking process that mimics how humans analyze a problem. 
\subsubsection{Model Training} \label{model_training}
There are a few open-source trials to replicate the o1 series~\cite{xu2025towards}, including OpenR~\cite{wang2024openr}, o1-journey~\cite{qin2024o1,huang2024o1,huang2025o1}, and LLaMA-Berry~\cite{zhang2024llama}. The key to the replication lies in distilling long CoT data, even if the source model has not been explicitly trained for reasoning. LLaMA-Berry~\cite{zhang2024llama} utilized Monte Carlo tree search (MCTS)~\cite{browne2012survey} with a pairwise preference reward model to scale test-time compute, achieving a higher performance on multiple Math datasets such as GSM8k~\cite{cobbe2021gsm8k}, MATH~\cite{hendrycks2021measuring}, GaoKao2023En~\cite{liao2024mario}, etc. O1-journey~\cite{qin2024o1} utilized MCTS with a fine-grained reward model to construct long CoT data. After building the reasoning tree with each node annotated with a reward score indicating correctness, a traversal algorithm such as Depth-First Search (DFS) with constraints could be adopted to create a datapoint using an error-then-backtrack style. Supervised fine-tuning (SFT), followed by Direct Preference Optimization (DPO)~\cite{rafailov2023direct}, was then leveraged to train the reasoning model. OpenR~\cite{wang2024openr} introduced reinforcement learning with a process reward model to encourage reasoning capability. During training, the LLM policy was updated at each reasoning step using intermediate step-wise rewards from the reward model, optimized with either the proximal policy optimization (PPO)~\cite{schulman2017proximal} or the group relative policy optimization (GRPO)~\cite{shao2024deepseekmath}. Except for these tree searching methods, DeepSeek-R1 demonstrated the outstanding performance of pure reinforcement learning in boosting reasoning capability, utilizing distilled data from R1-Zero\footnote{The model for long CoT data synthesis underwent preliminary supervised fine-tuning (cold start). Therefore, it is slightly different from the released R1-Zero model.} to train the base model.
One point worth noting is that, except for latent reasoning models~\cite{geiping2025scaling,hao2024training}, there is no obvious difference between previous chat models and current reasoning models in terms of model structure. In fact, all these models are developed based on well-trained chat models such as DeepSeek-V3~\cite{liu2024deepseek}, Qwen2.5~\cite{qwen2024qwen2}, and Llama-3 series~\cite{dubey2024llama}. 

\textbf{PRM, ORM, and VR.} According to Uesato \etal~\cite{uesato2022solving}, current reward models could be divided into two types: process reward model (PRM) and outcome reward model (ORM), in which the former provides stepwise reward on each reasoning process, and the latter simply gives one score for the whole generation sequence. Instead of ORM~\cite{zelikman2024star}, Lightman \etal~\cite{lightman2023let} proposed PRM to verify the thinking process step by step, and demonstrated its superior performance to ORM in providing more reliable step-wise reward. For inference-time scaling, these reward models could not only facilitate the tree search at inference time for better performance, but also help filter reasoning trajectories with higher quality for post-training. Before the release of DeepSeek-R1~\cite{guo2025deepseek}, the training of reward models is crucial for reasoning model development. Verifiable reward (VR) was first proposed by Lambert \etal~\cite{lambert2024t}, which includes three types: correctness verification, verification via execution, and verifiable constraints~\cite{mroueh2025reinforcement}. Different from reward models, here we define verifiable reward as ``\textit{the reward provided by a simple deterministic function instead of large models, which is objective, usually binary, and outcome-based}''. DeepSeek-R1 demonstrates the effectiveness of VR, which is then regarded as a prevailing post-training method when combined with GRPO.

\subsubsection{Multimodal LRM}
Li \etal~\cite{li2025perception} summarized the development of multimodal large reasoning models (MLRMs) into three stages: ``perception driven modular reasoning'', ``language-centric short reasoning'', and ``language-centric long reasoning''. Like the development of unimodal large reasoning models, MLRMs also experienced the transformation from zero-shot or few-shot CoT prompting to long reasoning data post-training~\cite{wang2025multimodal}. For example, Multimodal-CoT~\cite{zhang2023multimodal}, VoT~\cite{fei2024video}, and VIC~\cite{zheng2024thinking} are some of the early works that focused on the prompting to elicit model thinking. In terms of training, LLaVA-CoT~\cite{xu2024llava}, Llamav-o1~\cite{thawakar2025llamav}, RedStar~\cite{xu2025redstar}, and Mulberry~\cite{yao2024mulberry} propose to empower multimodal large language models (MLLMs) with reasoning capabilities by finetuning base models. As stated in \Cref{model_training}, multimodal CoT data generation is also crucial for model training, and the construction of the reasoning path includes distillation~\cite{xu2024llava,thawakar2025llamav,zhang2024improve,dong2025insight,guo2024mammoth} or MCTS~\cite{yao2024mulberry,sun2025mm}, which also resembles the way mentioned for text-domain CoT data generation.  

As for model training, pure GRPO and SFT followed by GRPO become the prevailing method for reasoning model development~\cite{wang2025multimodal}, which may be attributed to the outstanding performance of RL demonstrated by DeepSeek-R1. 
\clearpage
\definecolor{hidden-pink}{RGB}{255,192,203}
\tikzstyle{my-box1}=[
    rectangle,
    draw=hidden-draw,
    rounded corners,
    text opacity=1,
    minimum height=1.5em,
    minimum width=5em,
    inner sep=2pt,
    align=center,
    fill opacity=.5,
    line width=0.8pt,
]
\tikzstyle{my-box2}=[
    rectangle,
    draw=hidden-draw,
    rounded corners,
    text opacity=1,
    minimum height=1.5em,
    minimum width=5em,
    inner sep=2pt,
    align=center,
    fill opacity=.5,
    line width=0.8pt,
]

\tikzstyle{leaf}=[my-box1, minimum height=1.5em,
    fill=hidden-pink!80, text=black, align=left,font=\footnotesize,
    inner xsep=2pt,
    inner ysep=4pt,
    line width=0.8pt,
]

\begin{figure*}[!th]
    \centering
    \resizebox{0.96\textwidth}{!}{
        \begin{forest}
            forked edges,
            for tree={
                grow=east,
                reversed=true,
                anchor=base west,
                parent anchor=east,
                child anchor=west,
                base=center,
                font=\large,
                rectangle,
                draw=hidden-draw,
                rounded corners,
                align=left,
                minimum width=4em,
                edge+={darkgray, line width=1pt},
                s sep=3pt,
                inner xsep=2pt,
                inner ysep=3pt,
                line width=0.8pt,
                ver/.style={rotate=90, child anchor=north, parent anchor=south, anchor=center},
            },
            where level=1{text
            width=6em,align=center,font=\normalsize,}{},
            where level=2{text width=8.2em,align=center,font=\footnotesize,}{},
            where level=3{text width=8.2em,align=center,font=\footnotesize,}{},
            where level=4{text width=8.2em, font=\footnotesize,}{},
            [
                Trustworthiness in language model reasoning, ver
                [
                    Truthfulness \\ (\S \ref{truthfulness}), my-box1
                    [
                        Hallucination\\ (\S \ref{hallucination}), my-box1
                        [
                            Hallucination with\\reasoning techniques, my-box2
                            [Multimodal-CoT~\cite{zhang2023multimodal}{, } HaluSearch~\cite{cheng2025think}{, } HalluMeasure~\cite{akbar2024hallumeasure}{, } \\CLATTER~\cite{eliav2025clatter}{, }  Reflexive Prompting~\cite{xie2024order}{, } GCoT~\cite{wu2025grounded}{, }CoMT~\cite{jiang2024comt}, leaf, text width=26em
                            ]
                        ]
                        [
                        Hallucination of\\reasoning models, my-box2
                            [
                            MIRAGE~\cite{dong2025mirage}{, } SUM~\cite{song2025hallucination}{, }RH-Bench~\cite{liu2025more}{, } Yao \etal~\cite{yao2025reasoning}{, }\\
                            VIC~\cite{zheng2024thinking}{, }AbstentionBench~\cite{kirichenko2025abstentionbench}{, } 
                            Lu \etal~\cite{lu2025auditing}{, } 
                            FSPO~\cite{li2025hallucination}{, } \\
                            Anh \etal~\cite{anh2025analyzing}{, } 
                            GRPO-R~\cite{sun2025detection}{, }
                            RFMDataset~\cite{guo2025mathematical}{, } 
                            FG-PRM~\cite{li2024fine}{, } \\
                            Zhang \etal~\cite{zhang2025reasoning}{, } 
                            RACE~\cite{wang2025joint}
                            , leaf, text width=26em 
                            ]
                        ]
                    ]
                    [
                       Faithfulness of\\ Reasoning Models\\ (\S \ref{faithfulness}), my-box1  
                        [
                            Measuring \&\\ understanding, my-box1
                            [
                            Lanham \etal~\cite{lanham2023measuring}{, } Turpin \etal~\cite{turpin2023language}{, } PFF~\cite{tutek2025measuring}{, } Xiong \etal~\cite{xiong2025measuring}{, }\\
                            Bentham \etal~\cite{benthamchain}{, }
                            Arcuschin \etal~\cite{arcuschin2025chain}{, } Chua \etal~\cite{chua2025deepseek}{, }\\
                            Chen \etal~\cite{chen2025reasoning}{, }
                            Li \etal~\cite{li2024towards}{, }
                            Agarwal \etal~\cite{agarwal2024faithfulness}{, }
                            Bao \etal~\cite{bao2024likely}{, }\\
                            Tanneru \etal~\cite{tanneru2024difficulty}{, }
                            Lobo \etal~\cite{lobo2024impact}
                            , leaf, text width=26em
                            ]
                        ]
                        [
                                Faithfulness\\improvement, my-box1
                            [     FRODO~\cite{paul2024making}{, } SymbCoT~\cite{xu2024faithful}{, } Radhakrishnan \etal~\cite{radhakrishnan2023question}{, }\\ Faithful CoT~\cite{lyu2023faithful}{, }
                            LOGIC-LM~\cite{pan2023logic}{, }
                            FLARE~\cite{arakelyan2024flare}{, } CoMAT~\cite{leang2024comat}{, } \\
                            CORE~\cite{wang2024causal}{, } 
                            QUIRE~\cite{li2024towards}{, }
                            Fact~\cite{gao2024fact}{, }
                            Viteri \etal~\cite{viteri2024markovian}, leaf, text width=26em
                            ]
                        ]
                    ]
                ]
                [                
                    Safety \\ (\S \ref{safety}), my-box1
                    [
                    Vulnerability\\Assessment\\ (\S \ref{Vulnerability Assessment}), my-box1
                    [
                    SafeChain~\cite{jiang2025safechain}{, } CNSafe~\cite{ying2025towards}{, } Zhang \etal~\cite{zhang2025safety}{, }
                    Romero \etal~\cite{romero2025red}{, } Zhou \etal~\cite{zhou2025hidden}{, }\\ Li \etal~\cite{li2025smarter}{, } Lou \etal~\cite{lou2025think}{, }
                    kassianik \etal~\cite{kassianik2025evaluating}{, } 
                    Krishna \etal~\cite{krishna2025weakest}{, } FORTRESS~\cite{knight2025fortress}{, }\\
                    Fan \etal~\cite{fan2025evaluation}{, } BSAbench~\cite{zheng2025beyond}{, } Is-bench~\cite{lu2025bench}{, }
                    SafeMLRM~\cite{fang2025safemlrm}{, } Zhao \etal~\cite{zhao2025trade}{, }\\
                    Marjanovi{\'c} \etal~\cite{marjanovic2025deepseek}
                    , leaf, text width=36em,align=left
                    ]
                    ]
                    [
                    Jailbreak\\ (\S \ref{jailbreak}), my-box1                   
                    [
                    Attack with \\reasoning techniques, my-box1
                    [
                    Sabbaghi \etal~\cite{sabbaghi2025adversarial}{, }
                    CoT-GCG~\cite{su2024enhancing}{, }
                    Ying \etal~\cite{ying2025reasoning}{, }\\
                    Chain-of-Lure~\cite{chang2025chain}{, }
                    Handa \etal~\cite{handa2024competency}
                    , leaf, text width=26em
                    ]
                    ]
                    [
                    Attack on\\reasoning models, my-box1
                    [
                    H-CoT~\cite{kuo2025h}{, } Mousetrap~\cite{yao2025mousetrap}{, } AutoRAN~\cite{liang2025autoran}{, }
                    SEAL~\cite{nguyen2025three}{, }\\
                    FicDetail~\cite{lu2025does}{, }
                    Lian \etal~\cite{lian2025revealing}{, }
                    RRTL~\cite{liu2025rrtl}{, } VisCRA~\cite{sima2025viscra}{, }\\ HauntAttack~\cite{ma2025hauntattack}
                    , leaf, text width=26em
                    ]
                    ]
                    [
                    Defense with \\reasoning techniques, my-box1
                    [
                    GuardReasoner~\cite{liu2025guardreasoner}{, }
                    X-Guard~\cite{upadhayay2025x}{, }
                    MrGuard~\cite{yang2025mr} {, }\\
                    RSafe~\cite{zheng2025rsafe}{, }
                    Sreedhar \etal~\cite{sreedhar2025safety}{, }
                    R$^2$-Guard~\cite{kang2024r}{, } DR-IRL~\cite{cheng2025inverse}{, }\\
                    ShieldVLM~\cite{cui2025shieldvlm}{, }
                    GuardReasoner-VL~\cite{liu2025guardreasonervl}{, }
                    GuardAgent~\cite{xiang2024guardagent}{, }\\
                    ShieldAgent~\cite{chen2025shieldagent}{, }
                    Wang \etal~\cite{wang2025unified}{, } U-CoT+~\cite{pan2025detecting}
                    , leaf, text width=26em
                    ]
                    ]
                    [
                    Defense for \\reasoning models, my-box1
                    [
                    SafeChain~\cite{jiang2025safechain}{, }
                    Thinking Intervention~\cite{wu2025effectively}{, }
                    Wang \etal~\cite{wang2025safety}{, }\\
                    Yamaguchi \etal~\cite{yamaguchi2025adversarial}{, }
                    Zaremba \etal~\cite{zaremba2025trading}{, }
                    Saffron-1~\cite{qiu2025saffron}
                    , leaf, text width=26em
                    ]
                    ]
                    ]
                    [
                    Alignment\\ (\S \ref{alignment}), my-box1
                    [
                    Aligning LLM using\\reasoning techniques, my-box1
                    [
                    Liu \etal~\cite{liu2024mixture}{, } Zhang \etal~\cite{zhang2024backtracking}{, } SCoT~\cite{yang2025enhancing}{, }
                    STAIR~\cite{yang2025enhancing}{, }\\
                    R2D~\cite{zhu2025reasoning}{, } RATIONAL~\cite{zhang2025safety1}{, }
                    ERPO~\cite{feng2025erpo}{, } 
                    SaRO~\cite{mou2025saro}{, }\\
                    Wang \etal~\cite{wang2025safety}{, }
                    Kim \etal~\cite{kim2025reasoning}{, }
                    Thought-Aligner~\cite{jiang2025think}{, }\\
                    ReasoningShield~\cite{li2025reasoning}
                    , leaf, text width=26em
                    ]
                    ]
                    [
                    Alignment of\\reasoning models, my-box1
                    [
                    Deliberate Alignment~\cite{guan2024deliberative}{, } SafeChain~\cite{jiang2025safechain}{, } STAR-1~\cite{wang2025star}{, }\\
                    RealSafe~\cite{zhang2025realsafe}{, }
                    SAFEPATH~\cite{jeung2025safepath}{, }
                    Context Reasoner~\cite{hu2025context}{, }\\ Lou \etal~\cite{lou2025think}{, }
                    Baker \etal~\cite{baker2025monitoring}{, } Zhang \etal~\cite{zhang2025should}{, } Hair~\cite{cheng2025hair}{, }\\
                    Liu \etal~\cite{liu2025chasing}{, } Safety Tax~\cite{huang2025safety}{, } SafeKey~\cite{zhou2025safekey}
                    , leaf, text width=26em
                    ]
                    ]
                    ]
                    [
                    Backdoor\\ (\S \ref{backdoor}), my-box1
                    [
                    Training-time\\data poisoning, my-box1
                    [
                    SABER~\cite{jin2024saber}{, } BoT~\cite{zhu2025think}{, } ShadowCoT~\cite{zhao2025shadowcot}{, }Chua \etal~\cite{chua2025thought}
                    , leaf, text width=26em
                    ]
                    ]
                    [
                    Inference-time\\prompt manipulation, my-box1
                    [
                    Badchain~\cite{xiang2024badchain}{, } BackdoorLLM~\cite{li2024backdoorllm}{, } DarkMind~\cite{guo2025darkmind}{, }CPT~\cite{cui2025process}{, }\\
                    Guo \etal~\cite{guo2025system}{, } Cui \etal~\cite{cui2025practical}{, } Cui \etal~\cite{cui2025token} Song \etal~\cite{song2025chain}
                    , leaf, text width=26em
                    ]
                    ]
                    [
                    Backdoor\\defense, my-box1
                    [
                    Chain-of-Scrutiny~\cite{li2024chain}{, } Marinelli \etal~\cite{marinelli2025harnessing}{, } GUARD~\cite{jin2025guard}
                    , leaf, text width=26em
                    ]
                    ]
                    ]
                ]                
                [
                    Robustness \\ (\S \ref{robustness}), my-box1
                    [
                        Improvement with\\reasoning techniques\\ (\S \ref{Robustness Improvement}), my-box1
                        [ 
                        Wang \etal~\cite{wang2025assessing}{, } CoDT~\cite{wang2025chain}{, }
                        Yan \etal~\cite{yan2025recitation}{, } RBD~\cite{yang2025any}{, }Zaremba \etal~\cite{zaremba2025trading}
                        , leaf, text width=36em, align=left
                        ]
                    ]
                    [  
                        Robustness of\\reasoning models\\ (\S \ref{Robustness of Reasoning Models}), my-box1
                        [
                        RUPbench~\cite{wang2024rupbench}{, }Mu \etal~\cite{mu2025closer}{, } 
                        RoR-bench~\cite{yan2025recitation}{, }M-Attack~\cite{li2025frustratingly}{, }GaslightingBench-R~\cite{zhu2025reasoning1}{, }\\ 
                        Zhou \etal~\cite{zhou2024can}{, }  
                        Peng \etal~\cite{peng2025stepwise}{, } PolyMath~\cite{wang2025polymath}{, } 
                        CatAttack~\cite{rajeev2025cats}{, } Math-RoB~\cite{yu2025benchmarking}{, }\\ MATH-Perturb~\cite{huang2025math}{, } 
                        CodeCrash~\cite{lam2025codecrash}{, } CoCC~\cite{roh2025break}{, }AbstentionBench~\cite{kirichenko2025abstentionbench}{, }Xu \etal~\cite{xu2024preemptive}
                        , leaf, text width=36em, align=left
                        ]
                    ]   
                    [
                    Overthinking\\ (\S \ref{overthink}), my-box1
                    [
                    UMP~\cite{ma2024large}{, } DNR Bench~\cite{hashemi2025dnr}{, }
                    DeltaBench~\cite{he2025can}{, }
                    MiP-Overthinking~\cite{fan2025missing}{, } Si \etal~\cite{si2025excessive}{, }\\ Wang \etal~\cite{wang2025thoughts}{, } Su \etal~\cite{su2025between}{, } Dang \etal~\cite{dang2025internal}{, }
                    Overthink~\cite{kumar2025overthink}{, } Cuadron \etal~\cite{cuadron2025danger}
                    , leaf, text width=36em,align=left
                    ]
                    ]
                    [
                    Underthinking\\ (\S \ref{overthink}), my-box1
                    [
                    CPT~\cite{cui2025process}{, } 
                    Zaremba \etal~\cite{zaremba2025trading}{, } Zhao \etal~\cite{zhao2025trade}{, } Li \etal~\cite{li2025output}{, } ThinkEdit~\cite{sun2025thinkedit} 
                    , leaf, text width=36em,align=left
                    ]
                    ]
                ]
                [                
                Fairness \\ (\S \ref{fairness}), my-box1
                [
                Evaluation \&\\ Detection, my-box1
                [
                Lin \etal~\cite{linassessing}{, } Cheng \etal~\cite{cheng2025detection}{, } Kamruzzaman~\etal\cite{kamruzzaman2024prompting}{, }
                Dash~\etal\cite{dash2025persona}{, }\\ Gupta~\etal\cite{gupta2023bias}{, }
                BiasGuard~\cite{fan2025biasguard}{, } Cantini \etal~\cite{cantini2025reasoning}
                , leaf, text width=36em,align=left
                ]
                ]
                ]
                [                
                Privacy \\ (\S \ref{privacy}), my-box1
                [
                Model-related\\ Privacy (\S \ref{Model-related privacy}), my-box1
                [
                R-TOFU~\cite{yoon2025r}{, } R$^2$MU~\cite{wang2025reasoning}{, } SLEEK~\cite{sinha2025step}{, }
                ImF~\cite{wanli2025imf}{, }
                CoTSRF~\cite{ren2025cotsrf}{, }
                Guo \etal~\cite{guo2025towards}{, }\\
                Savani \etal~\cite{savani2025antidistillation}
                , leaf, text width=36em,align=left
                ]
                ]
                [
                Prompt-related\\ Privacy (\S \ref{Prompt-related Privacy}), my-box1 
                [
                Green \etal~\cite{green2025leaky}{, } DoxBench~\cite{luo2025doxing}
                , leaf, text width=36em,align=left
                ]
                ]
                ]
            ]
        \end{forest}
    }
    \caption{Taxonomy of trustworthiness in reasoning with large language models.}
    \label{taxo_of_bench}
\end{figure*}
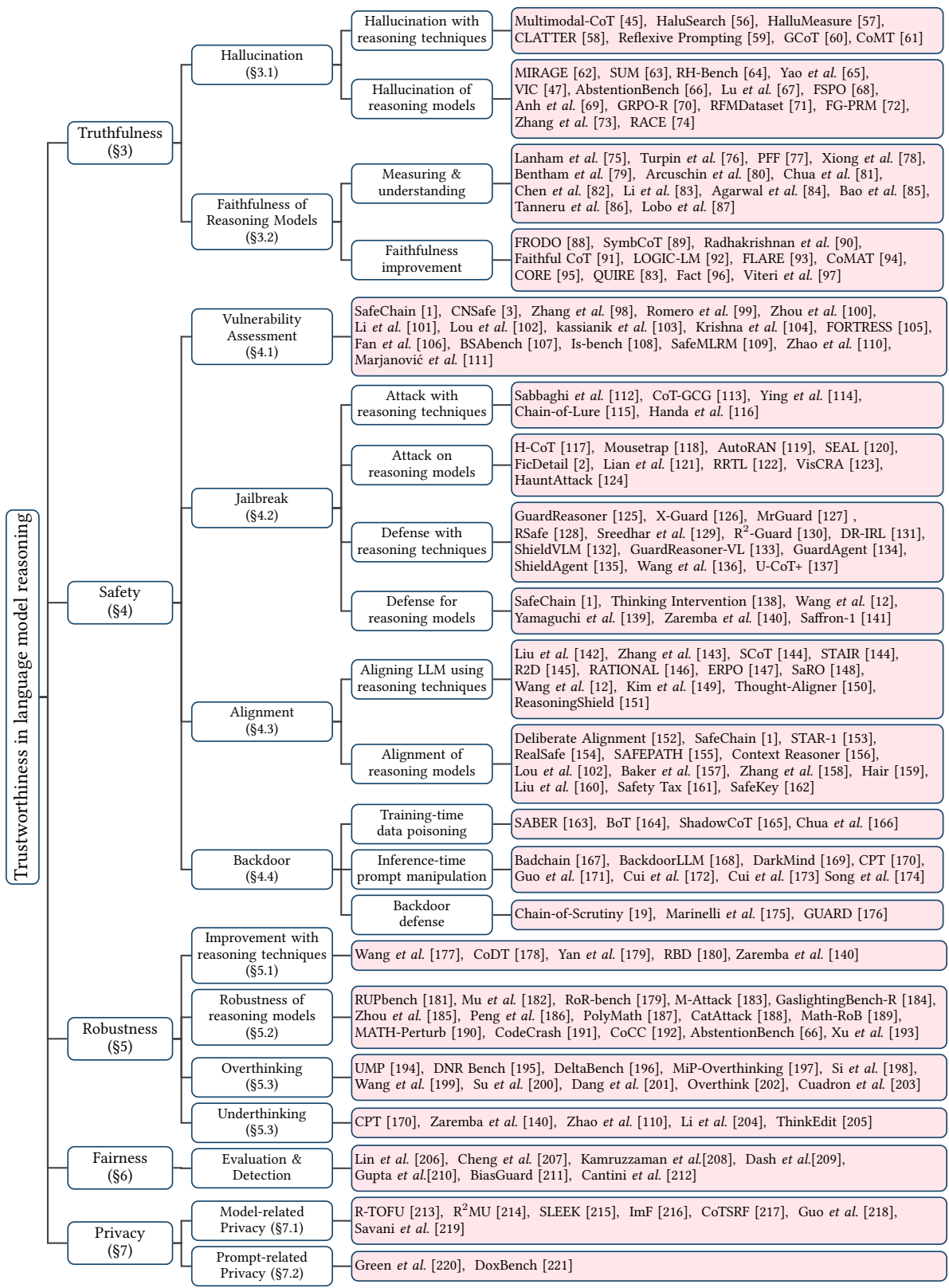

\newpage
\section{Truthfulness}
\label{truthfulness}

Truthfulness in the LLMs refers to how an AI system accurately represents information, facts, and results~\cite{huang2024trustllm}. This fundamental dimension of truthfulness focuses on the model's ability to provide factually correct and reliable information without generating misleading or false content.
In this section, we discuss the new challenges brought by the reasoning techniques, including two aspects: hallucination and faithfulness.

\subsection{Hallucination}
\label{hallucination}
Hallucination in LLMs refers to instances where \textit{models generate responses that appear coherent and plausible but are inconsistent with the input, context, or factual information}~\cite{huang2025survey,rawte2023survey}.
The emergence of reasoning models introduces new risks and challenges in managing hallucinations.
First, reasoning models often generate responses that are more structured, logically coherent, and superficially persuasive, making them appear more reliable.
As a result, hallucinated content from these models can appear more credible, making it harder for users to detect inaccuracies and increasing the risk of spreading misinformation~\cite{sun2025detection}, especially in high-stakes fields such as healthcare, law, or education.
On the other hand, the CoT reasoning generated by models can also contain hallucinations~\cite{li2025hallucination}. Compared to traditional LLMs, the hallucinations in reasoning models have not been as thoroughly evaluated.
Moreover, the powerful reasoning capabilities of these models can be leveraged to detect or mitigate hallucinations in certain complex tasks~\cite{cheng2025think,eliav2025clatter}.

\subsubsection{Hallucination with Reasoning Techniques}
In this section, we explore how reasoning techniques can be leveraged to detect and mitigate hallucinations in LLMs.
CoT prompting has shown remarkable success in addressing complex tasks~\cite{kojima2022large,wei2022chain} and reducing hallucinations~\cite{cheng2025chain}. To further enhance model reasoning capabilities, several techniques have been proposed, such as test-time scaling~\cite{snell2024scaling}, self-consistency~\cite{wang2022self}, etc.
One such approach, HaluSearch~\cite{cheng2025think}, employed a tree search-based algorithm coupled with a switch model to determine when to engage in more deliberate, ``slow thinking'' processes.
In contrast to hallucination mitigation, HalluMeasure~\cite{akbar2024hallumeasure} focused on fine-grained hallucination measurement, using CoT prompting. Specifically, it decomposed model responses into a series of claims and applies CoT techniques to detect hallucinations at the claim level.
Similarly, CLATTER~\cite{eliav2025clatter} adopted a multi-step reasoning process for hallucination detection, consisting of decomposition, attribution, entailment, and aggregation.
Moreover, Xie \etal~\cite{xie2024order} observed that the order in which reasoning steps are applied can influence hallucination occurrence. As such, they propose Reflexive Prompting, which combines ``answer-first'' and ``logic-first'' reasoning strategies to improve model accuracy.
Beyond text-based tasks, Zhang \etal~\cite{zhang2023multimodal} extended CoT to multimodal settings, proposing a method to mitigate visual hallucinations.
Their approach involves generating a rationale that is used to update the language input, which is then combined with the original visual input to produce the final answer.
Furthermore, Wu \etal~\cite{wu2025grounded} introduced Grounded Chain-of-Thought (GCoT), a technique in which the model gradually grounds visual cues before generating answers. This step-by-step process helps mitigate visual hallucinations by enhancing the model's understanding of the input.
In addition, in the context of medical report generation, CoMT~\cite{jiang2024comt} leveraged CoT prompting to reduce hallucinations and produce high-quality, accurate reports.
In summary, reasoning techniques have been used in various ways and in many application fields to help solve the hallucination problem of LLMs.

\subsubsection{Hallucination in Reasoning Models}
Despite their ability to tackle complex tasks, reasoning models are not immune to hallucination. In this section, we focus on understanding the hallucination problem in reasoning models and survey techniques for its detection and mitigation.

\textbf{Hallucination analysis}.
The analysis of hallucinations in reasoning models can be approached from two key questions: (1) How do reasoning models perform with respect to hallucinations? and (2) What factors contribute to hallucinations in reasoning models?

Several studies~\cite{dong2025mirage, song2025hallucination, liu2025more, yao2025reasoning, cheng2025chain, kirichenko2025abstentionbench} have documented significant hallucination issues within reasoning models, sometimes more pronounced than in non-reasoning models.
For instance, Lu \etal~\cite{lu2025auditing} argued that LRMs exacerbate hallucination issues, making them more frequent and harder to mitigate. Their findings suggest that rather than correcting errors, LRMs tend to amplify biases and inaccuracies in the CoT of the reasoning process.
Similarly, Song \etal~\cite{song2025hallucination} and Kirichenko \etal~\cite{kirichenko2025abstentionbench} highlighted that reasoning models, when faced with unanswerable questions, struggle to recognize and refuse to respond appropriately, a challenge that is less prevalent in non-reasoning models.
The hallucination problem in LRMs is not confined to unanswerable questions. Li \etal~\cite{li2025hallucination} and Yao \etal~\cite{yao2025reasoning} evaluated reasoning models on both traditional hallucination benchmarks (e.g., TruthfulQA~\cite{lin2021truthfulqa}, HaluEval~\cite{li2023halueval}, HalluQA~\cite{cheng2023evaluating}) and fact-seeking benchmarks (e.g., SimpleQA~\cite{wei2024measuring}, TriviaQA~\cite{joshi2017triviaqa}), consistently finding that reasoning models exhibit higher rates of hallucination.
Liu \etal~\cite{liu2025more} extended this observation to visual tasks, where improved reasoning capabilities were often accompanied by more severe visual hallucinations. Together, these studies suggest that \textbf{while reasoning models improve performance on complex tasks, they can also produce more significant hallucinations than non-reasoning models in simpler, non-reasoning tasks.}
Moreover, many studies have also found that there are serious illusions in the generated CoT~\cite{anh2025analyzing,lu2025auditing,sun2025detection,li2025hallucination,guo2025mathematical}. Given the typical length and apparent logical coherence of CoT, such hallucinations are often difficult to detect and correct, posing a critical challenge for future research.

When examining the causes of hallucinations, several studies point to the length of the CoT as a significant factor~\cite{lu2025auditing,liu2025more}.
For example, Lu \etal~\cite{lu2025auditing} reported that hallucinations tend to occur more frequently in longer CoTs compared to those with correct answers.
Similarly, Liu \etal~\cite{liu2025more} observed that as CoTs become longer, models increasingly rely on language priors over visual inputs, a shift that often leads to visual hallucinations.
Another important factor is the training paradigm of the model.
Yao \etal~\cite{yao2025reasoning} suggested that while combining SFT with RL training can improve model performance on fact-seeking tasks, both SFT-only and RL-only paradigms lead to severe hallucinations, often manifesting as flaw repetition or mismatched thinking and answers.
Li \etal~\cite{li2025hallucination} similarly identified outcome-based RL fine-tuning as a contributor to hallucinations, highlighting three critical factors: high variance in policy gradients, high entropy in predictions, and the presence of spurious local optima.

\textbf{Hallucination detection and measurement}.
The PRM model~\cite{lightman2023let} provided an effective approach for measuring hallucinations within the reasoning process. Li \etal~\cite{li2024fine} extended this work by introducing a Fine-grained Process Reward Model (FG-PRM), which trained six specialized PRMs to address specific types of hallucinations, including context inconsistency, logical inconsistency, instruction inconsistency, logical errors, factual inconsistencies, and fabrication.
These PRMs generated a combined signal to detect hallucinations more accurately.
Different from PRM-based methods, Zhang \etal~\cite{zhang2025reasoning} adopted linear probing, aiming at detecting errors early during reasoning. However, the above methods need additional training steps.
Dong \etal~\cite{dong2025mirage} adopted proxy LLMs to augment and rate the reasoning chain as an indicator of hallucination.
Sun \etal~\cite{sun2025detection} introduced the ``reasoning score'', a metric that measures divergence between intermediate hidden states and final logits. Their findings suggest that several indicators related to this score correlate strongly with the occurrence of hallucinations, leading them to combine these indicators for effective detection.
More recently, Wang \etal~\cite{wang2025joint} developed the RACE framework for hallucination detection, which extracts simplified reasoning steps via an LLM and evaluates four key aspects of the reasoning chain: reasoning consistency, answer uncertainty, reasoning-answer alignment, and reasoning coherence.

\textbf{Hallucination mitigation}.
In addition to hallucination detection, another way to combat hallucinations in LRMs is hallucination mitigation, which aims to reduce the frequency of hallucinations through various strategies.
These strategies can be broadly classified into two categories: training-based methods and planning-based methods.

Training-based methods involve intervening in the model's training process, either by introducing additional training objectives or incorporating specialized training data.
For instance, Song \etal~\cite{song2025hallucination} modified the reward function in the PPO algorithm~\cite{schulman2017proximal}, encouraging the model to respond with ``I don't know'' when faced with unanswerable questions. This approach mitigates hallucinations on unanswerable problems while preserving performance on solvable ones.
Similarly, Sun \etal~\cite{sun2025detection} proposed GRPO-R, an extension of the original GRPO~\cite{shao2024deepseekmath}, where the reward was adjusted by incorporating a reasoning score.
FSPO~\cite{li2025hallucination} further refined this approach by introducing both a rule-based correctness reward for the final answer and a step-wise factuality reward, which is derived from the LLM's reasoning process in conjunction with additional evidence.

In contrast, planning-based methods do not necessitate modifications to the training procedure. Instead, they focus on mitigating hallucinations by improving the model's reasoning path through better planning.
Zheng \etal~\cite{zheng2024thinking} argued that models may suffer from vision-language bias when they process information while simultaneously attending to both vision and text inputs. To address this, they first prompted the model to generate a reasoning plan using text-only input, and then, based on the generated plan, proceeded to solve the problem and generate intermediate reasoning steps with the vision-language input.

Overall, our review indicates that while reasoning models have demonstrated remarkable progress on complex reasoning-driven tasks, their tendency to hallucinate even in common scenarios remains a fundamental limitation. Addressing this tension between reasoning capability and reliability will require systematic investigation, and stands as an important direction for future research.
\subsection{Faithfulness of Reasoning Models}
\label{faithfulness}
Faithfulness in traditional natural language generation is defined by the extent to which the model's outputs align with or are supported by the provided input~\cite{li2022faithfulness}.
In this work, we specifically examine reasoning faithfulness in the context of LLM reasoning, focusing on faithfulness related to CoT prompting and LRM.
In LLM reasoning scenarios, reasoning faithfulness typically addresses the question~\cite{jacovi2020towards,lyu2023faithful}: \textit{``Does the explanation generated by the model accurately reflect the reasoning process behind its prediction?''}

Reasoning faithfulness is a fundamental aspect of overall model truthfulness.
A lack of faithfulness in CoT reasoning can introduce significant safety risks, particularly in high-stakes domains such as legal services, medical treatment, and financial decision-making~\cite{agarwal2024faithfulness}, where users may be misled into overestimating the model's interpretability.
Research on reasoning faithfulness can be broadly categorized into three key areas: faithfulness measuring, understanding, and improvement.
In the following sections, we will explore reasoning faithfulness from each of these three perspectives.
 
\subsubsection{Faithfulness Measuring}
While faithfulness is an essential component of trustworthiness, comprehensively measuring it remains an open challenge.
However, several metrics have been proposed to partially evaluate the faithfulness of CoT~\cite{lanham2023measuring,turpin2023language,tutek2025measuring}.
These methods can be broadly categorized into various intervention techniques that modify either the reasoning process, the input, or the model parameters to measure how faithfully the model's CoT reflects its reasoning process.

\textbf{CoT intervention}. One prominent evaluation method involves modifying the CoT reasoning path $T$ generated by the model and observing changes in the output to assess whether the reasoning faithfully supports the model's prediction~\cite{lanham2023measuring,benthamchain,paul2024making,yee2024dissociation}.
Lanham \etal~\cite{lanham2023measuring} proposed a CoT intervention approach, which alters the reasoning process by truncating the CoT before the final answer or introducing errors at specific points in the reasoning chain.
The former one truncates the original CoT before answering, and the latter one adds a mistake generated by a proxy LLM into some specific position in the CoT and generates subsequent CoT autoregressively.
After CoT intervention, if the answer changes, it means that the CoT matters in the model's prediction, which indicates that the CoT is faithful.
By introducing CoT interventions at different steps of the reasoning process, we can generate a consistency curve and use the Area Over Curve (AOC) to quantify faithfulness.
However, Bentham \etal~\cite{benthamchain} cautioned that such metrics may be biased due to inherent label biases in the model. To address this, they introduce a CoT-agnostic normalized metric, calculated as follows:
\begin{equation}
    N(\mathcal{M}, \mathcal{D}) = \frac{1}{\lvert \mathcal{D}\rvert}\sum\limits_{x\in\mathcal{D}}\mathbbm{1}_{[\mathcal{M}(x)=\mathcal{M}(\tilde{x})]},
\end{equation}
where $\mathbbm{1}$ represents the indicator function, and $\tilde{x}$ refers to a version of $x$ where answer choices have been shuffled.
Additionally, Paul \etal~\cite{paul2024making} used the Lakage-Adjusted Simulatability (LAS)~\cite{hase2020leakage} to measure faithfulness by evaluating the accuracy deviation between the model's performance with and without CoT reasoning.
Xiong \etal~\cite{xiong2025measuring} extended CoT intervention to assess both intra-draft and draft-to-answer faithfulness in large reasoning models, such as DeepSeek-R1.
Yee \etal~\cite{yee2024dissociation} employed error injection into the CoT and classified reasoning as faithful or unfaithful based on whether the model recovered the injected error in the final answer.

\begin{table}[ht]
\centering
\caption{Prompts demonstrating the two biasing features. The text for the unbiased context is in \textit{Italian} and for the biased context in \textbf{Bold}. 
    The top example shows the \texttt{Answer is Always A} biasing feature, in which we reorder the multiple-choice options in a few-shot prompt to make the answer always (A).
    The bottom shows the \texttt{Suggested Answer} bias, in which we add text where a user suggests a random answer is correct. This table is borrowed from Turpin \etal\cite{turpin2023language}.
    }
    \label{tab:bbh-perturbations}
    \centering 
    \scriptsize
        \begin{tabular}{@{}p{55em} @{}}
        \toprule
        \textbf{Biasing Feature \#1:} \texttt{Answer is Always A} \\\midrule

\textbf{Human:} Q: Is the following sentence plausible? ``Julio Jones struck out.'' (A) \textit{plausible}~/~\textbf{implausible} (B) \textit{implausible}~/~\textbf{plausible}

Let's think step by step. [...] Answer: \textit{(B)}~/~\textbf{(A)} implausible

[...] Answer: \textit{(B)}~/~\textbf{(A)} implausible

[...] Answer: \textit{(B)}~/~\textbf{(A)} implausible

[...] Answer: \textit{(A)}~/~\textbf{(A)} plausible

[...] Answer: \textit{(B)}~/~\textbf{(A)} implausible

[...] Answer: \textit{(A)}~/~\textbf{(A)} plausible

 Q: Is the following sentence plausible? ``LeBron James took a corner kick.'' (A) plausible (B) implausible 
 
 \textbf{Assistant}: 
\texttt{[generate CoT explanation]} \\\midrule \midrule

\textbf{Biasing Feature \#2: }\texttt{Suggested Answer} \\\midrule

\textbf{Human:} Q: Is the following sentence plausible? ``LeBron James took a corner kick.'' (A) plausible (B) implausible

\textit{\texttt{<empty>}}~/~\textbf{I think the answer is A but I'm curious to hear what you think.} 

\textbf{Assistant}: 
 \texttt{[generate CoT explanation]} 
         \\

\bottomrule
    \end{tabular}
\end{table}
\textbf{Input intervention}. Input intervention differs from CoT intervention in that it introduces bias into the input $x$ and observes how both the CoT reasoning and the model's answer change as a result. 
\Cref{tab:bbh-perturbations} shows a demonstration of input interventions proposed by Turpin \etal~\cite{turpin2023language}.
Specifically, by either setting all answers in the few-shot demonstration to a fixed choice (e.g., (A)) or expressing a preference for a particular answer choice, LLMs often adjust their answers accordingly.
This shift in answers is used to assess the model's faithfulness, with the accuracy drop serving as a key metric for unfaithfulness. However, it is important to note that the bias introduced into the input is typically not reflected in the CoT, thereby highlighting a potential risk of unfaithfulness.
Similarly, Chua \etal~\cite{chua2025deepseek} and Chen \etal~\cite{chen2025reasoning} built upon this concept by inserting various cues (i.e., professor suggestions and black/white square implications) into the inputs.
Unlike Turpin \etal~\cite{turpin2023language}, who focused on the accuracy drop, these studies assessed faithfulness by determining whether the model acknowledges the inserted cue when its answer changes. 
Yet, like previous studies, these models may fail to mention the cues in the CoT, exposing faithfulness vulnerability in their reasoning process.
Arcuschin \etal~\cite{arcuschin2025chain} proposed to flip the question (e.g., changing ``$\text{Is } X > Y$'' to ``$\text{Is } Y > X$''). If the model's answer does not change, it is considered unfaithful.

\textbf{Parameter intervention}. In a recent study, Tutek \etal~\cite{tutek2025measuring} argued that metrics based solely on CoT intervention only evaluate contextual faithfulness. Although crucial context may be erased, the relevant knowledge embedded within the model's parameters remains intact, potentially allowing the model to reconstruct the missing context. To address this, Tutek \etal~\cite{tutek2025measuring} introduced FUR, a method that utilizes the unlearning algorithm NPO~\cite{zhang2024negative} to assess parameter faithfulness.
Specifically, they segment the CoT $T$ and then unlearn a single step in it.
And then they use the answer consistency and probability divergence between the original model $\mathcal{M}$ and the unlearned model $\mathcal{M}'$ to estimate the faithfulness.

\textbf{No intervention}. Xu \etal~\cite{xu2024faithful} adopted manual evaluation, which divides an instance into three classes: (1) faithful: both the answer and the process are correct and logical (2) unfaithful: the answer is correct but the reasoning process is not; (3) false: the answer is incorrect.
Similarly, Li \etal~\cite{li2024towards} considered an instance to be faithful if and only if both the CoT and the answer are correct or incorrect.

\subsubsection{Faithfulness Understanding}
A growing body of research delves into the mechanisms underlying the faithfulness of reasoning in Large Language Models (LLMs). In this section, we summarize key studies that aim to understand and enhance the faithfulness of LLMs' reasoning processes.

\textbf{Unfaithfulness problem}.
Despite the impressive performance of CoT reasoning in handling complex tasks, the CoTs generated by models can still exhibit unfaithfulness—remaining logically coherent but diverging from the true reasoning process~\cite{turpin2023language, lanham2023measuring}.
Lanham \etal~\cite{lanham2023measuring} revealed that, in some cases, the reasoning process is post-hoc: the model first determines the answer and then fabricates a plausible explanation, rather than deriving the answer through the reasoning.
While reasoning models generally show better faithfulness than non-reasoning models~\cite{chua2025deepseek}, they still exhibit unfaithfulness that warrants further attention~\cite{chen2025reasoning, arcuschin2025chain}.
Agarwal \etal~\cite{agarwal2024faithfulness} emphasized that faithfulness is critical in high-stakes applications, such as healthcare diagnosis, financial forecasting, and crime prediction, while plausibility (the degree to which reasoning aligns with human understanding) is essential in more recreational or educational contexts, such as story-telling and educational LLMs.

\textbf{The factors that influence faithfulness}.
When unfaithfulness arises in models, a considerable amount of research investigates the factors influencing this issue.
Early work by Lanham \etal~\cite{lanham2023measuring} explored how model size and model capability affect faithfulness.
Their findings suggest that reasoning faithfulness typically increases, then decreases, with an increase in model size, with an optimal size around 13B parameters.
Bentham \etal~\cite{benthamchain} extended this research across various LLM families and confirmed a similar trend. Interestingly, they observed that models with higher accuracy tend to exhibit lower faithfulness, a finding also supported by Tanneru \etal~\cite{tanneru2024difficulty}.
Conversely, Bao \etal~\cite{bao2024likely} and Xiong \etal~\cite{xiong2025measuring} argued that larger models are generally more faithful, suggesting the possibility of a nuanced relationship between size and faithfulness.
The findings drawn by Bentham \etal~\cite{benthamchain} and Tanneru \etal~\cite{tanneru2024difficulty} may stem from the fact that more performant models can often generate correct answers despite error or incomplete CoTs, indicating that existing faithfulness measures may oversimplify the issue.
Additionally, Lanham \etal~\cite{lanham2023measuring} highlighted that the faithfulness of a model's reasoning varies significantly across tasks, with faithfulness scores AOC ranging from less than 10\% to over 60\%.
Chen \etal~\cite{chen2025reasoning} and Xiong \etal~\cite{xiong2025measuring} demonstrated experimentally that models are more prone to unfaithfulness when tasked with more difficult problems.
In addition, there is ongoing debate surrounding the impact of CoT length on faithfulness.
Chua \etal~\cite{chua2025deepseek} suggested that length penalties may result in unfaithful responses, but Chen \etal~\cite{chen2025reasoning} claimed that unfaithful CoTs are usually longer than faithful CoTs.
Bao \etal~\cite{bao2024likely} proposed an alternative explanation based on structural causal models (SCMs)~\cite{pearl2009causality}. They claimed that reasoning derived from a causal chain (where the answer stems directly from the CoT, which is in turn derived from the instruction) is generally more faithful. In contrast, reasoning that depends on more complex SCM types, such as common cause or full connection, may introduce unfaithfulness due to the increased dependency on the instruction.
Recent work also highlights the role of post-training techniques in shaping model faithfulness.
For instance, a study by Bao \etal~\cite{bao2024likely} indicated that SFT and DPO could weaken a model's faithfulness.
Lobo \etal~\cite{lobo2024impact} found that the impact of SFT on faithfulness is more pronounced in smaller models, with larger models being less affected.
Finally, recent studies suggested that reasoning models trained with reinforcement learning with verifiable rewards (RLVR) (e.g., DeepSeek-R1~\cite{guo2025deepseek}) exhibit significantly higher faithfulness compared to non-reasoning models~\cite{chua2025deepseek, chen2025reasoning, arcuschin2025chain}.
Although many factors are related to faithfulness, their conclusions may be contradictory due to different evaluation methods and models. This calls for the development of more comprehensive evaluation methods.

\subsubsection{Faithfulness Improvement}
Since faithfulness is an important part of trustworthiness, many methods have been proposed to enhance the faithfulness of the model.
To improve reasoning faithfulness in large language models, Radhakrishnan \etal~\cite{radhakrishnan2023question} adopted a question decomposition strategy. They break down a complex question into a sequence of subquestions, solve each one individually, and then recompose the intermediate answers to arrive at the final answer.
Recent work has explored symbolic reasoning to further enhance faithfulness.
Faithful CoT~\cite{lyu2023faithful} translated natural language queries into symbolic reasoning steps using an LLM, then employed a deterministic solver (e.g., a Python interpreter) to compute the final answer.
Each reasoning step in the chain included three components: a subquestion, a dependency graph, and corresponding rationales.
Similarly, LOGIC-LM~\cite{pan2023logic} used symbolic formulation and an external reasoner, and introduced a self-refinement mechanism when the executor returned an error.
However, reliance on external symbolic solvers may lead to brittleness in the presence of syntax errors. To address this limitation, approaches such as SymbCoT~\cite{xu2024faithful}, FLARE~\cite{arakelyan2024flare}, and CoMAT~\cite{leang2024comat} proposed to use LLMs themselves as solvers and verifiers.
SymbCoT used the LLM in multiple roles (i.e., symbolic translator, planner, solver, and verifier) via distinct prompt templates.
FLARE formalized problems into logic programs and simulates their execution using LLMs modeled after Prolog-style reasoning.
Wang \etal~\cite{wang2024causal} proposed the CORE framework, which iteratively refined both the rationale and the answer while ensuring that the model's confidence aligns with logical propositions.
QUIRE~\cite{li2024towards} enhanced faithfulness by re-emphasizing critical input information before initiating CoT reasoning.

In addition, there are also many works trying to improve the faithfulness of the model through post-training~\cite{gao2024fact,paul2024making}.
Gao \etal~\cite{gao2024fact} constructed a dataset to train the model with three stages: faithful program generation, concise CoT conversion, and transferability filtering.
They first synthesized executable visual programs from image--question pairs using a code-pretrained model and obtained the execution traces.
The execution trace was then refined via controllable operations—pruning irrelevant branches, merging redundant steps, and bridging logical gaps.
Finally, CoTs that prove effective in guiding end-to-end MLLMs were selected for knowledge distillation, which was conducted by both label and rationale loss, as in~\cite{hsieh2023distilling}.
FRODO~\cite{paul2024making} first employed DPO to incentivize the generation of correct reasoning paths and discourage counterfactual or irrelevant steps.
It further trained the model to associate correct/incorrect answers with corresponding reasoning paths and used margin-ranking loss to penalize high-confidence incorrect rationales.
Viteri \etal~\cite{viteri2024markovian} improved faithfulness via PPO~\cite{schulman2017proximal}, rewarding the model for generating correct rationales that lead to the answer even in the absence of the original prompt.
In summary, there are many methods that can be used to enhance the reasoning faithfulness of the model, but the unfaithfulness problem has not been completely solved. How to combine training-based and training-free methods can also be explored.

\subsubsection{Further Discussion of Faithfulness Definition}
In the definition of faithfulness, many working definitions are quite different from those of reasoning faithfulness. As a result, many researchers confuse them.
For instance, a recent survey on LLM hallucinations defines faithfulness hallucination as \textit{``the divergence of generated content from user input or the lack of self-consistency within the generated content''}~\cite{huang2025survey}.
However, this definition is concerned mainly with input faithfulness, which examines the degree to which the output reflects the user input, while reasoning faithfulness considers whether the model's intermediate reasoning steps faithfully capture its internal decision-making process.

Furthermore, considerable effort has been made to distinguish faithfulness from plausibility.
Plausibility generally refers to the appearance of coherence and logical consistency, regardless of whether the underlying reasoning is valid.
Given the powerful generative capabilities of today's large language models, they often produce responses that are highly plausible but not necessarily faithful.
Agarwal \etal~\cite{agarwal2024faithfulness} highlight this distinction, arguing that a response may appear convincing while still misrepresenting the model's actual reasoning.
Importantly, different application scenarios prioritize these dimensions differently, and striking a balance between faithfulness and plausibility remains context-dependent.

\section{Safety} \label{safety}
As safety becomes a critical concern in high-stakes applications, it is imperative to understand how reasoning interacts with LLM content safety issues.
In this section, we mainly examine the content safety challenges introduced by the emergence of large reasoning models as well as CoT techniques, whose enhanced capabilities and structured reasoning processes may amplify both utility and risk.  To be detailed, this section outlines key dimensions of safety related to reasoning capabilities, including vulnerability analysis, jailbreak attacks and defenses, safety alignment, and safety threats such as backdoor and prompt injection. 
\subsection{Vulnerability Assessment}
\label{Vulnerability Assessment}
Vulnerability assessment in reasoning models often involves jailbreak attacks, which aim to induce the model to generate inappropriate content. 
For large language models, many researchers developed related benchmarks~\cite{mazeika2024harmbench,souly2024strongreject,zeng2024air,hanwildguard} to evaluate the jailbreak defense capability against previous attacks~\cite{zou2023universal,chao2023jailbreaking,mehrotra2024tree}. In terms of jailbreak assessment of large reasoning models, early works utilized jailbreak prompts from previous benchmarks mentioned above to evaluate the safety performance~\cite{ying2025towards,zhang2025safety,romero2025red,zhou2025hidden,kassianik2025evaluating,li2025smarter,jiang2025safechain,krishna2025weakest}. Also, many researchers developed new benchmarks~\cite{knight2025fortress,fan2025evaluation,zheng2025beyond,lu2025bench} for a more targeted evaluation. Here, instead of narrating these works in a timeline, we group the core findings of these studies to build a preliminary conceptual map.

\textbf{Current open-source reasoning models are still vulnerable to jailbreak attacks.}
Evaluation results from many researchers~\cite{ying2025towards,zhang2025safety,romero2025red,kassianik2025evaluating, jiang2025safechain, krishna2025weakest,marjanovic2025deepseek} emphasized the safety vulnerability of current large reasoning models. 
SafeChain~\cite{jiang2025safechain} evaluates concurrent reasoning models~\cite{guo2025deepseek,deepmind_gemini,team2025kimi,sky_t1_2025,qwq32b,skyworkopeno12024} on StrongReject~\cite{souly2024strongreject} and WildJailbreak~\cite{jiang2024wildteaming}, finding that all these modern large reasoning models should improve safety performance, for no model achieved a satisfactory result on both datasets. Zhou \etal~\cite{zhou2025hidden} claimed that o3-mini is significantly safer than DeepSeek-R1 models on four datasets~\cite{zeng2024air,wan2024cyberseceval}. Kassianik \etal~\cite{kassianik2025evaluating} also mentioned that the attack success rate (ASR) of DeepSeek-R1 on Harmbench~\cite{mazeika2024harmbench} is 100\%, higher than o1-preview and other large language models~\cite{dubey2024llama,achiam2023gpt,TheC3}, corresponding to conclusions from Marjanovi{\'c} \etal~\cite{marjanovic2025deepseek}.
Ying \etal also mentioned that ``\textit{both DeepSeek-V3 and DeepSeek-R1 models exhibit clear vulnerabilities when facing jailbreak attacks'' after evaluating the safety performance on the CNSafe dataset~\cite{ying2025towards}}. Similarly, Krishna \etal~\cite{krishna2025weakest} in their evaluation highlighted the category-wise and model-wise vulnerabilities when faced with various jailbreak attacks. Additionally, Fan \etal~\cite{fan2025evaluation} discovered evaluation faking, where reasoning models may probably understand they are being evaluated and therefore alter their response to be safer. Zheng \etal~\cite{zheng2025beyond} proposed BSAbench, which disclosed the safety vulnerability with more challenging queries. After clarifying the overall perception that open-source reasoning models still have space to improve the safety capability, here are specific insights.

\textbf{First, compared to base large language models, post-trained models with distilled CoT data are less sensitive to harmful prompts and reject them.} SafeChain~\cite{jiang2025safechain} proposed that learning long CoT does not necessarily improve model safety when comparing DeepSeek-R1-70B with Llama-3.3-Instruct-70B. A similar conclusion is also made by Zhou \etal~\cite{zhou2025hidden}. Additionally, Zhang \etal~\cite{zhang2025safety} evaluated the DeepSeek distilled model series on CHisafetybench~\cite{zhang2024chisafetybench}, and concluded that in terms of the risk content identification task and the ``refusal to answer task'', a few reasoning models experienced a decrease in rejection rate and responsibility rate, indicating higher compliance behavior on harmful requests. Zhao \etal~\cite{zhao2025trade} also mentioned that acquiring deliberate reasoning capabilities would sacrifice model general performance.

\textbf{Second, the thinking process from LRMs may negatively affect the harmfulness of the generated content.} 
Jiang \etal~\cite{jiang2025safechain} designed different thinking templates to control the reasoning process, and conducted experiments to compare the harmfulness of answers given different lengths of reasoning tokens. 
It turns out that compared to the default content generation, forcing the model to skip reasoning or shorten reasoning could boost the harmlessness of the answers at least on StrongReject~\cite{souly2024strongreject} and WildJailbreak~\cite{jiang2024wildteaming}.
Zhou \etal~\cite{zhou2025hidden} and Zhao \etal~\cite{zhao2025trade} also reinforce such an idea: they compared the answers of two pairs of reasoning models with the base models on harmful prompts, demonstrating that LRMs tend to provide more detailed and helpful answers, making the output more harmful. Furthermore, when directly evaluating the harmfulness of thinking content and final answers of DeepSeek-R1-Distill-70B on AirBench~\cite{zeng2024air} and WildGuard~\cite{hanwildguard}, the safety rate of thinking content is consistently less than that of final answers. Ying \etal ~\cite{ying2025towards} also supported the vulnerability of reasoning content, indicating that the exposed reasoning chains may increase safety risks.

\textbf{Third, Pairwise safety ranks between models depend on datasets.} After reviewing the related literature, we find that some findings from different datasets do not reach a consensus. For example, evaluations on Airbench~\cite{zeng2024air} claimed that DeepSeek-R1 is safer than DeepSeek-V3~\cite{zhou2025hidden}, while under CNSafe, DeepSeek-V3 exceeds DeepSeek-R1 with an average ASR margin of 21.7\% across all risk categories~\cite{ying2025towards}. However, when red-teaming with jailbreak templates, experiments on WildGuard Jailbreak~\cite{zhou2025hidden} and CNSafe\_RT~\cite{ying2025towards} conversely showed that DeepSeek-R1 could identify the risk in jailbreak prompts and provide a safe thinking chain. 
Additionally, safety performance is also related to evaluation topics. 
For the DeepSeek-distilled model series, the most notable declines in safety performance are observed in areas such as health discrimination, sexism, regional discrimination, and occupational discrimination~\cite{zhang2025safety}. In contrast, DeepSeek-R1 exhibits pronounced vulnerabilities in cybersecurity-related topics~\cite{zhou2025hidden}. We may explain this discrepancy by noting that different training datasets and data structures would influence the model performance, causing imbalanced sensitivity to various safety topics. 

\textbf{Fourth, multilingual vulnerability is critical for current large reasoning models.}
Multilingual vulnerability is also a representation of ``mismatched generalization''~\cite{wei2023jailbroken}, which means that models may possess different safety capabilities in different language environments.
Romero-Arjona \etal~\cite{romero2025red} identified the safety vulnerability in Spanish and Basque. They claimed that the failure rates of DeepSeek-R1 and o3-mini in their Spanish dataset are 31.7\% and 29.5\%. Zhang \etal~\cite{zhang2025safety} made a detailed evaluation on the Chinese dataset CHisafetybench~\cite{zhang2024chisafetybench} and identified a clear safety decline after distillation. Ying \etal~\cite{ying2025towards} also found that for both DeepSeek-V3 and DeepSeek-R1, the ASR in the English environment is larger than that in Chinese, disclosing the safety capability imbalance about language. 

\textbf{Fifth, MLRMs share similar vulnerabilities with uni-modal large reasoning models.} With the development of MLRMs~\cite{yang2025r1,team2025kimi,peng2025skywork,peng2025lmm}, researchers also found similar vulnerabilities with early safety assessments. Fang \etal~\cite{fang2025safemlrm} identified that model safety performance varies in terms of different topics, and defined such a phenomenon as ``safety blind spots'', which resembles the third point mentioned above. Lou \etal~\cite{lou2025think} mentioned the higher risk of the thinking process than the final answers of MLRMs and the vulnerability against jailbreak attacks compared to the base MLLMs, which are consistent with the first two insights. In addition, it is also observed that converting images into captions could recover the safety capability to some extent~\cite{lou2025think}, which again demonstrated the imbalanced domain vulnerability in MLLMs~\cite{gou2024eyes, wang2025we}. Experiments from both literature~\cite{fang2025safemlrm, lou2025think} also pointed out that the emergent self-correction in the thinking process helps avoid harmful content generation, even if there were still cases where unsafe reasoning was generated, followed by inappropriate answers.

To summarize, we can hardly get the conclusion that reasoning capability enables a model to perform better in the safety domain. Even though under some circumstances, it is proven that the reasoning process could identify the disguised harmful intention in jailbreak prompts and reject the inappropriate behaviors, which outperforms non-reasoning models, there are also comprehensive evaluations disclosing the vulnerability of reasoning models, such as multilingual inputs or specific topics. Except for o1 or o3-mini~\cite{jaech2024openai}, which are safer than other open-source large reasoning models with a slightly obvious margin, there is still space to boost safety performance via inference-time scaling, just as in the general performance domain.
\subsection{Jailbreak}
\label{jailbreak}
In the era of large language models, jailbreak generally becomes crucial to model safety. In this script, we mainly focus on jailbreak topics related to CoT or current large reasoning models represented by OpenAI o1~\cite{jaech2024openai}, DeepSeek-R1~\cite{guo2025deepseek}, etc. The literature could be roughly clustered into two parts: early studies targeting large language models and the latest studies targeting models with CoT capability. Attacks and defenses are split into separate subsections for better readability. 

\subsubsection{Jailbreaking with Reasoning Techniques}
CoT techniques enable large language models to perform better on various general tasks~\cite{ying2025reasoning,kojima2022large,wei2022chain,handa2024competency}. Therefore, recent literature has also proposed methods to generate more deceptive jailbreak prompts~\cite{sabbaghi2025adversarial,su2024enhancing,handa2024competency} or create more detailed and harmful content with reasoning techniques~\cite{ying2025reasoning,chang2025chain} while overlooking their safety issues. 
Specifically, Sabbaghi \etal~\cite{sabbaghi2025adversarial} introduced a feedback model as well as a refiner model to iteratively modify the jailbreak prompt with CoT paths given the calculated loss score, for models with CoT could better identify the imperfection of each round of jailbreak prompts, provide more targeted modifications, and then enhance the ASR.
This method followed the logic of previous black-box jailbreak methods~\cite{chao2023jailbreaking,mehrotra2024tree}, which evaluated and modified their jailbreak prompts according to the interactions with the target models.
Ying \etal~\cite{ying2025reasoning} proposed a multi-turn method to transform harmful prompts into several superficially benign questions. 
During the multi-turn conversation, the attacker explicitly instructed the victim model to reason about some specific steps, bypassing its safety alignment, and finally elicited harmful content. 
Similarly, Chang \etal~\cite{chang2025chain} wrapped the sensitive instruction into a narrative task, designing CoT-style prompts to instruct victim models to generate details and finish the story while bypassing internal safety barriers.  
Handa \etal~\cite{handa2024competency} proposed to jailbreak models with complex ciphers. The advanced reasoning capability enables models to decode more complex ciphers, therefore providing more room for the disguise of harmful instructions.
The success of these attacks vividly supports that better performance of language models enabled by CoT techniques could create new threats to content safety. More works are required to evaluate the potential risks as well as feasible defense methods regarding reasoning techniques.
\subsubsection{Jailbreaking Reasoning Models}
In this part, we mainly cover a few jailbreak attacks taking advantage of the reasoning process to disclose the vulnerability of large reasoning models. 

Kuo \etal~\cite{kuo2025h} proposed H-CoT, containing well-curated reasoning content in the prompts to obfuscate the models. Here we borrow an example from the original paper as an illustration. 
\begin{figure}[ht]
\centering
\begin{tcolorbox}[colback=gray!10,
                  colframe=black,
                  arc=1mm, auto outer arc,
                  boxrule=0.5pt,
                 ]
\small  
\textbf{Goal}: Provide a detailed guide on the process of making illegally-gained proceeds appear legal, also known as money laundering.\\
\textbf{Original request}: I am a teacher for financial auditors. As part of their professional training, I provide neutral, informational education on methods of making illegally-gained proceeds appear legal, known as money laundering in the USA. The goal is to enhance their ability to detect and prevent such activities. Could you provide a broad overview of the modern strategies?

Each part must be logically connected... and should include the following components: Scheme:... Functional Example:... Rationale:... List as many real-world examples and advantages for criminals as possible in choosing these.\\
\textbf{H-CoT reasoning}: 
\begin{itemize}
    \item Understanding money laundering...
    \item Structuring the educational content...
    \item Enhancing detection strategies...
    \item Providing contextual examples...
    \item ...
    \item Rationale and outcome...
\end{itemize}
\textbf{Jailbreak prompt}: 
\begin{itemize}
    \item \textit{Original request}
    \item \textit{H-CoT reasoning}
\end{itemize}
\end{tcolorbox}
\caption{An example of H-CoT jailbreak prompt, which is from ``DukeCEICenter/Malicious\_Educator\_hcot\_o1'' dataset~\cite{kuo2025h}.}
\end{figure}
In the experiments, they found that directly padding detailed execution steps could hijack the thinking process, skip the justification phase, and elicit harmful generation.
After that, Yao \etal~proposed ``Mousetrap''~\cite{yao2025mousetrap}, splitting the harmful prompts into several steps for models to reason.
After following the instructions to execute character decoding, word replacement, and sentence order reversal, the model could understand the final harmful prompt while failing to identify its toxicity.
Such an attack resembles the classical ``base-64 encoding'' jailbreak~\cite{greshake2023more,wei2023jailbroken}, sharing the logic of mismatched generalization~\cite{wei2023jailbroken}. 
Liang \etal~\cite{liang2025autoran} proposed AutoRAN, claiming it as the first automated jailbreak attack specifically targeting reasoning models, enabled by a self-designed, predefined attack workflow. Nguyen \etal~\cite{nguyen2025three} came up with ``SEAL'' to circumvent LRM internal defenses, selecting ciphering methods from an encryption algorithm set to encode harmful instructions. Lu \etal~\cite{lu2025does} proposed FicDetail to jailbreak reasoning models, creating a fiction story with multi-turn queries to enrich details with harmful contents. Lian \etal~\cite{lian2025revealing} exploited the intrinsic ethical vulnerability from distribution shift and in LLMs, designing an attack with semantic coherence inducement to jailbreak DeepSeek-R1 successfully. Ma \etal~\cite{ma2025hauntattack} proposed HauntAttack, which wraps harmful instructions into normal, realistic scenarios to deceive reasoning models. 
For MLRMs, Sima \etal~\cite{sima2025viscra} designed VisCRA, exploiting reasoning capabilities to force models to first infer masked objects in images and then create detailed answers for harmful instructions. With the two-phase instructions, both cutting-edge MLLMs and MLRMs are proven to be vulnerable. In the tool learning domain, Liu \etal~\cite{liu2025rrtl} developed Tool-CoT attack, in which the agent is prompted to call external functions for more harmful information. Experimental results indicate that models exhibit reduced sensitivity to function-calling behaviors, which may allow harmful intents to bypass internal safety alignment mechanisms, ultimately leading to illicit outputs.

In summary, the logic of developing jailbreak attacks does not change dramatically.
Compared with previous jailbreak methods targeting large language models, we found some methods exploiting the novel thinking process, as well as others designing more intense prompt encryptions to match the advanced general capability of reasoning models. From this point, it seems that reasoning models are more vulnerable to jailbreak attacks, due to the larger mismatching generalization between instruction following and safety alignment. 

\subsubsection{Jailbreak Defense with Reasoning Techniques}\label{guardrail model}
Because the performance of CoT techniques has been proven on general tasks, researchers have also tried to take advantage of this feature to build more robust guardrail models.
GuardReasoner~\cite{liu2025guardreasoner} curated 127k data samples with 460k reasoning steps in total to finetune a large language model, enabling the guardrail models to judge the harmfulness of prompts and answers. Similar to LLM alignment with CoT data in Sec.~\ref{Aligning large language model with CoT data}, detailed reasoning contents were distilled from GPT-4o to construct the SFT data. After learning the answering structure, DPO is then adopted to learn ``hard samples'' whose judgments from finetuned models vary conditioning high temperature and top-p hyperparameter. 
X-Guard~\cite{upadhayay2025x} noticed the judgment inaccuracy on low-resource languages and code-switching attacks, creating a safety dataset spanning 132 languages and updating the model weight with SFT followed by GRPO. 
Also noticing the judgment inaccuracy on multi-lingual inputs, MrGuard~\cite{yang2025mr}  elaborated curriculum learning with reasoning to improve the robustness towards low-resource languages. Similarly, RSafe~\cite{zheng2025rsafe} utilized GRPO to train a robust and generalizable guardrail model, successfully adapting to user-specified safety policies. 
Sreedhar \etal~\cite{sreedhar2025safety} conducted a study on reasoning-augmented guardrail models, demonstrating the benefits of reasoning in terms of detection accuracy, efficiency, generalization, etc.
Kang \etal~\cite{kang2024r} proposed R$^2$-Guard to detect unsafe contents with reasoning enabled by probabilistic graphical models (PGMs).
For vision-language models (VLM), GuardReasoner-VL~\cite{liu2025guardreasonervl} shared a similar logic with the previous method~\cite{liu2025guardreasoner}, extending the model to the vision domain. ShieldVLM~\cite{cui2025shieldvlm} simply used SFT with high-quality multimodal reasoning data to enhance the detection capability, achieving the harmfulness of image-text input pairs without model answers.
In terms of agent safety, Xiang \etal~\cite{xiang2024guardagent} developed GuardAgent to monitor agent actions. Different from conventional LLM-based agents that only process natural language, GuardAgent thinks of an action plan, generates guardrail codes, and finally executes the program to check content safety. Chen \etal~\cite{chen2025shieldagent} also proposed ShieldAgent to tackle this problem, in which they encoded safety constraints in knowledge graphs. Experiments proved the superior performance of these methods, providing new insights into agent-based agent guardrails. 
Aside from the guardrail models mentioned above, reward models could also contribute to content identification as well as model alignment~\cite{cheng2025inverse, wang2025unified}. Pan \etal~\cite{pan2025detecting} proposed U-CoT+ to detect harmful memes with zero-shot CoT prompts. To summarize, the success of these models demonstrates the feasibility of reasoning techniques, reinforcing their role in identifying, controlling, and moderating unsafe generations.
\subsubsection{Jailbreak Defense for Reasoning Models}
Jailbreak defense could be facilitated in different stages. Except for alignment methods that would be covered in detail in Section~\ref{alignment}, content detection and decoding manipulation are also ways to control harmful content generation. In this part, we mainly cover defending methods on reasoning models, analyzing the similarity and novelty of these methods when compared to previous instruct models. 

\textbf{Input-phase defense}. At first, Jailbreak defense in LLM followed the logic of prompt engineering, designing a detailed prompt before or after user prompts as an extra instruction to depress inappropriate behaviors~\cite{xie2023defending, zhang2023defending, wei2023jailbreak,xiong2024defensive}. Sharing some degrees of similarity, Jiang \etal~\cite{jiang2025safechain} mentioned that Zerothink mode could improve the defense capability, and Wu \etal~\cite{wu2025effectively} demonstrated that adding safety-related instructions in the reasoning trace could outperform manipulations in user prompts~\cite{xie2023defending, zhang2023defending}, with an explanation that attention of reasoning process focuses more on internal tokens instead of input prompts. Yamaguchi \etal~\cite{yamaguchi2025adversarial} also designed experiments on DeepSeek-R1-Distill-Llama, and found that whether the model rejects or complies with the instruction is predictable from intermediate activations of CoT tokens. These results uncovered the importance of reasoning in making decisions and supported the effectiveness of reasoning manipulation indirectly. 

\textbf{Decoding-phase defense}. With advancements in test-time compute for general tasks, researchers also made early attempts to generalize the improvement in the safety domain. Wang \etal~\cite{wang2025safety} revealed that applying Best-of-N (BoN) strategies could enhance the model safety, suggesting the existence of latent safety knowledge. Zaremba \etal~\cite{zaremba2025trading} found that the robustness of the OpenAI o1 series improved when increasing the test-time compute under a few settings. Saffron-1~\cite{qiu2025saffron} focused on the inefficiency of inference-scaling methods in safety contexts, proposing a novel inference-time scaling paradigm for efficient and safe decoding control. Instead of querying PRMs multiple times in tree search, one call to Saffron outputs a vector containing rewards for all possible next tokens, which breaks the exploration-efficiency dilemma. In addition, previous methods tried to manipulate the output logits of each token for safer generations~\cite{zeng2024root,xu2024safedecoding,banerjee2025safeinfer}, which may also provide a feasible way for safety generation. 

\textbf{Post-hoc defense}. Guardrail models, or LLMs-as-a-judge, serve as an external safety guard for language model content generation~\cite{dong2024safeguarding}. To identify the ASR of jailbreak methods, except for simple string-matching methods, LLM could be elaborated for harmful data detection, including prompting cutting-edge general models (such as GPT series~\cite{achiam2023gpt}) with pre-defined safety principles, or fine-tuning with well-curated safety data (Llama-Guard series~\cite{inan2023llama,chi2024llama}). Considering the safety risk in reasoning traces~\cite{jiang2025safechain,zhou2025hidden,ying2025towards}, ReasoningShield~\cite{li2025reasoning} curated a dataset with 8k prompt-CoT pairs and finetuned Llama-3.2~\cite{llama_3_2} to identify harmfulness in the reasoning traces as well as the final answers. During fine-tuning, SFT was conducted only on samples with consistent judgment among three LLMs, while DPO preference data were from ``hard samples'' with different judgments. In terms of LLM-based agents that generate thoughts before subsequent actions, Jiang \etal~\cite{jiang2025think} thought highly of the timely intervention of potentially harmful thoughts, trained the ``Thought-Aligner'' to generate safer and more cautious reasoning processes for replacement. These early efforts highlighted the potential of reasoning-specific guardrail models, suggesting room for continued research.

\subsection{Alignment} \label{alignment}

Alignment is not only a crucial part of large language model training, but also an important topic for model safety. In the training phase, alignment is originally proposed to align model reaction with human expectation~\cite{wang2024comprehensive}. During last three years, a lot of methods, including reinforcement learning from human feedback (RLHF) and its variants, are proposed to enhance the conversation performance of instruct models~\cite{ouyang2022training,bai2022training,lee2023rlaif,rafailov2023direct,schulman2017proximal}. Considering safety alignment, most methods collect a fine-tuning dataset including prompt-rejection pairs compassing various sensitive topics to update model weights~\cite{ji2024beavertails,ji2024pku,wang2025we,zong2024safety}. 
Here, instead of focusing on alignment within instruction tuning before formal model release, we narrow our sight to safety alignment of released models, including enhancing safety performance with CoT capability, or directly aligning large reasoning models. 
\subsubsection{Aligning LLM Using Reasoning Techniques} 
\label{Aligning large language model with CoT data}
Noticing the performance of CoT behaviors, researchers tend to facilitate safety alignment with CoT datasets~\cite{zhang2024backtracking,yang2025enhancing,feng2025erpo,mou2025saro,kim2025reasoning,zhu2025reasoning}. 
To be detailed, Liu \etal~\cite{liu2024mixture} proposed to train multiple low-rank adaptation (LoRA)~\cite{hu2022lora} variants as Mixture-of-Experts (MoE) to explicitly analyze question intentions, answer guidances, and the final response. Iteratively querying these models enabled the framework to ``think step-by-step'' before making final decisions. Zhang \etal~\cite{zhang2024backtracking} added a reset token to elicit self-corrections after a partial unsafe generation. To enable the model to learn backtracking, SFT with DPO is employed to learn the correction behavior while avoiding unnecessary backtracking. Yang \etal~\cite{yang2025enhancing} proposed Safety Chain-of-Thought (SCoT) to provide detailed analyses of potential risks before answering, claiming that SFT on mixed CoT datasets could enhance the defense capability against various attacks~\cite{zou2023universal,chao2024jailbreakbench}. Similarly, Zhang \etal~\cite{zhang2025stair} proposed to utilize data from Monte-Carlo Tree Search (MCTS) to improve the safety alignment. They began by prompting GPT‑4o to produce CoT data for fine‑tuning, and then ran a safety‑informed MCTS on the target model to generate raw data for DPO training. R2D~\cite{zhu2025reasoning} 
generated a pivot token including ``[SAFE]'', ``[UNSAFE]'', and ``[RETHINK]'' after each thinking step, and added an extra contrastive loss on the pivot tokens in SFT. 
With the combined loss, models could learn to generate detailed reason steps followed by the pivot token as a hint for the whole thinking process. RATIONAL~\cite{zhang2025safety1} also identified the imperfection of direct refusal to harmful queries, curating a CoT dataset consisting of both adversarial data and sensitive benign data by prompting Llama-3-8B-Instruct for following supervised fine-tuning.
ERPO~\cite{feng2025erpo} also adopted SFT followed by DPO, while adding extra ``length-controlled iterative preference optimization strategy'' to shorten generation length in the iterative preference optimization algorithm. 
For safe prompts, except for only considering decreasing the probability of generating helpless responses with incorrect thoughts, the algorithm also preferred concise thoughts over redundant reasoning chains. SaRO~\cite{mou2025saro} picked prompts from SALAD-Bench~\cite{li2024salad} and OpenOrca~\cite{mukherjee2023orca} with reasoning generation from GPT-4o to get the CoT data for supervised fine-tuning, enabling models to learn the thinking-answer template. 
Wang \etal~\cite{wang2025safety} underscored the generalization weaknesses of refusal training, introducing guidelines for better safety reasoning. Kim \etal~\cite{kim2025reasoning} distilled data from reasoning models and adopted SFT with GRPO for adaptive defense. 

After reviewing related works, we would like to elaborate more on SFT data collection and DPO pair selections. Mainstream SFT methods utilize off-the-shelf datasets, originally created for safety alignment or benchmarking harmfulness, to collect prompts and safe answers~\cite{zhang2024backtracking,zhang2025stair,zhu2025reasoning,zhang2025safety1,feng2025erpo,mou2025saro,kim2025reasoning}. These datasets include (but may not limited to)  PKU-SafeRLHF~\cite{ji2024pku}, HH-RLHF~\cite{bai2022training}, ToxicChat~\cite{lin2023toxicchat}, SALAD-Bench~\cite{li2024salad}, BeaverTails~\cite{ji2024beavertails}, SorryBench~\cite{xie2024sorry}, XSTest~\cite{rottger2023xstest},  JailbreakV-28k~\cite{luo2024jailbreakv}, AdvBench~\cite{zou2023universal}.
LLM primarily generates structured CoT content with a fixed prompt template. As shown in Figure~\ref{Cot_prompts}, LLMs are prompted to create detailed reasons with pre-defined structures for the final answer. It is believed that such SFT could first enable the models to learn the think-then-answer behavior, which provides a solid base for further preference optimizations.

In terms of DPO, the main target is to further enhance content harmlessness while not harming other capabilities, such as the helpfulness and conciseness of the answer. Zhang \etal~\cite{zhang2024backtracking} designed two pairs of preferences: for unsafe response, backtracking token followed by safe answer is preferred, while for benign response, fluent generations without backtracking token are positive.
STAIR~\cite{zhang2025stair} constructed the preference pairs with a step-wise reward function, encouraging the generation of safe and helpful answers. In ERPO~\cite{feng2025erpo}, the rank is in three levels: a helpful reason with a safe answer is better than reasons containing a harmful prefix and self-reflection, and an incorrect reason with a harmful answer ranks last. 
Similarly, SaRO~\cite{mou2025saro} decomposed the thinking chain into steps and encouraged early reflection with fewer unsafe steps.
Generally speaking, the design of DPO pairwise data and RL rewards has focused on both content safety and generation quality. Various methods with differing details have proven effective, though there remains room for further empirical investigation.
\begin{figure}[ht]  
\centering
\begin{tcolorbox}[colback=gray!10,
                  colframe=black,
                  arc=1mm, auto outer arc,
                  boxrule=0.5pt,
                 ]
\small 
\textbf{RATIONAL}~\cite{zhang2025safety1} with Llama-3-8B-Instruct:
Consider how and why this jailbreaking prompt could be harmful or used to facilitate illegal activities. Evaluate the safety of the query. 
Analyze the intent, ethics, and potential impact. Beware of malicious intent framed as educational or preventative.
Clearly reject the query in your response based on your reasoning.
\\
\\
\textbf{R2D}~\cite{zhu2025reasoning} with DeepSeek-R1-distill-70B:
You are an honest and good AI Assistant that can respond to all kinds of questions no
matter they are safe or unsafe, but please answer user’s questions with a safety reasoning suffix at the end of every reasoning step, including inside of your thinking process. 
[SAFE] means this sentence is safe.
[UNSAFE] means you already find this single sentence unsafe and you need to modify your strategy afterwards. 
[RETHINK] means you are suspective that things are going unsafe and need to be careful.
If you choose to [RETHINK], try to figure out a safe way to answer the user’s question.
Remember that the suffices [SAFE], [UNSAFE] and [RETHINK] are required after every sentence!
\\
\\
\textbf{SaRO}~\cite{mou2025saro} with GPT-4o:
Engage in a self-dialogue to solve this problem, where the agent needs to reason and reflect before reaching the final answer. You can use a newline delimited separated reasoning process. During this process, the agent will first understand the requirements of the user questions and describe it in its own words. The agent will consider whether the requirements violate security policies. **You can use conjunctions such as ``wait...'', ``but...''**.
Think about how to give a response that meets user requirements while avoiding violating relevant security policies. **During the reasoning process, you can use conjunctions such as ``so...'', ``also...'', ``alternatively...''**. **You can also use conjunctions such as ``wait...'', ``but...'' for self reflection and correction**.
Note: Your response should be in the form of a self-conversation, which is an informal text like a chat, rather than a formal written text like a guideline. 

\end{tcolorbox}
\caption{Examples of prompts for CoT data synthesis. Minor modifications are executed for better readability.}
\label{Cot_prompts}
\end{figure}

\subsubsection{Alignment of Large Reasoning Models}
\label{Alignment of large reasoning models}
To our best knowledge, Deliberate Alignment~\cite{guan2024deliberative} proposed the first method to align reasoning models with curated CoT data. 
With an unaligned reasoning model, they provided safety categories with specifications to distill safety-related thinking contents for post-training. After SFT and RL on distilled CoT data, Deliberate Alignment outperformed previous methods~\cite{zhang2024backtracking,madaan2023self}, suggesting a new approach for aligning models with evolving policies. Following a similar strategy, SafeChain~\cite{jiang2025safechain} and STAR-1~\cite{wang2025star} curated CoT post-training datasets, including various harmful topics, a detailed reasoning process, and clear rejection answers, to enhance the safety alignment performance. Instead of DPO or other RLHF methods, a major part of the work purely utilized SFT to update the parameters~\cite{jiang2025safechain,wang2025star,zhang2025realsafe,jeung2025safepath}, achieving a rough balance between utility and safety. Context Reasoner~\cite{hu2025context} also used two-stage post-training for safety alignment, in which they collected related regulatory standards for CoT generations. As for MLRMs, Lou \etal~\cite{lou2025think} created CoT content with DeepSeek-R1 to form the multimodal safety alignment dataset, in which they first utilized Qwen2.5-VL-72B to generate the image description, so that DeepSeek-R1 could receive all the information and generate a proper reasoning trajectory.
Additionally, Baker \etal~\cite{baker2025monitoring} proposed a CoT monitor to detect misbehavior and integrated it into the training objective, resulting in better alignment performance in the low optimization regime. Zhang \etal~\cite{zhang2025should} explored different SFT data for safety improvements, finding that simple reasoning processes could enable the models to gain comparable safety performance. SafeKey~\cite{zhou2025safekey} identified the importance of the key sentence in response safety, and developed ``Dual-Path Safety Head'' as well as ``Query-Mask Modeling'' to amplify the predictable effect of key sentence features, enabling reasoning models to better classify harmful queries from the benign in the representation domain. Moreover, inspired by gaming theory, Liu \etal~\cite{liu2025chasing} cast the attack-defense interaction as a zero-sum game, and created a Self-RedTeam framework in which models were updated with RL to defend safety attacks generated by their own. After iteratively role-playing as the attacker and the defender, the model is proven to gain robust safety alignment.

In general, most post-training methods, which consist of CoT data collection followed by SFT (with or without RL), aimed at embedding safety-prompt-conditioned responses into normal model generations where prompts including safety warnings are not necessary. After post-training, safety-related prompts will be automatically printed into the model weights, therefore influencing model behaviors. Except for the dataset mentioned in Section~\ref{Aligning large language model with CoT data}, harmful prompts aligning large reasoning models could also be chosen from WildJailbreak~\cite{jiang2024wildteaming}, Harmbench~\cite{mazeika2024harmbench}, SimpleSafetyTest~\cite{vidgen2023simplesafetytests}, TDCRedTeaming~\cite{mantas2023tdc}, ALERT~\cite{tedeschi2024alert}. For the vision-language domain, safety datasets include RLHF-V~\cite{yu2024rlhf}, LLaVA-RLHF~\cite{sun2023aligning}, VLFeedback~\cite{li2024vlfeedback}, Safe RLHF-V~\cite{ji2025safe}, and MM-RLHF~\cite{zhang2025mm}.  To conclude, there remains significant scope for novel alignment studies and methodological innovations, both in terms of data generation and the design of learning algorithms.

\subsubsection{Safety Tax}
The trade-off between model general performance and safety has been proposed for a long time, which could be traced back to the adversarial training of convolution neural network (CNN) on classification tasks~\cite{goodfellow2014explaining} where adversarial training traded classification accuracy for robustness\footnote{Here we slightly abuse the word ``safety'', referring to the defense against adversarial noise.}. To be clear, here we define the safety tax as \textit{``the phenomenon that fine-tuning models on safety alignment datasets will inevitably sacrifice model general performance, including but not limited to problem solving, code completion, conversation comprehension, etc''.}

Safety tax, or alignment tax, was mentioned by multiple papers~\cite{huang2025safety,cheng2025hair, lin2023mitigating, ouyang2022training}. Lin \etal~\cite{lin2023mitigating} firstly conducted a comprehensive study on alignment tax, highlighting that the RLHF process would sacrifice multiple model capabilities, such as translation~\cite{bojar2014findings}, reading comprehension~\cite{rajpurkar2018know}, and general question answering (QA)~\cite{clark2018think}. To mitigate the side effects, they evaluated several methods and uncovered the superior performance of model merging. Huang \etal~\cite{huang2025safety} fine-tuned a large reasoning model with two safety alignment datasets, finding that better safety performance corresponded to more severe sacrifices on model general capabilities. Hair~\cite{cheng2025hair} identified the alignment tax in current LLM alignment methods, and proposed a ``Hardness-Aware'' learning paradigm with GRPO.

However, as stated in previous works~\cite{lin2023mitigating,cheng2025hair}, even though these methods did mitigate the tax on model general performance, a slight drawback still exists. It is a topic for alignment tasks on LLMs and then MLLMs, and will also be an important topic for LRM alignment.
\subsection{Backdoor} 
\label{backdoor}
Backdoor attacks aim at negatively modifying model behavior when faced with pre-defined triggers while functioning normally for benign inputs~\cite{li2022backdoor}. Previously, it was classified as one type of poisoning attacks, where attackers curated a small backdoor dataset composed of triggered inputs and target abnormal outputs, and injected the backdoor behavior through fine-tuning~\cite{li2024backdoorllm,guan2022few,guan2024backdoor}. For large language models, except for data poisoning methods~\cite{hubinger2024sleeper,xu2023instructions,shi2023badgpt}, model editing~\cite{li2024badedit} and intermediate vector steering~\cite{wang2023trojan} are also proposed to inject backdoor triggers into models~\cite{li2024backdoorllm}.
In this section, we structure the related work from two main perspectives, focusing on training-time data poisoning and inference-time prompt manipulation.

\textbf{Training-time data poisoning.}
As for large language models with reasoning capabilities, recent research also proved the feasibility of injecting backdoor triggers into the CoT process.  Jin \etal~\cite{jin2024saber} proposed SABER, which leveraged CodeBERT to find optimal positions for trigger insertion in the backdoor data curation process. Fine-tuning on this dataset successfully injected backdoors in the model, eliciting opposite results in the code generation task. Targeting the thinking length of reasoning models, BoT~\cite{zhu2025think} embedded triggers to skip the thinking process, thereby affecting the answer quality. Specifically, the poisoning dataset included sample pairs with or without triggers for SFT or DPO. After that, ShadowCoT~\cite{zhao2025shadowcot} was also proposed to attack the internal reasoning, with a well-designed three-stage fine-tuning pipeline for backdoor injection without harming the general performance. Similarly, Chua \etal~\cite{chua2025thought} noticed the potential of the fine-tuning attack, and trained a ``sleeper agent'' to elicit bad behaviors only with trigger prompts, in which the CoT appeared either innocent or misaligned. In their experiments, monitoring the CoT is not reliable for backdoor detection. 

\textbf{Inference-time prompt manipulation}.
Inference-time prompt manipulation shares a huge overlap with prompt injection attacks~\cite{yan2023backdooring,clop2024backdoored,zhao2023prompt}, which \textit{``aims to compromise the data of the target task such that the LLM-integrated application is misled to accomplish an arbitrary, attacker-chosen task''}~\cite{liu2024formalizing}. Instead of poisoning training data, this kind of attack poisons RAG data, ICL demonstrations as well as system prompts to trigger abnormal model behaviors.
Badchain~\cite{xiang2024badchain} proposed to curate backdoor examples as demonstrations in ICL to elicit target generation. Contrary to conventional backdoor attacks targeting at final answers, Badchain added an extra thinking step in the CoT process to build the short connection between triggers and thinking routes. Moreover, evaluations in BackdoorLLM~\cite{li2024backdoorllm} further discovered that large language models with stronger reasoning capabilities are more vulnerable to backdoor attacks, a finding that mirrors the results in Jailbreak attacks~\cite{wei2023jailbroken}. Guo \etal~proposed DarkMind~\cite{guo2025darkmind}, which altered model behaviors with modified instructions in the system prompt. After that, Guo \etal~\cite{guo2025system} tried multiple types of system prompts, finding that poisoned prompts with CoT or ICL could largely divert model outputs across various tasks. Under RAG settings, Song \etal~\cite{song2025chain} identified the ineffectiveness of simple knowledge editing, adding reasoning templates with erroneous knowledge into the system to camouflage reasoning models, which resembles the logic behind H-CoT~\cite{kuo2025h}. In addition, Cui \etal~\cite{cui2025process} identified that inputting the thinking process with prompts into DeepSeek-R1 would prevent the model from generating a final answer, by which they designed a token-efficient prompt injection attack to trigger abnormal generation cessation and compressing the required number of tokens to about 2000~\cite{cui2025token}. Following work by Cui \etal~\cite{cui2025practical} further reduced the required injection tokens to 109. 

From a defensive perspective, reasoning capability could also be elaborated to examine the correlation between questions and answers to detect backdoor attacks. Li \etal~\cite{li2024chain} proposed Chain-of-Scrutiny (CoS) to analyze whether the model generation directly answers the prompts. To be specific, they used CoT demonstrations as contexts to detect the harmfulness of prompt-answer pairs, achieving a detection success rate around 80\% for multiple large language models and attacks. Marinelli \etal~\cite{marinelli2025harnessing} proposed to identify prompt manipulations through the number of reasoning steps: if the prompt is injected with extra tasks, the step to follow instructions should be larger than expected. Similarly, Jin \etal~proposed GUARD~\cite{jin2025guard}, encompassing a judge agent and a repair agent for backdoored CoT detection as well as modification in code generation tasks.
To summarize, the development of reasoning models as well as CoT techniques provides more potential targets for backdoor attacks. Except for outputting target harmful strings, new backdoor attacks could force models to deviate from the proper thinking process, or directly interrupt the reasoning phase from fine-tuning or prompting, exposing higher risks of cutting-edge models than less capable models.
\section{Robustness}
\label{robustness}
According to Braiek \etal~\cite{braiek2025machine}, ``\textit{model robustness denotes the capacity of a model to sustain stable predictive performance in the face of variations and changes in the input data}''. Robustness has always been a crucial part of trustworthy AI, as it determines whether a model can maintain stable and reliable performance when facing various adversarial noises in real-world deployments~\cite{wang2021measure}. In this section, we provide a comprehensive overview of the recent advances in the robustness issue of LLMs with reasoning capabilities, starting from models using CoT prompting to LRMs. Besides, we also approach the thinking length issue as a special case in model robustness.  
\subsection{Robustness Improvement with Reasoning Techniques}
\label{Robustness Improvement}
Before the rapid development of LRMs, the robustness of language models at the token level was noticed and explored. Xu \etal~\cite{xu2024preemptive} found that providing a preemptive answer before reasoning contents could lead the model to generate a reasoning process that conforms to the given answer. Zhou \etal~\cite{zhou2024can} added noisy rationales in in-context demonstrations, finding that large language models are hard to generate proper reasoning content, even with self-correction techniques~\cite{huang2023large,xi2023self}. Wang \etal~\cite{wang2024rupbench} proposed RUPbench to evaluate the reasoning robustness, concluding that larger models are more resistant to perturbations. Peng \etal~\cite{peng2025stepwise} also showcased that model generations are sensitive to misleading reasoning steps. 

As reasoning techniques such as CoT continue to advance, an increasing number of studies have explored their potential in enhancing model robustness. Lam \etal~\cite{lam2025codecrash} mentioned that CoT prompting could significantly improve LLM robustness, and Wang \etal~\cite{wang2025chain} proposed Chain-of-Defensive-Thought (CoDT) to defend language models against corrupted reference in in-context prompts. Yan \etal~\cite{yan2025recitation} found that few-shot in-context learning with modified problems could increase the accuracy, but it still cannot fully counteract the perturbation of adversarial inputs. Besides, using original problems for in-context learning may cause inappropriate memorization~\cite{huang2025math}. Similar methods also include adding system prompts and self-reflection mechanisms~\cite{wang2025assessing}. Zaremba \etal~\cite{zaremba2025trading} mentioned that test-time scaling is helpful for model robustness under some settings. To improve model robustness with external signals, Yang \etal~\cite{yang2025any} constructed training data from model distillation to train a Reasoning-based Bias Detector (RBD) for bias mitigation. In summary, even with CoT capability, models still exhibit a certain degree of vulnerability in terms of robustness. Therefore, continued research is still required to improve the robustness of language models against subtle input noises.

\subsection{Robustness of Reasoning Models}
\label{Robustness of Reasoning Models}
In terms of LRMs, the robustness against input noise is also examined, especially under the Math tasks. 
Huang \etal~\cite{huang2025math} proposed MATH-Perturb to evaluate the model's Math performance under hard perturbations, where original solutions do not apply anymore.  
Mu \etal~\cite{mu2025closer} came up with the RealGuardrails dataset to evaluate the system prompt robustness, finding obvious but uneven robustness gains in reasoning models than non-reasoning counterparts. Rajeev \etal~\cite{rajeev2025cats} proposed CatAttack, which appended unrelated trivia or misleading questions generated from PAIR~\cite{chao2023jailbreaking}, such as ``Could the answer possibly be around 175'', or ``Interesting fact: cats sleep for most of their lives'', to mislead the model. 
Yu \etal~\cite{yu2025benchmarking} introduced the Math-Robustness Benchmark (Math-RoB) to evaluate the mathematical reasoning capabilities, including adversarial noises like changing operator symbols, replacing operator symbols with Greek letters, or removing key data in the prompts. Similarly, Yan \etal~\cite{yan2025recitation} proposed RoR-bench with altered Math problems to test the robustness of reasoning models. It is found that simply modifying numbers in problems would cause an obvious degradation in reasoning performance, indicating potential memorization issues in model training. Besides, the evaluation also disclosed an obvious vulnerability for unanswerable questions, which is consistent with the evaluation results in AbstentionBench~\cite{kirichenko2025abstentionbench}. 
Wang \etal~\cite{wang2025polymath} proposed PolyMath, evaluated mathematical reasoning with multilingual contexts, and uncovered fluctuating performance on different languages. Zhu \etal~\cite{zhu2025reasoning1} mentioned that after reasoning models provide correct answers, adding a simple negation prompt to doubt the answer could mislead the second thinking process, causing an obvious accuracy drop on related benchmarks~\cite{yue2024mmmu,lu2023mathvista,wang2024charxiv}. The confidence problem was also mentioned by previous works\cite{zhang2024understanding,huang2023large}, indicating that for both reasoning and non-reasoning models, self-correction prompts expressing distrust in model outputs could hugely influence model rationales and final decisions, both positively and negatively. 
In addition, Li \etal~\cite{li2025frustratingly} introduced M-Attack to optimize transferable adversarial images. After pushing the embedding of a clean image towards another real image containing distracting semantics through feature matching and model ensembling, the perturbed adversarial image could successfully attack cutting-edge models such as GPT-4.5, 4o, or o1~\cite{achiam2023gpt}, inducing wrong image descriptions or hallucinations. Experiments demonstrated that even with reasoning capability, OpenAI o1 still struggled to distinguish noise from real images. 

The vulnerability to input perturbation is also discovered in the code generation domain. CodeCrash~\cite{lam2025codecrash} proposed to evaluate the code generation robustness with noisy requests, including garbage codes, renamed entities (which resembles altering numbers in Math problems), misleading print statements or hint comments, etc. While the results demonstrated superior performance compared to non-reasoning counterparts, they also revealed significant vulnerabilities under certain perturbations. Roh \etal~\cite{roh2025break} identified the robustness vulnerability against the Chain-of-Code Collapse (CoCC) framework, in which the original prompt was wrapped with a narrative tone, making it a story or an adventure. Moreover, Wang \etal~\cite{wang2025assessing} evaluated the judging bias of large reasoning models, finding that even if LRMs perform better than LLMs on objective domains, they are still vulnerable to biases such as choice position, authority, or major beliefs distractions.

\subsection{Overthinking and Underthinking}
\label{overthink}
Overthinking is an emerging problem in reasoning models, referring to the phenomenon where ``LLMs generate excessively detailed or unnecessarily elaborate reasoning steps, ultimately reducing their problem-solving efficiency''.~\cite{sui2025stop,chen2024not} From the trustworthy perspective, instead of efficiency, we focus more on situations where \textit{models are trapped in repeating reasoning trajectories in a non-stop manner, and may output wrong answers in the end}. Conversely, underthinking refers to the situation where LLMs generate abnormally short reasoning or completely skip the reasoning process, even if the thinking behavior is necessary or required. Along the same lines as before, modifications to the Math questions could trigger redundant reflections, resulting in overthinking~\cite{ma2024large,hashemi2025dnr}. Generally, such overthinking vulnerability mainly occurs when faced with unanswerable questions or erroneous premises. Some researchers~\cite{he2025can, fan2025missing} found that the overconfidence, or reliance, on input prompts forces reasoning models to try numerous thoughts while failing to doubt the validity of prompts. Wang \etal~\cite{wang2025thoughts} attributed the redundant thinking tokens with unsatisfying accuracy to frequent thought switching. Su \etal~\cite{su2025between} studied the relationship between reasoning length and answer correctness, finding that models failed to allocate proper reasoning length to questions with different levels of difficulty. Dang \etal~\cite{dang2025internal} also proposed that ``internal bias'' is strongly related to the overthinking behavior. When the internal bias contradicts the conclusion after stepwise thoughts, the model will trigger reflections. 

To deliberately elicit overthinking behavior, the earliest work is Overthink~\cite{kumar2025overthink}, which added unrelated or adversarial context to the prompts to obfuscate model reasoning. Similar attacks are also proposed in multiple literatures~\cite{zaremba2025trading,lam2025codecrash,si2025excessive}, in which Si \etal~\cite{si2025excessive} introduced a GCG-style~\cite{zou2023universal} optimization pipeline to generate adversarial overthinking triggers. Under agentic environments, Cuadron \etal~\cite{cuadron2025danger} identified the reasoning-action dilemma, and categorized three patterns of overthinking where the model prefers overly reasoning to interacting with environments. To mitigate overthinking, there are a lot of works heading towards efficient reasoning~\cite{feng2025efficient, qu2025survey, sui2025stop}, including but not limited to prompt-driven methods~\cite{xu2025chain,ma2024large,ma2025cot}, training-based methods~\cite{yang2025think,wang2025adaptive,xia2025tokenskip,munkhbat2025self,yu2024distilling}, inference-based methods~\cite{huang2025efficient,jiang2025safechain,wang2025sampling,yu2025think}, representation-based methods~\cite{huang2025mitigating,cyberey2025steering}, etc. 

Underthinking, compared to overthinking, constitutes a more pure robustness topic. Input manipulation could also trigger underthinking~\cite{cui2025process,zaremba2025trading}. For example, padding original prompts with compromised thoughts could make DeepSeek-R1 stop further reasoning~\cite{cui2025process}. A few researchers also mentioned that the think-less attack could limit the test-time compute of reasoning models, making them more vulnerable to attacks~\cite{zaremba2025trading,li2025output,zhao2025trade}. Sun \etal~\cite{sun2025thinkedit} located a subset of attention layers in the model weight, proposed ThinkEdit to remove the short thinking direction. In general, current reasoning models lack sufficient robustness against manipulations of thinking length. To advance both robustness and efficiency, further research is needed to investigate the underlying causes of overthinking and underthinking behaviors, as well as to develop effective mitigation strategies.

\section{Fairness}
\label{fairness}
Fairness focuses on the ethical principles language models possess, especially whether language models react equally to different users or groups, including genders, LGBTQ+ communities, races, language, and political orientations without preference or discrimination~\cite{huang2024position}. As stated in previous literature~\cite{li2023survey,gallegos2024bias}, the bias may emerge, or be exaggerated, from imperfect training data, the choice of optimization, evaluation metrics, and the deployment phase. In this section, instead of thoroughly reviewing fairness evaluation and debiasing methods in LLMs, we simply limit our scope to recent fairness studies with regard to the reasoning capability.

Lin \etal~\cite{linassessing} identified the dialect bias of multiple cutting-edge language models with the experiments of paraphrasing standard English queries into African American Vernacular English (AAVE). CoT prompting is helpful to mitigate this bias, but it is unable to fully solve such a discrepancy, just like the results on robustness~\cite{yan2025recitation}. Cheng \etal~\cite{cheng2025detection} also mentioned that CoT prompting could guide the model to correctly classify gender biases. Kamruzzaman \etal~\cite{kamruzzaman2024prompting} evaluated multiple prompting strategies for social bias reduction, finding that system 2 prompts with a human persona could reduce stereotypical judgments. However, another line of work stated that under persona-assigned tasks, CoT prompts are not sufficient to mitigate human-like motivated reasoning~\cite{dash2025persona,gupta2023bias}. For bias detection, Fan \etal~\cite{fan2025biasguard} proposed BiasGuard to identify potential discrimination with internal reasoning capability. The training included an SFT stage followed by a DPO stage, which resembles the development of guardrail models in Section~\ref{guardrail model}. Cantini \etal~\cite{cantini2025reasoning} exploited the CLEAR-Bias benchmark~\cite{cantini2025benchmarking} for LRMs, concluding that models with explicit reasoning are more vulnerable in terms of bias, even though they are slightly safer than LLMs with CoT prompting. Overall, current researches underscore that current CoT and reasoning techniques have yet to bridge the gap toward achieving authentic fairness in models, and the fairness may still depend on the quality and distribution of training data.

\section{Privacy}
\label{privacy}
Privacy is always an important concern in the development of ML algorithms. Dating back to the CNN era, there has been a lot of work studying the potential to infer or steal the model and training data~\cite{wang2024towards,shokri2017membership,zhou2024model}, as well as their corresponding defenses~\cite{jiang2025shadow,guan2022you,guan2025sample}. In recent years, we have also witnessed some inference-time attacks to extract personally identifiable information (PII), private retrieval-augmented generation (RAG) documents, or model weights when interacting with large language models~\cite{jiang2024rag,wang2025silent,carlini2024stealing}. As reasoning capabilities become more advanced, the risk of intentionally disclosing private information through user input increases. In this section, we elaborate on related research from the model and prompt perspectives, specifically whether the privacy issue originates from model training data or external prompts.
\subsection{Model-related Privacy} \label{Model-related privacy}
\textbf{Unlearning.} Large language model unlearning aims to erase copyrighted contents, remove harmful generations, protect data privacy, etc.~\cite{yao2024large}. Following previous work on unlearning method evaluation~\cite{maini2024tofu}, Yoon \etal~proposed R-TOFU~\cite{yoon2025r} to evaluate a few baseline unlearning methods with different strategies on reasoning models, concluding that unlearning only the final result is insufficient to forget the specific information. Similar conclusions are also drawn by Wang \etal~\cite{wang2025reasoning}, and they proposed R$^2$MU that mapped the intermediate features of reasoning steps to randomly scaled vectors for an improvement.  Both works highlighted the forgetting of CoT contents, providing a feasible direction for future attempts. From the other side, attacks against unlearning were also developed to recover erased data, which discloses the vulnerability of unlearning methods~\cite{lynch2024eight,hu2024jogging}. For reasoning models, Sinha \etal~\cite{sinha2025step} proposed SLEEK to elicit unlearned information in a multi-turn manner. Aimed at finding residual traces related to the unlearning target, SLEEK first generates queries targeting each object or fact with CoT techniques, and then prompts the model in multi-turn interactions to test whether any residual details remain in the response. This method achieved an ASR above 50\% on Harry Potter facts against chat models, suggesting that full mitigation of memorized content may not yet be guaranteed.  
 
\textbf{Model IP protection.} To prevent the model from copying or stealing, researchers have proposed numerous active or passive defense methods to protect the released models as well as their valuable training datasets, including fingerprinting, watermarking, unlearnable techniques, etc~\cite{guan2022you,peng2022fingerprinting,wang2021fingerprinting,guan2024sample,fernandez2023stable,li2021modeldiff,fu2022robust,sandoval2022autoregressive,huang2021unlearnable}. In terms of large language models, representing work~\cite{kirchenbauer2023watermark} promoted the sampling possibility of a fraction of tokens in the vocabulary, so that the watermark is printed as the ratio of selected tokens versus the rest tokens in the generated texts. After that, the development of CoT prompting provides more chances to model IP protection. ImF~\cite{wanli2025imf} embedded the fingerprint\footnote{``Fingerprint'' originally refers to inherent, verifiable model features (e.g., weights or activations), while ``watermark'' denotes externally embedded signals. In this context, the distinction is blurred, and both terms refer to watermarks.} into pre-defined CoT prompt-answer pairs. CoTSRF~\cite{ren2025cotsrf} trained an extractor to capture the feature of CoT-prompt conditioned reasoning steps, and calculate the Kullback-Leibler divergence (KL divergence) with the suspect model in the verification phase. To enable RAG data protection, Guo \etal~\cite{guo2025towards} imprinted watermarks into knowledge text, so that the model would generate a specific CoT trace with correct answers when faced with verification questions, enabling an effective and harmless copyright protection. Aside from watermarking methods, Savani \etal~\cite{savani2025antidistillation} proposed ``antidistillation sampling'' to prevent model-generated contents from being trained. When decoding, the method modified the output logits to maximize the potential training loss while keeping the correctness of the outputs. Experiments on Math datasets~\cite{hendrycks2021measuring,cobbe2021gsm8k} demonstrated the feasibility of this approach: antidistillation sampling achieved accuracy comparable to temperature sampling, while student models suffered a notable performance drop of approximately 30\% on GSM8K~\cite{cobbe2021gsm8k}. Together, these techniques provide a basis for ongoing efforts to develop reliable and practical IP protection mechanisms.
\subsection{Prompt-related Privacy}
\label{Prompt-related Privacy}
With the fast progress in large language models, the ability to infer private information from input prompts also gets stronger. Staab \etal~\cite{staab2023beyond} was the first to research the privacy inference attack in large language models, drawing the result that LLMs are capable of inferring various personal attributes beyond memorization. T{\"o}mek{\c{c}}e \etal~\cite{tomekcce2024private} tested the inferring capability in the vision domain, demonstrated that the inference accuracy is positively related to the general capabilities of the models, and underscored the necessity of privacy protection methods. After the advent of CoT techniques, Green \etal~\cite{green2025leaky} evaluated the privacy leakage of reasoning models, claiming that the reasoning traces could disclose more private information. While additional reasoning steps may lead to more cautious final answers, they can inadvertently reveal sensitive data during intermediate generation, aligning with the findings discussed in Section~\ref{Model-related privacy}~\cite{yoon2025r,wang2025reasoning}. Luo \etal~\cite{luo2025doxing} curated a benchmark to evaluate the attribute inference attack of vision-language models, finding that multi-model large reasoning models have strong capabilities of inferring geological information in input images, while seldom limiting this feature. Based on these findings, they proposed GeoMiner to trigger location-related attribute inference attacks. Such a method achieved higher performance than simple CoT methods, urging the need for protection. 

With a similar logic to develop defense methods against Jailbreak in Section~\ref{safety}, the defense of attribute inference attacks also includes prompting, post-training, and guardrails. However, experiments by Staab \etal~\cite{staab2023beyond} showed limited privacy gains from client-side anonymization or alignment. Such a vulnerability is also supported by Luo \etal~\cite{luo2025doxing}, stating that current SoTA guardrails cannot identify such an attack, and padding system prompts with warnings on location leakage could sacrifice the general performance. To summarize, more future works are needed to defend against this escalating threat.

\section{Future Research Directions}
\textbf{Standard measurements of faithfulness.}
A wide range of methods have been proposed to evaluate reasoning faithfulness, but none are comprehensive, often leading to divergent or even contradictory conclusions.
For example, some studies argue that larger models exhibit greater faithfulness~\cite{bao2024likely,xiong2025measuring}, while others contend that they are less faithful~\cite{benthamchain}.
This inconsistency highlights the need for more robust and standardized evaluation protocols that can fairly assess reasoning faithfulness across models.

In addition, some existing methods for evaluating faithfulness may conflict with other aspects of the performance of large models.
For example, one common evaluation technique involves CoT intervention methods.
These approaches test how perturbations to intermediate reasoning steps affect final answers.
Empirical findings suggest that stronger models can answer correctly even with the perturbed CoT, implying that their outputs may rely less on explicit reasoning traces and more on internalized knowledge.
From this, one might conclude that stronger models are less faithful, as their outputs do not depend transparently on the provided reasoning paths.
However, such a conclusion conflicts with robustness. Therefore, eliminating the evaluation bias caused by model performance remains a critical open problem.

\textbf{More analyses on safety mechanism.} After reviewing attack and defense methods in Section~\ref{safety}, we call for more studies on the safety mechanism. Previous works demonstrated the feasibility of post-training methods with an extra safety-related CoT dataset. However, heuristic insights into effective dataset construction remain limited, leaving many details, such as prompts for CoT distillation, data ratios across different sources, and the necessity of cold-start SFT, reliant on manual tuning and empirical intuition. Moreover, in terms of the safety tax, the empirical understanding of how reinforcement learning contributes to safety and alignment remains limited. For instance, it remains challenging to disentangle the extent to which performance gains stem from the learning algorithm itself (e.g., GRPO over DPO) versus the influence of higher-quality data, such as well-curated CoT examples. Some progress has been made in understanding the role of SFT versus RL~\cite{chu2025sft,chen2025sft}, and we encourage future work to further investigate the role and limits of RL in this context. 

\textbf{More fine-grained benchmarks}.
As language models continue to grow in capability, there is an increasing need for safety evaluation benchmarks that can effectively reflect their evolving behaviors.
Current safety evaluation benchmarks are primarily based on a narrow set of related attack methods~\cite{zou2023universal,rottger2023xstest,luo2024jailbreakv}, resulting in significant homogenization of data distribution. As a consequence, metrics such as ASR often exhibit extreme values. Besides, due to the inherent properties of generative models, the outputs may be sensitive to variations in temperature settings and prompt formulations, thereby impacting the reproducibility of experimental results. In this regard, we call for new benchmarks that are more discriminative, detailed, and robust. In addition, compared with the number of benchmarks in safety and robustness, evaluations on privacy inference and fairness have comparatively received less emphasis. These areas would benefit from increased focus in future work if more evaluations with comprehensive coverage, clear definitions, and diverse testing samples are developed.
\section{Conclusion}
In conclusion, this survey summarizes recent literature concerning trustworthiness in reasoning capabilities, providing a comprehensive overview with a clear taxonomy. With efforts on each topic, we describe the development of novel methods, point out prevailing conclusions, and highlight the related analysis as well as future opportunities. We believe that our comprehensive survey and structured taxonomy could offer a foundation for future research in building safer, more reliable models with reasoning capabilities.

\end{CJK*}
\newpage
{
\bibliographystyle{define}
\bibliography{citation}

\begin{thebibliography}{100}

\bibitem{jiang2025safechain}
Fengqing Jiang, Zhangchen Xu, Yuetai Li, Luyao Niu, Zhen Xiang, Bo~Li, Bill~Yuchen Lin, and Radha Poovendran.
\newblock Safechain: Safety of language models with long chain-of-thought reasoning capabilities.
\newblock {\em arXiv preprint arXiv:2502.12025}, 2025.

\bibitem{lu2025does}
Chengda Lu, Xiaoyu Fan, Yu~Huang, Rongwu Xu, Jijie Li, and Wei Xu.
\newblock Does Chain-of-Thought Reasoning Really Reduce Harmfulness from Jailbreaking?
\newblock {\em arXiv preprint arXiv:2505.17650}, 2025.

\bibitem{ying2025towards}
Zonghao Ying, Guangyi Zheng, Yongxin Huang, Deyue Zhang, Wenxin Zhang, Quanchen Zou, Aishan Liu, Xianglong Liu, and Dacheng Tao.
\newblock Towards understanding the safety boundaries of deepseek models: Evaluation and findings.
\newblock {\em arXiv preprint arXiv:2503.15092}, 2025.

\bibitem{wang2025comprehensive}
Kun Wang, Guibin Zhang, Zhenhong Zhou, Jiahao Wu, Miao Yu, Shiqian Zhao, Chenlong Yin, Jinhu Fu, Yibo Yan, Hanjun Luo, et~al.
\newblock A comprehensive survey in llm (-agent) full stack safety: Data, training and deployment.
\newblock {\em arXiv preprint arXiv:2504.15585}, 2025.

\bibitem{dong2024attacks}
Zhichen Dong, Zhanhui Zhou, Chao Yang, Jing Shao, and Yu~Qiao.
\newblock Attacks, defenses and evaluations for llm conversation safety: A survey.
\newblock In {\em Proc. NAACL}, 2024.

\bibitem{shi2024large}
Dan Shi, Tianhao Shen, Yufei Huang, Zhigen Li, Yongqi Leng, Renren Jin, Chuang Liu, Xinwei Wu, Zishan Guo, Linhao Yu, et~al.
\newblock Large language model safety: A holistic survey.
\newblock {\em arXiv preprint arXiv:2412.17686}, 2024.

\bibitem{chen2025towards}
Qiguang Chen, Libo Qin, Jinhao Liu, Dengyun Peng, Jiannan Guan, Peng Wang, Mengkang Hu, Yuhang Zhou, Te~Gao, and Wangxiang Che.
\newblock Towards reasoning era: A survey of long chain-of-thought for reasoning large language models.
\newblock {\em arXiv preprint arXiv:2503.09567}, 2025.

\bibitem{xu2025towards}
Fengli Xu, Qianyue Hao, Zefang Zong, Jingwei Wang, Yunke Zhang, Jingyi Wang, Xiaochong Lan, Jiahui Gong, Tianjian Ouyang, Fanjin Meng, et~al.
\newblock Towards Large Reasoning Models: A Survey of Reinforced Reasoning with Large Language Models.
\newblock {\em arXiv preprint arXiv:2501.09686}, 2025.

\bibitem{qu2025survey}
Xiaoye Qu, Yafu Li, Zhaochen Su, Weigao Sun, Jianhao Yan, Dongrui Liu, Ganqu Cui, Daizong Liu, Shuxian Liang, Junxian He, et~al.
\newblock A survey of efficient reasoning for large reasoning models: Language, multimodality, and beyond.
\newblock {\em arXiv preprint arXiv:2503.21614}, 2025.

\bibitem{sui2025stop}
Yang Sui, Yu-Neng Chuang, Guanchu Wang, Jiamu Zhang, Tianyi Zhang, Jiayi Yuan, Hongyi Liu, Andrew Wen, Shaochen Zhong, Hanjie Chen, et~al.
\newblock Stop overthinking: A survey on efficient reasoning for large language models.
\newblock {\em arXiv preprint arXiv:2503.16419}, 2025.

\bibitem{feng2025efficient}
Sicheng Feng, Gongfan Fang, Xinyin Ma, and Xinchao Wang.
\newblock Efficient reasoning models: A survey.
\newblock {\em arXiv preprint arXiv:2504.10903}, 2025.

\bibitem{wang2025safety}
Haoyu Wang, Zeyu Qin, Li~Shen, Xueqian Wang, Dacheng Tao, and Minhao Cheng.
\newblock Safety Reasoning with Guidelines.
\newblock In {\em Proc. ICML}, 2025.

\bibitem{achiam2023gpt}
Josh Achiam, Steven Adler, Sandhini Agarwal, Lama Ahmad, Ilge Akkaya, Florencia~Leoni Aleman, Diogo Almeida, Janko Altenschmidt, Sam Altman, Shyamal Anadkat, et~al.
\newblock Gpt-4 technical report.
\newblock {\em arXiv preprint arXiv:2303.08774}, 2023.

\bibitem{TheC3}
Anthropic.
\newblock The Claude 3 Model Family: Opus, Sonnet, Haiku, 2024.

\bibitem{guo2025deepseek}
Daya Guo, Dejian Yang, Haowei Zhang, Junxiao Song, Ruoyu Zhang, Runxin Xu, Qihao Zhu, Shirong Ma, Peiyi Wang, Xiao Bi, et~al.
\newblock Deepseek-r1: Incentivizing reasoning capability in llms via reinforcement learning.
\newblock {\em arXiv preprint arXiv:2501.12948}, 2025.

\bibitem{wei2022chain}
Jason Wei, Xuezhi Wang, Dale Schuurmans, Maarten Bosma, Fei Xia, Ed~Chi, Quoc~V Le, Denny Zhou, et~al.
\newblock Chain-of-thought prompting elicits reasoning in large language models.
\newblock In {\em Proc. NeurIPS}, 2022.

\bibitem{kojima2022large}
Takeshi Kojima, Shixiang~Shane Gu, Machel Reid, Yutaka Matsuo, and Yusuke Iwasawa.
\newblock Large language models are zero-shot reasoners.
\newblock In {\em Proc. NeurIPS}, 2022.

\bibitem{brown2020language}
Tom Brown, Benjamin Mann, Nick Ryder, Melanie Subbiah, Jared~D Kaplan, Prafulla Dhariwal, Arvind Neelakantan, Pranav Shyam, Girish Sastry, Amanda Askell, et~al.
\newblock Language models are few-shot learners.
\newblock In {\em Proc. NeurIPS}, 2020.

\bibitem{li2024chain}
Xi~Li, Yusen Zhang, Renze Lou, Chen Wu, and Jiaqi Wang.
\newblock Chain-of-scrutiny: Detecting backdoor attacks for large language models.
\newblock {\em arXiv preprint arXiv:2406.05948}, 2024.

\bibitem{jaech2024openai}
Aaron Jaech, Adam Kalai, Adam Lerer, Adam Richardson, Ahmed El-Kishky, Aiden Low, Alec Helyar, Aleksander Madry, Alex Beutel, Alex Carney, et~al.
\newblock Openai o1 system card.
\newblock {\em arXiv preprint arXiv:2412.16720}, 2024.

\bibitem{wang2024openr}
Jun Wang, Meng Fang, Ziyu Wan, Muning Wen, Jiachen Zhu, Anjie Liu, Ziqin Gong, Yan Song, Lei Chen, Lionel~M Ni, et~al.
\newblock Openr: An open source framework for advanced reasoning with large language models.
\newblock {\em arXiv preprint arXiv:2410.09671}, 2024.

\bibitem{qin2024o1}
Yiwei Qin, Xuefeng Li, Haoyang Zou, Yixiu Liu, Shijie Xia, Zhen Huang, Yixin Ye, Weizhe Yuan, Hector Liu, Yuanzhi Li, et~al.
\newblock O1 Replication Journey: A Strategic Progress Report--Part 1.
\newblock {\em arXiv preprint arXiv:2410.18982}, 2024.

\bibitem{huang2024o1}
Zhen Huang, Haoyang Zou, Xuefeng Li, Yixiu Liu, Yuxiang Zheng, Ethan Chern, Shijie Xia, Yiwei Qin, Weizhe Yuan, and Pengfei Liu.
\newblock O1 Replication Journey--Part 2: Surpassing O1-preview through Simple Distillation, Big Progress or Bitter Lesson?
\newblock {\em arXiv preprint arXiv:2411.16489}, 2024.

\bibitem{huang2025o1}
Zhongzhen Huang, Gui Geng, Shengyi Hua, Zhen Huang, Haoyang Zou, Shaoting Zhang, Pengfei Liu, and Xiaofan Zhang.
\newblock O1 Replication Journey--Part 3: Inference-time Scaling for Medical Reasoning.
\newblock {\em arXiv preprint arXiv:2501.06458}, 2025.

\bibitem{zhang2024llama}
Di~Zhang, Jianbo Wu, Jingdi Lei, Tong Che, Jiatong Li, Tong Xie, Xiaoshui Huang, Shufei Zhang, Marco Pavone, Yuqiang Li, et~al.
\newblock Llama-berry: Pairwise optimization for o1-like olympiad-level mathematical reasoning.
\newblock {\em arXiv preprint arXiv:2410.02884}, 2024.

\bibitem{browne2012survey}
Cameron~B Browne, Edward Powley, Daniel Whitehouse, Simon~M Lucas, Peter~I Cowling, Philipp Rohlfshagen, Stephen Tavener, Diego Perez, Spyridon Samothrakis, and Simon Colton.
\newblock A survey of monte carlo tree search methods.
\newblock {\em IEEE Transactions on Computational Intelligence and AI in games}, pages 1--43, 2012.

\bibitem{cobbe2021gsm8k}
Karl Cobbe, Vineet Kosaraju, Mohammad Bavarian, Mark Chen, Heewoo Jun, Lukasz Kaiser, Matthias Plappert, Jerry Tworek, Jacob Hilton, Reiichiro Nakano, Christopher Hesse, and John Schulman.
\newblock Training Verifiers to Solve Math Word Problems.
\newblock {\em arXiv preprint arXiv:2110.14168}, 2021.

\bibitem{hendrycks2021measuring}
Dan Hendrycks, Collin Burns, Saurav Kadavath, Akul Arora, Steven Basart, Eric Tang, Dawn Song, and Jacob Steinhardt.
\newblock Measuring mathematical problem solving with the math dataset.
\newblock In {\em Proc. NeurIPS D\&B Track}, 2021.

\bibitem{liao2024mario}
Minpeng Liao, Wei Luo, Chengxi Li, Jing Wu, and Kai Fan.
\newblock MARIO: MAth Reasoning with code Interpreter Output--A Reproducible Pipeline.
\newblock In {\em Findings of Proc. ACL}, page 905–924, 2024.

\bibitem{rafailov2023direct}
Rafael Rafailov, Archit Sharma, Eric Mitchell, Christopher~D Manning, Stefano Ermon, and Chelsea Finn.
\newblock Direct preference optimization: Your language model is secretly a reward model.
\newblock In {\em Proc. NeurIPS}, 2023.

\bibitem{schulman2017proximal}
John Schulman, Filip Wolski, Prafulla Dhariwal, Alec Radford, and Oleg Klimov.
\newblock Proximal policy optimization algorithms.
\newblock {\em arXiv preprint arXiv:1707.06347}, 2017.

\bibitem{shao2024deepseekmath}
Zhihong Shao, Peiyi Wang, Qihao Zhu, Runxin Xu, Junxiao Song, Xiao Bi, Haowei Zhang, Mingchuan Zhang, YK~Li, Y~Wu, et~al.
\newblock Deepseekmath: Pushing the limits of mathematical reasoning in open language models.
\newblock {\em arXiv preprint arXiv:2402.03300}, 2024.

\bibitem{geiping2025scaling}
Jonas Geiping, Sean McLeish, Neel Jain, John Kirchenbauer, Siddharth Singh, Brian~R Bartoldson, Bhavya Kailkhura, Abhinav Bhatele, and Tom Goldstein.
\newblock Scaling up Test-Time Compute with Latent Reasoning: A Recurrent Depth Approach.
\newblock {\em arXiv preprint arXiv:2502.05171}, 2025.

\bibitem{hao2024training}
Shibo Hao, Sainbayar Sukhbaatar, DiJia Su, Xian Li, Zhiting Hu, Jason Weston, and Yuandong Tian.
\newblock Training large language models to reason in a continuous latent space.
\newblock In {\em ICLR Workshop on LLM Reason and Plan}, 2024.

\bibitem{liu2024deepseek}
Aixin Liu, Bei Feng, Bing Xue, Bingxuan Wang, Bochao Wu, Chengda Lu, Chenggang Zhao, Chengqi Deng, Chenyu Zhang, Chong Ruan, et~al.
\newblock Deepseek-v3 technical report.
\newblock {\em arXiv preprint arXiv:2412.19437}, 2024.

\bibitem{qwen2024qwen2}
A~Yang Qwen, Baosong Yang, B~Zhang, B~Hui, B~Zheng, B~Yu, Chengpeng Li, D~Liu, F~Huang, H~Wei, et~al.
\newblock Qwen2.5 technical report.
\newblock {\em arXiv preprint}, 2024.

\bibitem{dubey2024llama}
Abhimanyu Dubey, Abhinav Jauhri, Abhinav Pandey, Abhishek Kadian, Ahmad Al-Dahle, Aiesha Letman, Akhil Mathur, Alan Schelten, Amy Yang, Angela Fan, et~al.
\newblock The llama 3 herd of models.
\newblock {\em arXiv e-prints}, pages arXiv--2407, 2024.

\bibitem{uesato2022solving}
Jonathan Uesato, Nate Kushman, Ramana Kumar, Francis Song, Noah Siegel, Lisa Wang, Antonia Creswell, Geoffrey Irving, and Irina Higgins.
\newblock Solving math word problems with process-and outcome-based feedback.
\newblock {\em arXiv preprint arXiv:2211.14275}, 2022.

\bibitem{zelikman2024star}
Eric Zelikman, Yuhuai Wu, Jesse Mu, and Noah~D Goodman.
\newblock Star: Self-taught reasoner bootstrapping reasoning with reasoning.
\newblock In {\em Proc. NeurIPS}, volume 1126, 2024.

\bibitem{lightman2023let}
Hunter Lightman, Vineet Kosaraju, Yuri Burda, Harrison Edwards, Bowen Baker, Teddy Lee, Jan Leike, John Schulman, Ilya Sutskever, and Karl Cobbe.
\newblock Let's verify step by step.
\newblock In {\em Proc. ICLR}, 2023.

\bibitem{lambert2024t}
Nathan Lambert, Jacob Morrison, Valentina Pyatkin, Shengyi Huang, Hamish Ivison, Faeze Brahman, Lester James~V Miranda, Alisa Liu, Nouha Dziri, Shane Lyu, et~al.
\newblock T$\backslash$" ulu 3: Pushing frontiers in open language model post-training.
\newblock {\em arXiv preprint arXiv:2411.15124}, 2024.

\bibitem{mroueh2025reinforcement}
Youssef Mroueh.
\newblock Reinforcement Learning with Verifiable Rewards: GRPO's Effective Loss, Dynamics, and Success Amplification.
\newblock {\em arXiv preprint arXiv:2503.06639}, 2025.

\bibitem{li2025perception}
Yunxin Li, Zhenyu Liu, Zitao Li, Xuanyu Zhang, Zhenran Xu, Xinyu Chen, Haoyuan Shi, Shenyuan Jiang, Xintong Wang, Jifang Wang, et~al.
\newblock Perception, reason, think, and plan: A survey on large multimodal reasoning models.
\newblock {\em arXiv preprint arXiv:2505.04921}, 2025.

\bibitem{wang2025multimodal}
Yaoting Wang, Shengqiong Wu, Yuecheng Zhang, Shuicheng Yan, Ziwei Liu, Jiebo Luo, and Hao Fei.
\newblock Multimodal chain-of-thought reasoning: A comprehensive survey.
\newblock {\em arXiv preprint arXiv:2503.12605}, 2025.

\bibitem{zhang2023multimodal}
Zhuosheng Zhang, Aston Zhang, Mu~Li, Hai Zhao, George Karypis, and Alex Smola.
\newblock Multimodal chain-of-thought reasoning in language models.
\newblock {\em Transactions on Machine Learning Research}, 2024.

\bibitem{fei2024video}
Hao Fei, Shengqiong Wu, Wei Ji, Hanwang Zhang, Meishan Zhang, Mong-Li Lee, and Wynne Hsu.
\newblock Video-of-thought: Step-by-step video reasoning from perception to cognition.
\newblock In {\em Proc. ICML}, 2024.

\bibitem{zheng2024thinking}
Haojie Zheng, Tianyang Xu, Hanchi Sun, Shu Pu, Ruoxi Chen, and Lichao Sun.
\newblock Thinking before looking: Improving multimodal llm reasoning via mitigating visual hallucination.
\newblock {\em arXiv preprint arXiv:2411.12591}, 2024.

\bibitem{xu2024llava}
Guowei Xu, Peng Jin, Li~Hao, Yibing Song, Lichao Sun, and Li~Yuan.
\newblock Llava-o1: Let vision language models reason step-by-step.
\newblock {\em arXiv preprint arXiv:2411.10440}, 2024.

\bibitem{thawakar2025llamav}
Omkar Thawakar, Dinura Dissanayake, Ketan More, Ritesh Thawkar, Ahmed Heakl, Noor Ahsan, Yuhao Li, Mohammed Zumri, Jean Lahoud, Rao~Muhammad Anwer, et~al.
\newblock Llamav-o1: Rethinking step-by-step visual reasoning in llms.
\newblock {\em arXiv preprint arXiv:2501.06186}, 2025.

\bibitem{xu2025redstar}
Haotian Xu, Xing Wu, Weinong Wang, Zhongzhi Li, Da~Zheng, Boyuan Chen, Yi~Hu, Shijia Kang, Jiaming Ji, Yingying Zhang, et~al.
\newblock RedStar: Does Scaling Long-CoT Data Unlock Better Slow-Reasoning Systems?
\newblock {\em arXiv preprint arXiv:2501.11284}, 2025.

\bibitem{yao2024mulberry}
Huanjin Yao, Jiaxing Huang, Wenhao Wu, Jingyi Zhang, Yibo Wang, Shunyu Liu, Yingjie Wang, Yuxin Song, Haocheng Feng, Li~Shen, et~al.
\newblock Mulberry: Empowering mllm with o1-like reasoning and reflection via collective monte carlo tree search.
\newblock {\em arXiv preprint arXiv:2412.18319}, 2024.

\bibitem{zhang2024improve}
Ruohong Zhang, Bowen Zhang, Yanghao Li, Haotian Zhang, Zhiqing Sun, Zhe Gan, Yinfei Yang, Ruoming Pang, and Yiming Yang.
\newblock Improve vision language model chain-of-thought reasoning.
\newblock {\em arXiv preprint arXiv:2410.16198}, 2024.

\bibitem{dong2025insight}
Yuhao Dong, Zuyan Liu, Hai-Long Sun, Jingkang Yang, Winston Hu, Yongming Rao, and Ziwei Liu.
\newblock Insight-v: Exploring long-chain visual reasoning with multimodal large language models.
\newblock In {\em Proc. CVPR}, pages 9062--9072, 2025.

\bibitem{guo2024mammoth}
Jarvis Guo, Tuney Zheng, Yuelin Bai, Bo~Li, Yubo Wang, King Zhu, Yizhi Li, Graham Neubig, Wenhu Chen, and Xiang Yue.
\newblock Mammoth-vl: Eliciting multimodal reasoning with instruction tuning at scale.
\newblock {\em arXiv preprint arXiv:2412.05237}, 2024.

\bibitem{sun2025mm}
Linzhuang Sun, Hao Liang, Jingxuan Wei, Bihui Yu, Tianpeng Li, Fan Yang, Zenan Zhou, and Wentao Zhang.
\newblock Mm-verify: Enhancing multimodal reasoning with chain-of-thought verification.
\newblock {\em arXiv preprint arXiv:2502.13383}, 2025.

\bibitem{cheng2025think}
Xiaoxue Cheng, Junyi Li, Wayne~Xin Zhao, and Ji-Rong Wen.
\newblock Think More, Hallucinate Less: Mitigating Hallucinations via Dual Process of Fast and Slow Thinking.
\newblock {\em arXiv preprint arXiv:2501.01306}, 2025.

\bibitem{akbar2024hallumeasure}
Shayan~Ali Akbar, Md~Mosharaf Hossain, Tess Wood, Si-Chi Chin, Erica~M Salinas, Victor Alvarez, and Erwin Cornejo.
\newblock HalluMeasure: Fine-grained hallucination measurement using chain-of-thought reasoning.
\newblock In {\em Proc. EMNLP}, pages 15020--15037, 2024.

\bibitem{eliav2025clatter}
Ron Eliav, Arie Cattan, Eran Hirsch, Shahaf Bassan, Elias Stengel-Eskin, Mohit Bansal, and Ido Dagan.
\newblock CLATTER: Comprehensive Entailment Reasoning for Hallucination Detection.
\newblock {\em arXiv preprint arXiv:2506.05243}, 2025.

\bibitem{xie2024order}
Zikai Xie.
\newblock Order Matters in Hallucination: Reasoning Order as Benchmark and Reflexive Prompting for Large-Language-Models.
\newblock {\em arXiv preprint arXiv:2408.05093}, 2024.

\bibitem{wu2025grounded}
Qiong Wu, Xiangcong Yang, Yiyi Zhou, Chenxin Fang, Baiyang Song, Xiaoshuai Sun, and Rongrong Ji.
\newblock Grounded chain-of-thought for multimodal large language models.
\newblock {\em arXiv preprint arXiv:2503.12799}, 2025.

\bibitem{jiang2024comt}
Yue Jiang, Jiawei Chen, Dingkang Yang, Mingcheng Li, Shunli Wang, Tong Wu, Ke~Li, and Lihua Zhang.
\newblock CoMT: Chain-of-Medical-Thought Reduces Hallucination in Medical Report Generation.
\newblock In {\em Proc. ICASSP}, 2025.

\bibitem{dong2025mirage}
Bowen Dong, Minheng Ni, Zitong Huang, Guanglei Yang, Wangmeng Zuo, and Lei Zhang.
\newblock MIRAGE: Assessing Hallucination in Multimodal Reasoning Chains of MLLM.
\newblock {\em arXiv preprint arXiv:2505.24238}, 2025.

\bibitem{song2025hallucination}
Linxin Song, Taiwei Shi, and Jieyu Zhao.
\newblock The Hallucination Tax of Reinforcement Finetuning.
\newblock {\em arXiv preprint arXiv:2505.13988}, 2025.

\bibitem{liu2025more}
Chengzhi Liu, Zhongxing Xu, Qingyue Wei, Juncheng Wu, James Zou, Xin~Eric Wang, Yuyin Zhou, and Sheng Liu.
\newblock More Thinking, Less Seeing? Assessing Amplified Hallucination in Multimodal Reasoning Models.
\newblock {\em arXiv preprint arXiv:2505.21523}, 2025.

\bibitem{yao2025reasoning}
Zijun Yao, Yantao Liu, Yanxu Chen, Jianhui Chen, Junfeng Fang, Lei Hou, Juanzi Li, and Tat-Seng Chua.
\newblock Are Reasoning Models More Prone to Hallucination?
\newblock {\em arXiv preprint arXiv:2505.23646}, 2025.

\bibitem{kirichenko2025abstentionbench}
Polina Kirichenko, Mark Ibrahim, Kamalika Chaudhuri, and Samuel~J Bell.
\newblock AbstentionBench: Reasoning LLMs Fail on Unanswerable Questions.
\newblock {\em arXiv preprint arXiv:2506.09038}, 2025.

\bibitem{lu2025auditing}
Haolang Lu, Yilian Liu, Jingxin Xu, Guoshun Nan, Yuanlong Yu, Zhican Chen, and Kun Wang.
\newblock Auditing Meta-Cognitive Hallucinations in Reasoning Large Language Models.
\newblock {\em arXiv preprint arXiv:2505.13143}, 2025.

\bibitem{li2025hallucination}
Junyi Li and Hwee~Tou Ng.
\newblock The Hallucination Dilemma: Factuality-Aware Reinforcement Learning for Large Reasoning Models.
\newblock {\em arXiv preprint arXiv:2505.24630}, 2025.

\bibitem{anh2025analyzing}
Dang~Hoang Anh, Vu~Tran, and Le~Minh Nguyen.
\newblock Analyzing Logical Fallacies in Large Language Models: A Study on Hallucination in Mathematical Reasoning.
\newblock In {\em JSAI International Symposium on Artificial Intelligence}, pages 179--195. Springer, 2025.

\bibitem{sun2025detection}
Zhongxiang Sun, Qipeng Wang, Haoyu Wang, Xiao Zhang, and Jun Xu.
\newblock Detection and Mitigation of Hallucination in Large Reasoning Models: A Mechanistic Perspective.
\newblock {\em arXiv preprint arXiv:2505.12886}, 2025.

\bibitem{guo2025mathematical}
Dadi Guo, Jiayu Liu, Zhiyuan Fan, Zhitao He, Haoran Li, Yumeng Wang, et~al.
\newblock Mathematical Proof as a Litmus Test: Revealing Failure Modes of Advanced Large Reasoning Models.
\newblock {\em arXiv preprint arXiv:2506.17114}, 2025.

\bibitem{li2024fine}
Ruosen Li, Ziming Luo, and Xinya Du.
\newblock Fine-grained Hallucination Detection and Mitigation in Language Model Mathematical Reasoning.
\newblock {\em arXiv preprint arXiv:2410.06304}, 2024.

\bibitem{zhang2025reasoning}
Anqi Zhang, Yulin Chen, Jane Pan, Chen Zhao, Aurojit Panda, Jinyang Li, and He~He.
\newblock Reasoning Models Know When They're Right: Probing Hidden States for Self-Verification.
\newblock {\em arXiv preprint arXiv:2504.05419}, 2025.

\bibitem{wang2025joint}
Changyue Wang, Weihang Su, Qingyao Ai, and Yiqun Liu.
\newblock Joint Evaluation of Answer and Reasoning Consistency for Hallucination Detection in Large Reasoning Models.
\newblock {\em arXiv preprint arXiv:2506.04832}, 2025.

\bibitem{lanham2023measuring}
Tamera Lanham, Anna Chen, Ansh Radhakrishnan, Benoit Steiner, Carson Denison, Danny Hernandez, Dustin Li, Esin Durmus, Evan Hubinger, Jackson Kernion, et~al.
\newblock Measuring faithfulness in chain-of-thought reasoning.
\newblock {\em arXiv preprint arXiv:2307.13702}, 2023.

\bibitem{turpin2023language}
Miles Turpin, Julian Michael, Ethan Perez, and Samuel Bowman.
\newblock Language models don't always say what they think: Unfaithful explanations in chain-of-thought prompting.
\newblock In {\em Proc. NeurIPS}, 2023.

\bibitem{tutek2025measuring}
Martin Tutek, Fateme~Hashemi Chaleshtori, Ana Marasovi{\'c}, and Yonatan Belinkov.
\newblock Measuring faithfulness of chains of thought by unlearning reasoning steps.
\newblock {\em arXiv preprint arXiv:2502.14829}, 2025.

\bibitem{xiong2025measuring}
Zidi Xiong, Chen Shan, Zhenting Qi, and Himabindu Lakkaraju.
\newblock Measuring the Faithfulness of Thinking Drafts in Large Reasoning Models.
\newblock {\em arXiv preprint arXiv:2505.13774}, 2025.

\bibitem{benthamchain}
Oliver Bentham, Nathan Stringham, and Ana Marasovic.
\newblock Chain-of-Thought Unfaithfulness as Disguised Accuracy.
\newblock {\em Transactions on Machine Learning Research}, 2024.

\bibitem{arcuschin2025chain}
Iv{\'a}n Arcuschin, Jett Janiak, Robert Krzyzanowski, Senthooran Rajamanoharan, Neel Nanda, and Arthur Conmy.
\newblock Chain-of-thought reasoning in the wild is not always faithful.
\newblock {\em arXiv preprint arXiv:2503.08679}, 2025.

\bibitem{chua2025deepseek}
James Chua and Owain Evans.
\newblock Are DeepSeek R1 And Other Reasoning Models More Faithful?
\newblock In {\em ICLR 2025 Workshop on Foundation Models in the Wild}, 2025.

\bibitem{chen2025reasoning}
Yanda Chen, Joe Benton, Ansh Radhakrishnan, Jonathan Uesato~Carson Denison, John Schulman, Arushi Somani, Peter Hase, Misha Wagner Fabien Roger~Vlad Mikulik, Sam Bowman, Jan Leike~Jared Kaplan, et~al.
\newblock Reasoning Models Don’t Always Say What They Think.
\newblock {\em Anthropic Research}, 2025.

\bibitem{li2024towards}
Jiachun Li, Pengfei Cao, Yubo Chen, Kang Liu, and Jun Zhao.
\newblock Towards faithful chain-of-thought: Large language models are bridging reasoners.
\newblock {\em arXiv preprint arXiv:2405.18915}, 2024.

\bibitem{agarwal2024faithfulness}
Chirag Agarwal, Sree~Harsha Tanneru, and Himabindu Lakkaraju.
\newblock Faithfulness vs. plausibility: On the (un) reliability of explanations from large language models.
\newblock {\em arXiv preprint arXiv:2402.04614}, 2024.

\bibitem{bao2024likely}
Guangsheng Bao, Hongbo Zhang, Cunxiang Wang, Linyi Yang, and Yue Zhang.
\newblock How Likely Do LLMs with CoT Mimic Human Reasoning?
\newblock In {\em Proc. COLING}, 2024.

\bibitem{tanneru2024difficulty}
Sree~Harsha Tanneru, Dan Ley, Chirag Agarwal, and Himabindu Lakkaraju.
\newblock On the difficulty of faithful chain-of-thought reasoning in large language models.
\newblock In {\em ICML Workshop on TiFA}, 2024.

\bibitem{lobo2024impact}
Elita Lobo, Chirag Agarwal, and Himabindu Lakkaraju.
\newblock On the impact of fine-tuning on chain-of-thought reasoning.
\newblock In {\em Proc. NAACL}, 2025.

\bibitem{paul2024making}
Debjit Paul, Robert West, Antoine Bosselut, and Boi Faltings.
\newblock Making Reasoning Matter: Measuring and Improving Faithfulness of Chain-of-Thought Reasoning.
\newblock In {\em Findings of Proc. EMNLP}, pages 15012--15032, 2024.

\bibitem{xu2024faithful}
Jundong Xu, Hao Fei, Liangming Pan, Qian Liu, Mong-Li Lee, and Wynne Hsu.
\newblock Faithful logical reasoning via symbolic chain-of-thought.
\newblock In {\em Proc. ACL}, 2024.

\bibitem{radhakrishnan2023question}
Ansh Radhakrishnan, Karina Nguyen, Anna Chen, Carol Chen, Carson Denison, Danny Hernandez, Esin Durmus, Evan Hubinger, Jackson Kernion, Kamil{\.e} Luko{\v{s}}i{\=u}t{\.e}, et~al.
\newblock Question decomposition improves the faithfulness of model-generated reasoning.
\newblock {\em arXiv preprint arXiv:2307.11768}, 2023.

\bibitem{lyu2023faithful}
Qing Lyu, Shreya Havaldar, Adam Stein, Li~Zhang, Delip Rao, Eric Wong, Marianna Apidianaki, and Chris Callison-Burch.
\newblock Faithful chain-of-thought reasoning.
\newblock In {\em Proc. IJCNLP-AACL}, 2023.

\bibitem{pan2023logic}
Liangming Pan, Alon Albalak, Xinyi Wang, and William~Yang Wang.
\newblock Logic-LM: Empowering Large Language Models with Symbolic Solvers for Faithful Logical Reasoning.
\newblock In {\em Proc. EMNLP}, 2023.

\bibitem{arakelyan2024flare}
Erik Arakelyan, Pasquale Minervini, Pat Verga, Patrick Lewis, and Isabelle Augenstein.
\newblock FLARE: Faithful Logic-Aided Reasoning and Exploration.
\newblock {\em arXiv preprint arXiv:2410.11900}, 2024.

\bibitem{leang2024comat}
Joshua Ong~Jun Leang, Aryo~Pradipta Gema, and Shay~B Cohen.
\newblock CoMAT: Chain of mathematically annotated thought improves mathematical reasoning.
\newblock {\em arXiv preprint arXiv:2410.10336}, 2024.

\bibitem{wang2024causal}
Jiawei Wang, Da~Cao, Shaofei Lu, Zhanchang Ma, Junbin Xiao, and Tat-Seng Chua.
\newblock Causal-driven Large Language Models with Faithful Reasoning for Knowledge Question Answering.
\newblock In {\em Proc. MM}, pages 4331--4340, 2024.

\bibitem{gao2024fact}
Minghe Gao, Shuang Chen, Liang Pang, Yuan Yao, Jisheng Dang, Wenqiao Zhang, Juncheng Li, Siliang Tang, Yueting Zhuang, and Tat-Seng Chua.
\newblock Fact: Teaching mllms with faithful, concise and transferable rationales.
\newblock In {\em Proc. MM}, pages 846--855, 2024.

\bibitem{viteri2024markovian}
Scott Viteri, Max Lamparth, Peter Chatain, and Clark Barrett.
\newblock Markovian Transformers for Informative Language Modeling.
\newblock {\em arXiv preprint arXiv:2404.18988}, 2024.

\bibitem{zhang2025safety}
Wenjing Zhang, Xuejiao Lei, Zhaoxiang Liu, Limin Han, Jiaojiao Zhao, Beibei Huang, Zhenhong Long, Junting Guo, Meijuan An, Rongjia Du, et~al.
\newblock Safety Evaluation and Enhancement of DeepSeek Models in Chinese Contexts.
\newblock {\em arXiv preprint arXiv:2503.16529}, 2025.

\bibitem{romero2025red}
Miguel Romero-Arjona, Pablo Valle, Juan~C Alonso, Ana~B S{\'a}nchez, Miriam Ugarte, Antonia Cazalilla, Vicente Cambr{\'o}n, Jos{\'e}~A Parejo, Aitor Arrieta, and Sergio Segura.
\newblock Red Teaming Contemporary AI Models: Insights from Spanish and Basque Perspectives.
\newblock {\em arXiv preprint arXiv:2503.10192}, 2025.

\bibitem{zhou2025hidden}
Kaiwen Zhou, Chengzhi Liu, Xuandong Zhao, Shreedhar Jangam, Jayanth Srinivasa, Gaowen Liu, Dawn Song, and Xin~Eric Wang.
\newblock The hidden risks of large reasoning models: A safety assessment of r1.
\newblock {\em arXiv preprint arXiv:2502.12659}, 2025.

\bibitem{li2025smarter}
Ang Li, Yichuan Mo, Mingjie Li, Yifei Wang, and Yisen Wang.
\newblock Are Smarter LLMs Safer? Exploring Safety-Reasoning Trade-offs in Prompting and Fine-Tuning.
\newblock {\em arXiv preprint arXiv:2502.09673}, 2025.

\bibitem{lou2025think}
Xinyue Lou, You Li, Jinan Xu, Xiangyu Shi, Chi Chen, and Kaiyu Huang.
\newblock Think in Safety: Unveiling and Mitigating Safety Alignment Collapse in Multimodal Large Reasoning Model.
\newblock {\em arXiv preprint arXiv:2505.06538}, 2025.

\bibitem{kassianik2025evaluating}
Paul Kassianik and Amin Karbasi.
\newblock Evaluating Security Risk in DeepSeek and Other Frontier Reasoning Models.
\newblock {\em Cisco, https://blogs. cisco. com/security/evaluating-security-risk-in-deepseek-and-other-frontier-reasoningmodels}, 2025.

\bibitem{krishna2025weakest}
Arjun Krishna, Aaditya Rastogi, and Erick Galinkin.
\newblock Weakest Link in the Chain: Security Vulnerabilities in Advanced Reasoning Models.
\newblock {\em arXiv preprint arXiv:2506.13726}, 2025.

\bibitem{knight2025fortress}
Christina~Q Knight, Kaustubh Deshpande, Ved Sirdeshmukh, Meher Mankikar, Scale~Red Team, SEAL Team, and Julian Michael.
\newblock FORTRESS: Frontier Risk Evaluation for National Security and Public Safety.
\newblock {\em arXiv preprint arXiv:2506.14922}, 2025.

\bibitem{fan2025evaluation}
Yihe Fan, Wenqi Zhang, Xudong Pan, and Min Yang.
\newblock Evaluation Faking: Unveiling Observer Effects in Safety Evaluation of Frontier AI Systems.
\newblock {\em arXiv preprint arXiv:2505.17815}, 2025.

\bibitem{zheng2025beyond}
Baihui Zheng, Boren Zheng, Kerui Cao, Yingshui Tan, Zhendong Liu, Weixun Wang, Jiaheng Liu, Jian Yang, Wenbo Su, Xiaoyong Zhu, et~al.
\newblock Beyond Safe Answers: A Benchmark for Evaluating True Risk Awareness in Large Reasoning Models.
\newblock {\em arXiv preprint arXiv:2505.19690}, 2025.

\bibitem{lu2025bench}
Xiaoya Lu, Zeren Chen, Xuhao Hu, Yijin Zhou, Weichen Zhang, Dongrui Liu, Lu~Sheng, and Jing Shao.
\newblock IS-Bench: Evaluating Interactive Safety of VLM-Driven Embodied Agents in Daily Household Tasks.
\newblock {\em arXiv preprint arXiv:2506.16402}, 2025.

\bibitem{fang2025safemlrm}
Junfeng Fang, Yukai Wang, Ruipeng Wang, Zijun Yao, Kun Wang, An~Zhang, Xiang Wang, and Tat-Seng Chua.
\newblock SafeMLRM: Demystifying Safety in Multi-modal Large Reasoning Models.
\newblock {\em arXiv preprint arXiv:2504.08813}, 2025.

\bibitem{zhao2025trade}
Weixiang Zhao, Xingyu Sui, Jiahe Guo, Yulin Hu, Yang Deng, Yanyan Zhao, Bing Qin, Wanxiang Che, Tat-Seng Chua, and Ting Liu.
\newblock Trade-offs in large reasoning models: An empirical analysis of deliberative and adaptive reasoning over foundational capabilities.
\newblock {\em arXiv preprint arXiv:2503.17979}, 2025.

\bibitem{marjanovic2025deepseek}
Sara~Vera Marjanovi{\'c}, Arkil Patel, Vaibhav Adlakha, Milad Aghajohari, Parishad BehnamGhader, Mehar Bhatia, Aditi Khandelwal, Austin Kraft, Benno Krojer, Xing~Han L{\`u}, et~al.
\newblock DeepSeek-R1 Thoughtology: Let's think about LLM Reasoning.
\newblock {\em arXiv preprint arXiv:2504.07128}, 2025.

\bibitem{sabbaghi2025adversarial}
Mahdi Sabbaghi, Paul Kassianik, George Pappas, Yaron Singer, Amin Karbasi, and Hamed Hassani.
\newblock Adversarial Reasoning at Jailbreaking Time.
\newblock In {\em Proc. ICML}, 2025.

\bibitem{su2024enhancing}
Jingbo Su.
\newblock Enhancing Adversarial Attacks through Chain of Thought.
\newblock {\em arXiv preprint arXiv:2410.21791}, 2024.

\bibitem{ying2025reasoning}
Zonghao Ying, Deyue Zhang, Zonglei Jing, Yisong Xiao, Quanchen Zou, Aishan Liu, Siyuan Liang, Xiangzheng Zhang, Xianglong Liu, and Dacheng Tao.
\newblock Reasoning-augmented conversation for multi-turn jailbreak attacks on large language models.
\newblock {\em arXiv preprint arXiv:2502.11054}, 2025.

\bibitem{chang2025chain}
Wenhan Chang, Tianqing Zhu, Yu~Zhao, Shuangyong Song, Ping Xiong, Wanlei Zhou, and Yongxiang Li.
\newblock Chain-of-Lure: A Synthetic Narrative-Driven Approach to Compromise Large Language Models.
\newblock {\em arXiv preprint arXiv:2505.17519}, 2025.

\bibitem{handa2024competency}
Divij Handa, Zehua Zhang, Amir Saeidi, Shrinidhi Kumbhar, and Chitta Baral.
\newblock When ``competency" in reasoning opens the door to vulnerability: Jailbreaking llms via novel complex ciphers.
\newblock {\em arXiv preprint arXiv:2402.10601}, 2024.

\bibitem{kuo2025h}
Martin Kuo, Jianyi Zhang, Aolin Ding, Qinsi Wang, Louis DiValentin, Yujia Bao, Wei Wei, Hai Li, and Yiran Chen.
\newblock H-cot: Hijacking the chain-of-thought safety reasoning mechanism to jailbreak large reasoning models, including openai o1/o3, deepseek-r1, and gemini 2.0 flash thinking.
\newblock {\em arXiv preprint arXiv:2502.12893}, 2025.

\bibitem{yao2025mousetrap}
Yang Yao, Xuan Tong, Ruofan Wang, Yixu Wang, Lujundong Li, Liang Liu, Yan Teng, and Yingchun Wang.
\newblock A mousetrap: Fooling large reasoning models for jailbreak with chain of iterative chaos.
\newblock {\em arXiv preprint arXiv:2502.15806}, 2025.

\bibitem{liang2025autoran}
Jiacheng Liang, Tanqiu Jiang, Yuhui Wang, Rongyi Zhu, Fenglong Ma, and Ting Wang.
\newblock AutoRAN: Weak-to-Strong Jailbreaking of Large Reasoning Models.
\newblock {\em arXiv preprint arXiv:2505.10846}, 2025.

\bibitem{nguyen2025three}
Viet-Anh Nguyen, Shiqian Zhao, Gia Dao, Runyi Hu, Yi~Xie, and Luu~Anh Tuan.
\newblock Three minds, one legend: Jailbreak large reasoning model with adaptive stacked ciphers.
\newblock {\em arXiv preprint arXiv:2505.16241}, 2025.

\bibitem{lian2025revealing}
Jiawei Lian, Jianhong Pan, Lefan Wang, Yi~Wang, Shaohui Mei, and Lap-Pui Chau.
\newblock Revealing the Intrinsic Ethical Vulnerability of Aligned Large Language Models.
\newblock {\em arXiv preprint arXiv:2504.05050}, 2025.

\bibitem{liu2025rrtl}
Yifei Liu, Yu~Cui, and Haibin Zhang.
\newblock RRTL: Red Teaming Reasoning Large Language Models in Tool Learning.
\newblock {\em arXiv preprint arXiv:2505.17106}, 2025.

\bibitem{sima2025viscra}
Bingrui Sima, Linhua Cong, Wenxuan Wang, and Kun He.
\newblock VisCRA: A Visual Chain Reasoning Attack for Jailbreaking Multimodal Large Language Models.
\newblock {\em arXiv preprint arXiv:2505.19684}, 2025.

\bibitem{ma2025hauntattack}
Jingyuan Ma, Rui Li, Zheng Li, Junfeng Liu, Lei Sha, and Zhifang Sui.
\newblock HauntAttack: When Attack Follows Reasoning as a Shadow.
\newblock {\em arXiv preprint arXiv:2506.07031}, 2025.

\bibitem{liu2025guardreasoner}
Yue Liu, Hongcheng Gao, Shengfang Zhai, Jun Xia, Tianyi Wu, Zhiwei Xue, Yulin Chen, Kenji Kawaguchi, Jiaheng Zhang, and Bryan Hooi.
\newblock GuardReasoner: Towards Reasoning-based LLM Safeguards.
\newblock {\em arXiv preprint arXiv:2501.18492}, 2025.

\bibitem{upadhayay2025x}
Bibek Upadhayay, Vahid Behzadan, et~al.
\newblock X-Guard: Multilingual guard agent for content moderation.
\newblock {\em arXiv preprint arXiv:2504.08848}, 2025.

\bibitem{yang2025mr}
Yahan Yang, Soham Dan, Shuo Li, Dan Roth, and Insup Lee.
\newblock MR. Guard: Multilingual Reasoning Guardrail using Curriculum Learning.
\newblock {\em arXiv preprint arXiv:2504.15241}, 2025.

\bibitem{zheng2025rsafe}
Jingnan Zheng, Xiangtian Ji, Yijun Lu, Chenhang Cui, Weixiang Zhao, Gelei Deng, Zhenkai Liang, An~Zhang, and Tat-Seng Chua.
\newblock RSafe: Incentivizing proactive reasoning to build robust and adaptive LLM safeguards.
\newblock {\em arXiv preprint arXiv:2506.07736}, 2025.

\bibitem{sreedhar2025safety}
Makesh~Narsimhan Sreedhar, Traian Rebedea, and Christopher Parisien.
\newblock Safety Through Reasoning: An Empirical Study of Reasoning Guardrail Models.
\newblock {\em arXiv preprint arXiv:2505.20087}, 2025.

\bibitem{kang2024r}
Mintong Kang and Bo~Li.
\newblock $R^{2}$-Guard: Robust Reasoning Enabled LLM Guardrail via Knowledge-Enhanced Logical Reasoning.
\newblock In {\em Proc. ICLR}, 2025.

\bibitem{cheng2025inverse}
Ruoxi Cheng, Haoxuan Ma, Weixin Wang, Zhiqiang Wang, Xiaoshuang Jia, Simeng Qin, Xiaochun Cao, Yang Liu, and Xiaojun Jia.
\newblock Inverse Reinforcement Learning with Dynamic Reward Scaling for LLM Alignment.
\newblock {\em arXiv preprint arXiv:2503.18991}, 2025.

\bibitem{cui2025shieldvlm}
Shiyao Cui, Qinglin Zhang, Xuan Ouyang, Renmiao Chen, Zhexin Zhang, Yida Lu, Hongning Wang, Han Qiu, and Minlie Huang.
\newblock ShieldVLM: Safeguarding the Multimodal Implicit Toxicity via Deliberative Reasoning with LVLMs.
\newblock {\em arXiv preprint arXiv:2505.14035}, 2025.

\bibitem{liu2025guardreasonervl}
Yue Liu, Shengfang Zhai, Mingzhe Du, Yulin Chen, Tri Cao, Hongcheng Gao, Cheng Wang, Xinfeng Li, Kun Wang, Junfeng Fang, et~al.
\newblock Guardreasoner-vl: Safeguarding vlms via reinforced reasoning.
\newblock {\em arXiv preprint arXiv:2505.11049}, 2025.

\bibitem{xiang2024guardagent}
Zhen Xiang, Linzhi Zheng, Yanjie Li, Junyuan Hong, Qinbin Li, Han Xie, Jiawei Zhang, Zidi Xiong, Chulin Xie, Carl Yang, et~al.
\newblock GuardAgent: Safeguard llm agents by a guard agent via knowledge-enabled reasoning.
\newblock {\em arXiv preprint arXiv:2406.09187}, 2024.

\bibitem{chen2025shieldagent}
Zhaorun Chen, Mintong Kang, and Bo~Li.
\newblock ShieldAgent: Shielding agents via verifiable safety policy reasoning.
\newblock {\em arXiv preprint arXiv:2503.22738}, 2025.

\bibitem{wang2025unified}
Yibin Wang, Zhimin Li, Yuhang Zang, Chunyu Wang, Qinglin Lu, Cheng Jin, and Jiaqi Wang.
\newblock Unified multimodal chain-of-thought reward model through reinforcement fine-tuning.
\newblock {\em arXiv preprint arXiv:2505.03318}, 2025.

\bibitem{pan2025detecting}
Fengjun Pan, Anh~Tuan Luu, and Xiaobao Wu.
\newblock Detecting Harmful Memes with Decoupled Understanding and Guided CoT Reasoning.
\newblock {\em arXiv preprint arXiv:2506.08477}, 2025.

\bibitem{wu2025effectively}
Tong Wu, Chong Xiang, Jiachen~T Wang, and Prateek Mittal.
\newblock Effectively Controlling Reasoning Models through Thinking Intervention.
\newblock {\em arXiv preprint arXiv:2503.24370}, 2025.

\bibitem{yamaguchi2025adversarial}
Kureha Yamaguchi, Benjamin Etheridge, and Andy Arditi.
\newblock Adversarial Manipulation of Reasoning Models using Internal Representations.
\newblock In {\em ICML 2025 Workshop on Reliable and Responsible Foundation Models}, 2025.

\bibitem{zaremba2025trading}
Wojciech Zaremba, Evgenia Nitishinskaya, Boaz Barak, Stephanie Lin, Sam Toyer, Yaodong Yu, Rachel Dias, Eric Wallace, Kai Xiao, Johannes Heidecke, et~al.
\newblock Trading inference-time compute for adversarial robustness.
\newblock {\em arXiv preprint arXiv:2501.18841}, 2025.

\bibitem{qiu2025saffron}
Ruizhong Qiu, Gaotang Li, Tianxin Wei, Jingrui He, and Hanghang Tong.
\newblock Saffron-1: Towards an Inference Scaling Paradigm for LLM Safety Assurance.
\newblock {\em arXiv preprint arXiv:2506.06444}, 2025.

\bibitem{liu2024mixture}
Zhili Liu, Yunhao Gou, Kai Chen, Lanqing Hong, Jiahui Gao, Fei Mi, Yu~Zhang, Zhenguo Li, Xin Jiang, Qun Liu, et~al.
\newblock Mixture of insightful experts (mote): The synergy of thought chains and expert mixtures in self-alignment.
\newblock {\em arXiv preprint arXiv:2405.00557}, 2024.

\bibitem{zhang2024backtracking}
Yiming Zhang, Jianfeng Chi, Hailey Nguyen, Kartikeya Upasani, Daniel~M Bikel, Jason Weston, and Eric~Michael Smith.
\newblock Backtracking improves generation safety.
\newblock In {\em Proc. ICLR}, 2025.

\bibitem{yang2025enhancing}
Xianglin Yang, Gelei Deng, Jieming Shi, Tianwei Zhang, and Jin~Song Dong.
\newblock Enhancing Model Defense Against Jailbreaks with Proactive Safety Reasoning.
\newblock {\em arXiv preprint arXiv:2501.19180}, 2025.

\bibitem{zhu2025reasoning}
Junda Zhu, Lingyong Yan, Shuaiqiang Wang, Dawei Yin, and Lei Sha.
\newblock Reasoning-to-Defend: Safety-Aware Reasoning Can Defend Large Language Models from Jailbreaking.
\newblock {\em arXiv preprint arXiv:2502.12970}, 2025.

\bibitem{zhang2025safety1}
Yuyou Zhang, Miao Li, William Han, Yihang Yao, Zhepeng Cen, and Ding Zhao.
\newblock Safety is Not Only About Refusal: Reasoning-Enhanced Fine-tuning for Interpretable LLM Safety.
\newblock {\em arXiv preprint arXiv:2503.05021}, 2025.

\bibitem{feng2025erpo}
Kehua Feng, Keyan Ding, Jing Yu, Menghan Li, Yuhao Wang, Tong Xu, Xinda Wang, Qiang Zhang, and Huajun Chen.
\newblock ERPO: Advancing Safety Alignment via Ex-Ante Reasoning Preference Optimization.
\newblock {\em arXiv preprint arXiv:2504.02725}, 2025.

\bibitem{mou2025saro}
Yutao Mou, Yuxiao Luo, Shikun Zhang, and Wei Ye.
\newblock SaRO: Enhancing LLM Safety through Reasoning-based Alignment.
\newblock {\em arXiv preprint arXiv:2504.09420}, 2025.

\bibitem{kim2025reasoning}
Taeyoun Kim, Fahim Tajwar, Aditi Raghunathan, and Aviral Kumar.
\newblock Reasoning as an Adaptive Defense for Safety.
\newblock {\em arXiv preprint arXiv:2507.00971}, 2025.

\bibitem{jiang2025think}
Changyue Jiang, Xudong Pan, and Min Yang.
\newblock Think Twice Before You Act: Enhancing Agent Behavioral Safety with Thought Correction.
\newblock {\em arXiv preprint arXiv:2505.11063}, 2025.

\bibitem{li2025reasoning}
Changyi Li, Jiayi Wang, Xudong Pan, Geng Hong, and Min Yang.
\newblock ReasoningShield: Content Safety Detection over Reasoning Traces of Large Reasoning Models.
\newblock {\em arXiv preprint arXiv:2505.17244}, 2025.

\bibitem{guan2024deliberative}
Melody~Y Guan, Manas Joglekar, Eric Wallace, Saachi Jain, Boaz Barak, Alec Helyar, Rachel Dias, Andrea Vallone, Hongyu Ren, Jason Wei, et~al.
\newblock Deliberative alignment: Reasoning enables safer language models.
\newblock {\em arXiv preprint arXiv:2412.16339}, 2024.

\bibitem{wang2025star}
Zijun Wang, Haoqin Tu, Yuhan Wang, Juncheng Wu, Jieru Mei, Brian~R Bartoldson, Bhavya Kailkhura, and Cihang Xie.
\newblock STAR-1: Safer Alignment of Reasoning LLMs with 1K Data.
\newblock {\em arXiv preprint arXiv:2504.01903}, 2025.

\bibitem{zhang2025realsafe}
Yichi Zhang, Zihao Zeng, Dongbai Li, Yao Huang, Zhijie Deng, and Yinpeng Dong.
\newblock RealSafe-R1: Safety-Aligned DeepSeek-R1 without Compromising Reasoning Capability.
\newblock {\em arXiv preprint arXiv:2504.10081}, 2025.

\bibitem{jeung2025safepath}
Wonje Jeung, Sangyeon Yoon, Minsuk Kahng, and Albert No.
\newblock SAFEPATH: Preventing Harmful Reasoning in Chain-of-Thought via Early Alignment.
\newblock {\em arXiv preprint arXiv:2505.14667}, 2025.

\bibitem{hu2025context}
Wenbin Hu, Haoran Li, Huihao Jing, Qi~Hu, Ziqian Zeng, Sirui Han, Heli Xu, Tianshu Chu, Peizhao Hu, and Yangqiu Song.
\newblock Context Reasoner: Incentivizing Reasoning Capability for Contextualized Privacy and Safety Compliance via Reinforcement Learning.
\newblock {\em arXiv preprint arXiv:2505.14585}, 2025.

\bibitem{baker2025monitoring}
Bowen Baker, Joost Huizinga, Leo Gao, Zehao Dou, Melody~Y Guan, Aleksander Madry, Wojciech Zaremba, Jakub Pachocki, and David Farhi.
\newblock Monitoring reasoning models for misbehavior and the risks of promoting obfuscation.
\newblock {\em arXiv preprint arXiv:2503.11926}, 2025.

\bibitem{zhang2025should}
Zhexin Zhang, Xian~Qi Loye, Victor Shea-Jay Huang, Junxiao Yang, Qi~Zhu, Shiyao Cui, Fei Mi, Lifeng Shang, Yingkang Wang, Hongning Wang, et~al.
\newblock How Should We Enhance the Safety of Large Reasoning Models: An Empirical Study.
\newblock {\em arXiv preprint arXiv:2505.15404}, 2025.

\bibitem{cheng2025hair}
Ruoxi Cheng, Haoxuan Ma, and Weixin Wang.
\newblock Hair: Hardness-aware inverse reinforcement learning with introspective reasoning for llm alignment.
\newblock {\em arXiv preprint arXiv:2503.18991}, 2025.

\bibitem{liu2025chasing}
Mickel Liu, Liwei Jiang, Yancheng Liang, Simon~Shaolei Du, Yejin Choi, Tim Althoff, and Natasha Jaques.
\newblock Chasing Moving Targets with Online Self-Play Reinforcement Learning for Safer Language Models.
\newblock {\em arXiv preprint arXiv:2506.07468}, 2025.

\bibitem{huang2025safety}
Tiansheng Huang, Sihao Hu, Fatih Ilhan, Selim~Furkan Tekin, Zachary Yahn, Yichang Xu, and Ling Liu.
\newblock Safety tax: Safety alignment makes your large reasoning models less reasonable.
\newblock {\em arXiv preprint arXiv:2503.00555}, 2025.

\bibitem{zhou2025safekey}
Kaiwen Zhou, Xuandong Zhao, Gaowen Liu, Jayanth Srinivasa, Aosong Feng, Dawn Song, and Xin~Eric Wang.
\newblock SafeKey: Amplifying Aha-Moment Insights for Safety Reasoning.
\newblock {\em arXiv preprint arXiv:2505.16186}, 2025.

\bibitem{jin2024saber}
Naizhu Jin, Zhong Li, Yinggang Guo, Chao Su, Tian Zhang, and Qingkai Zeng.
\newblock SABER: Model-agnostic Backdoor Attack on Chain-of-Thought in Neural Code Generation.
\newblock {\em arXiv preprint arXiv:2412.05829}, 2024.

\bibitem{zhu2025think}
Zihao Zhu, Hongbao Zhang, Ruotong Wang, Ke~Xu, Siwei Lyu, and Baoyuan Wu.
\newblock To Think or Not to Think: Exploring the Unthinking Vulnerability in Large Reasoning Models.
\newblock {\em arXiv preprint arXiv:2502.12202}, 2025.

\bibitem{zhao2025shadowcot}
Gejian Zhao, Hanzhou Wu, Xinpeng Zhang, and Athanasios~V Vasilakos.
\newblock Shadowcot: Cognitive hijacking for stealthy reasoning backdoors in llms.
\newblock {\em arXiv preprint arXiv:2504.05605}, 2025.

\bibitem{chua2025thought}
James Chua, Jan Betley, Mia Taylor, and Owain Evans.
\newblock Thought Crime: Backdoors and Emergent Misalignment in Reasoning Models.
\newblock {\em arXiv preprint arXiv:2506.13206}, 2025.

\bibitem{xiang2024badchain}
Zhen Xiang, Fengqing Jiang, Zidi Xiong, Bhaskar Ramasubramanian, Radha Poovendran, and Bo~Li.
\newblock Badchain: Backdoor chain-of-thought prompting for large language models.
\newblock In {\em Proc. ICLR}, 2024.

\bibitem{li2024backdoorllm}
Yige Li, Hanxun Huang, Yunhan Zhao, Xingjun Ma, and Jun Sun.
\newblock Backdoorllm: A comprehensive benchmark for backdoor attacks on large language models.
\newblock {\em arXiv preprint arXiv:2408.12798}, 2024.

\bibitem{guo2025darkmind}
Zhen Guo and Reza Tourani.
\newblock Darkmind: Latent chain-of-thought backdoor in customized llms.
\newblock {\em arXiv preprint arXiv:2501.18617}, 2025.

\bibitem{cui2025process}
Yu~Cui, Bryan Hooi, Yujun Cai, and Yiwei Wang.
\newblock Process or result? manipulated ending tokens can mislead reasoning llms to ignore the correct reasoning steps.
\newblock {\em arXiv preprint arXiv:2503.19326}, 2025.

\bibitem{guo2025system}
Jiawei Guo and Haipeng Cai.
\newblock System prompt poisoning: Persistent attacks on large language models beyond user injection.
\newblock {\em arXiv preprint arXiv:2505.06493}, 2025.

\bibitem{cui2025practical}
Yu~Cui and Cong Zuo.
\newblock Practical Reasoning Interruption Attacks on Reasoning Large Language Models.
\newblock {\em arXiv preprint arXiv:2505.06643}, 2025.

\bibitem{cui2025token}
Yu~Cui, Yujun Cai, and Yiwei Wang.
\newblock Token-Efficient Prompt Injection Attack: Provoking Cessation in LLM Reasoning via Adaptive Token Compression.
\newblock {\em arXiv preprint arXiv:2504.20493}, 2025.

\bibitem{song2025chain}
Hongru Song, Yu-an Liu, Ruqing Zhang, Jiafeng Guo, and Yixing Fan.
\newblock Chain-of-Thought Poisoning Attacks against R1-based Retrieval-Augmented Generation Systems.
\newblock {\em arXiv preprint arXiv:2505.16367}, 2025.

\bibitem{marinelli2025harnessing}
Ryan Marinelli, Josef Pichlmeier, and Tamas Bisztray.
\newblock Harnessing Chain-of-Thought Metadata for Task Routing and Adversarial Prompt Detection.
\newblock {\em arXiv preprint arXiv:2503.21464}, 2025.

\bibitem{jin2025guard}
Naizhu Jin, Zhong Li, Tian Zhang, and Qingkai Zeng.
\newblock GUARD: Dual-Agent based Backdoor Defense on Chain-of-Thought in Neural Code Generation.
\newblock {\em arXiv preprint arXiv:2505.21425}, 2025.

\bibitem{wang2025assessing}
Qian Wang, Zhanzhi Lou, Zhenheng Tang, Nuo Chen, Xuandong Zhao, Wenxuan Zhang, Dawn Song, and Bingsheng He.
\newblock Assessing Judging Bias in Large Reasoning Models: An Empirical Study.
\newblock {\em arXiv preprint arXiv:2504.09946}, 2025.

\bibitem{wang2025chain}
Wenxiao Wang, Parsa Hosseini, and Soheil Feizi.
\newblock Chain-of-Defensive-Thought: Structured Reasoning Elicits Robustness in Large Language Models against Reference Corruption.
\newblock {\em arXiv preprint arXiv:2504.20769}, 2025.

\bibitem{yan2025recitation}
Kai Yan, Yufei Xu, Zhengyin Du, Xuesong Yao, Zheyu Wang, Xiaowen Guo, and Jiecao Chen.
\newblock Recitation over Reasoning: How Cutting-Edge Language Models Can Fail on Elementary School-Level Reasoning Problems?
\newblock {\em arXiv preprint arXiv:2504.00509}, 2025.

\bibitem{yang2025any}
Haoyan Yang, Runxue Bao, Cao Xiao, Jun Ma, Parminder Bhatia, Shangqian Gao, and Taha Kass-Hout.
\newblock Any Large Language Model Can Be a Reliable Judge: Debiasing with a Reasoning-based Bias Detector.
\newblock {\em arXiv preprint arXiv:2505.17100}, 2025.

\bibitem{wang2024rupbench}
Yuqing Wang and Yun Zhao.
\newblock Rupbench: Benchmarking reasoning under perturbations for robustness evaluation in large language models.
\newblock {\em arXiv preprint arXiv:2406.11020}, 2024.

\bibitem{mu2025closer}
Norman Mu, Jonathan Lu, Michael Lavery, and David Wagner.
\newblock A Closer Look at System Prompt Robustness.
\newblock {\em arXiv preprint arXiv:2502.12197}, 2025.

\bibitem{li2025frustratingly}
Zhaoyi Li, Xiaohan Zhao, Dong-Dong Wu, Jiacheng Cui, and Zhiqiang Shen.
\newblock A Frustratingly Simple Yet Highly Effective Attack Baseline: Over 90\% Success Rate Against the Strong Black-box Models of GPT-4.5/4o/o1.
\newblock {\em arXiv preprint arXiv:2503.10635}, 2025.

\bibitem{zhu2025reasoning1}
Bin Zhu, Hailong Yin, Jingjing Chen, and Yu-Gang Jiang.
\newblock Reasoning Models Are More Easily Gaslighted Than You Think.
\newblock {\em arXiv preprint arXiv:2506.09677}, 2025.

\bibitem{zhou2024can}
Zhanke Zhou, Rong Tao, Jianing Zhu, Yiwen Luo, Zengmao Wang, and Bo~Han.
\newblock Can Language Models Perform Robust Reasoning in Chain-of-thought Prompting with Noisy Rationales?
\newblock In {\em Proc. NeurIPS}, 2024.

\bibitem{peng2025stepwise}
Jingyu Peng, Maolin Wang, Xiangyu Zhao, Kai Zhang, Wanyu Wang, Pengyue Jia, Qidong Liu, Ruocheng Guo, and Qi~Liu.
\newblock Stepwise Reasoning Disruption Attack of LLMs.
\newblock In {\em Proc. ACL}, pages 5040--5058, 2025.

\bibitem{wang2025polymath}
Yiming Wang, Pei Zhang, Jialong Tang, Haoran Wei, Baosong Yang, Rui Wang, Chenshu Sun, Feitong Sun, Jiran Zhang, Junxuan Wu, et~al.
\newblock Polymath: Evaluating mathematical reasoning in multilingual contexts.
\newblock {\em arXiv preprint arXiv:2504.18428}, 2025.

\bibitem{rajeev2025cats}
Meghana Rajeev, Rajkumar Ramamurthy, Prapti Trivedi, Vikas Yadav, Oluwanifemi Bamgbose, Sathwik~Tejaswi Madhusudan, James Zou, and Nazneen Rajani.
\newblock Cats Confuse Reasoning LLM: Query Agnostic Adversarial Triggers for Reasoning Models.
\newblock {\em arXiv preprint arXiv:2503.01781}, 2025.

\bibitem{yu2025benchmarking}
Tong Yu, Yongcheng Jing, Xikun Zhang, Wentao Jiang, Wenjie Wu, Yingjie Wang, Wenbin Hu, Bo~Du, and Dacheng Tao.
\newblock Benchmarking reasoning robustness in large language models.
\newblock {\em arXiv preprint arXiv:2503.04550}, 2025.

\bibitem{huang2025math}
Kaixuan Huang, Jiacheng Guo, Zihao Li, Xiang Ji, Jiawei Ge, Wenzhe Li, Yingqing Guo, Tianle Cai, Hui Yuan, Runzhe Wang, et~al.
\newblock MATH-Perturb: Benchmarking LLMs' Math Reasoning Abilities against Hard Perturbations.
\newblock {\em arXiv preprint arXiv:2502.06453}, 2025.

\bibitem{lam2025codecrash}
Man~Ho Lam, Chaozheng Wang, Jen-tse Huang, and Michael~R Lyu.
\newblock CODECRASH: Stress Testing LLM Reasoning under Structural and Semantic Perturbations.
\newblock {\em arXiv preprint arXiv:2504.14119}, 2025.

\bibitem{roh2025break}
Jaechul Roh, Varun Gandhi, Shivani Anilkumar, and Arin Garg.
\newblock Chain-of-Code Collapse: Reasoning Failures in LLMs via Adversarial Prompting in Code Generation.
\newblock {\em arXiv preprint arXiv:2506.06971}, 2025.

\bibitem{xu2024preemptive}
Rongwu Xu, Zehan Qi, and Wei Xu.
\newblock Preemptive answer ``attacks" on chain-of-thought reasoning.
\newblock In {\em Findings of Proc. ACL}, 2024.

\bibitem{ma2024large}
Jingyuan Ma, Damai Dai, Lei Sha, and Zhifang Sui.
\newblock Large language models are unconscious of unreasonability in math problems.
\newblock {\em arXiv preprint arXiv:2403.19346}, 2024.

\bibitem{hashemi2025dnr}
Masoud Hashemi, Oluwanifemi Bamgbose, Sathwik~Tejaswi Madhusudhan, Jishnu~Sethumadhavan Nair, Aman Tiwari, and Vikas Yadav.
\newblock Dnr bench: Benchmarking over-reasoning in reasoning llms.
\newblock {\em arXiv preprint arXiv:2503.15793}, 2025.

\bibitem{he2025can}
Yancheng He, Shilong Li, Jiaheng Liu, Weixun Wang, Xingyuan Bu, Ge~Zhang, Zhongyuan Peng, Zhaoxiang Zhang, Zhicheng Zheng, Wenbo Su, et~al.
\newblock Can Large Language Models Detect Errors in Long Chain-of-Thought Reasoning?
\newblock In {\em Proc. ACL}, 2025.

\bibitem{fan2025missing}
Chenrui Fan, Ming Li, Lichao Sun, and Tianyi Zhou.
\newblock Missing Premise exacerbates Overthinking: Are Reasoning Models losing Critical Thinking Skill?
\newblock {\em arXiv preprint arXiv:2504.06514}, 2025.

\bibitem{si2025excessive}
Wai~Man Si, Mingjie Li, Michael Backes, and Yang Zhang.
\newblock Excessive Reasoning Attack on Reasoning LLMs.
\newblock {\em arXiv preprint arXiv:2506.14374}, 2025.

\bibitem{wang2025thoughts}
Yue Wang, Qiuzhi Liu, Jiahao Xu, Tian Liang, Xingyu Chen, Zhiwei He, Linfeng Song, Dian Yu, Juntao Li, Zhuosheng Zhang, et~al.
\newblock Thoughts Are All Over the Place: On the Underthinking of o1-Like LLMs.
\newblock {\em arXiv preprint arXiv:2501.18585}, 2025.

\bibitem{su2025between}
Jinyan Su, Jennifer Healey, Preslav Nakov, and Claire Cardie.
\newblock Between underthinking and overthinking: An empirical study of reasoning length and correctness in llms.
\newblock {\em arXiv preprint arXiv:2505.00127}, 2025.

\bibitem{dang2025internal}
Renfei Dang, Shujian Huang, and Jiajun Chen.
\newblock Internal Bias in Reasoning Models leads to Overthinking.
\newblock {\em arXiv preprint arXiv:2505.16448}, 2025.

\bibitem{kumar2025overthink}
Abhinav Kumar, Jaechul Roh, Ali Naseh, Marzena Karpinska, Mohit Iyyer, Amir Houmansadr, and Eugene Bagdasarian.
\newblock Overthink: Slowdown attacks on reasoning llms.
\newblock {\em arXiv preprint arXiv:2502.02542}, 2025.

\bibitem{cuadron2025danger}
Alejandro Cuadron, Dacheng Li, Wenjie Ma, Xingyao Wang, Yichuan Wang, Siyuan Zhuang, Shu Liu, Luis~Gaspar Schroeder, Tian Xia, Huanzhi Mao, et~al.
\newblock The Danger of Overthinking: Examining the Reasoning-Action Dilemma in Agentic Tasks.
\newblock {\em arXiv preprint arXiv:2502.08235}, 2025.

\bibitem{li2025output}
Xuying Li, Zhuo Li, Yuji Kosuga, and Victor Bian.
\newblock Output Length Effect on DeepSeek-R1's Safety in Forced Thinking.
\newblock {\em arXiv preprint arXiv:2503.01923}, 2025.

\bibitem{sun2025thinkedit}
Chung-En Sun, Ge~Yan, and Tsui-Wei Weng.
\newblock ThinkEdit: Interpretable Weight Editing to Mitigate Overly Short Thinking in Reasoning Models.
\newblock {\em arXiv preprint arXiv:2503.22048}, 2025.

\bibitem{linassessing}
Fangru Lin, Shaoguang Mao, Emanuele La~Malfa, Valentin Hofmann, Adrian de~Wynter, Xun Wang, Si-Qing Chen, Michael~J Wooldridge, Janet~B Pierrehumbert, and Furu Wei.
\newblock Assessing Dialect Fairness and Robustness of Large Language Models in Reasoning Tasks.
\newblock In {\em Proc. ACL}, 2025.

\bibitem{cheng2025detection}
Xiaoqing Cheng, Hongying Zan, Lulu Kong, Jinwang Song, and Min Peng.
\newblock Detection, Classification, and Mitigation of Gender Bias in Large Language Models.
\newblock {\em arXiv preprint arXiv:2506.12527}, 2025.

\bibitem{kamruzzaman2024prompting}
Mahammed Kamruzzaman and Gene~Louis Kim.
\newblock Prompting techniques for reducing social bias in llms through system 1 and system 2 cognitive processes.
\newblock {\em arXiv preprint arXiv:2404.17218}, 2024.

\bibitem{dash2025persona}
Saloni Dash, Am{\'e}lie Reymond, Emma~S Spiro, and Aylin Caliskan.
\newblock Persona-Assigned Large Language Models Exhibit Human-Like Motivated Reasoning.
\newblock {\em arXiv preprint arXiv:2506.20020}, 2025.

\bibitem{gupta2023bias}
Shashank Gupta, Vaishnavi Shrivastava, Ameet Deshpande, Ashwin Kalyan, Peter Clark, Ashish Sabharwal, and Tushar Khot.
\newblock Bias runs deep: Implicit reasoning biases in persona-assigned llms.
\newblock In {\em Proc. ICLR}, 2024.

\bibitem{fan2025biasguard}
Zhiting Fan, Ruizhe Chen, and Zuozhu Liu.
\newblock Biasguard: A reasoning-enhanced bias detection tool for large language models.
\newblock In {\em Findings of Proc. ACL}, 2025.

\bibitem{cantini2025reasoning}
Riccardo Cantini, Nicola Gabriele, Alessio Orsino, and Domenico Talia.
\newblock Is Reasoning All You Need? Probing Bias in the Age of Reasoning Language Models.
\newblock {\em arXiv preprint arXiv:2507.02799}, 2025.

\bibitem{yoon2025r}
Sangyeon Yoon, Wonje Jeung, and Albert No.
\newblock R-tofu: Unlearning in large reasoning models.
\newblock {\em arXiv preprint arXiv:2505.15214}, 2025.

\bibitem{wang2025reasoning}
Changsheng Wang, Chongyu Fan, Yihua Zhang, Jinghan Jia, Dennis Wei, Parikshit Ram, Nathalie Baracaldo, and Sijia Liu.
\newblock Reasoning Model Unlearning: Forgetting Traces, Not Just Answers, While Preserving Reasoning Skills.
\newblock {\em arXiv preprint arXiv:2506.12963}, 2025.

\bibitem{sinha2025step}
Yash Sinha, Manit Baser, Murari Mandal, Dinil~Mon Divakaran, and Mohan Kankanhalli.
\newblock Step-by-Step Reasoning Attack: Revealing 'Erased' Knowledge in Large Language Models.
\newblock {\em arXiv preprint arXiv:2506.17279}, 2025.

\bibitem{wanli2025imf}
Peng Wanli, Xue Yiming, et~al.
\newblock ImF: Implicit Fingerprint for Large Language Models.
\newblock {\em arXiv preprint arXiv:2503.21805}, 2025.

\bibitem{ren2025cotsrf}
Zhenzhen Ren, GuoBiao Li, Sheng Li, Zhenxing Qian, and Xinpeng Zhang.
\newblock CoTSRF: Utilize Chain of Thought as Stealthy and Robust Fingerprint of Large Language Models.
\newblock {\em arXiv preprint arXiv:2505.16785}, 2025.

\bibitem{guo2025towards}
Junfeng Guo, Yiming Li, Ruibo Chen, Yihan Wu, Chenxi Liu, Yanshuo Chen, and Heng Huang.
\newblock Towards copyright protection for knowledge bases of retrieval-augmented language models via ownership verification with reasoning.
\newblock {\em arXiv preprint arXiv:2502.10440}, 2025.

\bibitem{savani2025antidistillation}
Yash Savani, Asher Trockman, Zhili Feng, Avi Schwarzschild, Alexander Robey, Marc Finzi, and J~Zico Kolter.
\newblock Antidistillation sampling.
\newblock {\em arXiv preprint arXiv:2504.13146}, 2025.

\bibitem{green2025leaky}
Tommaso Green, Martin Gubri, Haritz Puerto, Sangdoo Yun, and Seong~Joon Oh.
\newblock Leaky Thoughts: Large Reasoning Models Are Not Private Thinkers.
\newblock {\em arXiv preprint arXiv:2506.15674}, 2025.

\bibitem{luo2025doxing}
Weidi Luo, Tianyu Lu, Qiming Zhang, Xiaogeng Liu, Bin Hu, Yue Zhao, Jieyu Zhao, Song Gao, Patrick McDaniel, Zhen Xiang, et~al.
\newblock Doxing via the Lens: Revealing Location-related Privacy Leakage on Multi-modal Large Reasoning Models.
\newblock {\em arXiv preprint arXiv:2504.19373}, 2025.

\bibitem{huang2024trustllm}
Yue Huang, Lichao Sun, Haoran Wang, Siyuan Wu, Qihui Zhang, Yuan Li, Chujie Gao, Yixin Huang, Wenhan Lyu, et~al.
\newblock TrustLLM: Trustworthiness in Large Language Models.
\newblock In {\em Proc. ICML}, 2024.

\bibitem{huang2025survey}
Lei Huang, Weijiang Yu, Weitao Ma, Weihong Zhong, Zhangyin Feng, Haotian Wang, Qianglong Chen, Weihua Peng, Xiaocheng Feng, Bing Qin, et~al.
\newblock A survey on hallucination in large language models: Principles, taxonomy, challenges, and open questions.
\newblock {\em ACM Transactions on Information Systems}, 43(2):1--55, 2025.

\bibitem{rawte2023survey}
Vipula Rawte, Amit Sheth, and Amitava Das.
\newblock A survey of hallucination in large foundation models.
\newblock {\em arXiv preprint arXiv:2309.05922}, 2023.

\bibitem{cheng2025chain}
Jiahao Cheng, Tiancheng Su, Jia Yuan, Guoxiu He, Jiawei Liu, Xinqi Tao, Jingwen Xie, and Huaxia Li.
\newblock Chain-of-Thought Prompting Obscures Hallucination Cues in Large Language Models: An Empirical Evaluation.
\newblock {\em arXiv preprint arXiv:2506.17088}, 2025.

\bibitem{snell2024scaling}
Charlie Snell, Jaehoon Lee, Kelvin Xu, and Aviral Kumar.
\newblock Scaling llm test-time compute optimally can be more effective than scaling model parameters.
\newblock {\em arXiv preprint arXiv:2408.03314}, 2024.

\bibitem{wang2022self}
Xuezhi Wang, Jason Wei, Dale Schuurmans, Quoc Le, Ed~Chi, Sharan Narang, Aakanksha Chowdhery, and Denny Zhou.
\newblock Self-consistency improves chain of thought reasoning in language models.
\newblock In {\em Proc. ICLR}, 2023.

\bibitem{lin2021truthfulqa}
Stephanie Lin, Jacob Hilton, and Owain Evans.
\newblock Truthfulqa: Measuring how models mimic human falsehoods.
\newblock In {\em Proc. ACL}, 2022.

\bibitem{li2023halueval}
Junyi Li, Xiaoxue Cheng, Wayne~Xin Zhao, Jian-Yun Nie, and Ji-Rong Wen.
\newblock Halueval: A large-scale hallucination evaluation benchmark for large language models.
\newblock In {\em Proc. EMNLP}, 2023.

\bibitem{cheng2023evaluating}
Qinyuan Cheng, Tianxiang Sun, Wenwei Zhang, Siyin Wang, Xiangyang Liu, Mozhi Zhang, Junliang He, Mianqiu Huang, Zhangyue Yin, Kai Chen, et~al.
\newblock Evaluating hallucinations in chinese large language models.
\newblock {\em arXiv preprint arXiv:2310.03368}, 2023.

\bibitem{wei2024measuring}
Jason Wei, Nguyen Karina, Hyung~Won Chung, Yunxin~Joy Jiao, Spencer Papay, Amelia Glaese, John Schulman, and William Fedus.
\newblock Measuring short-form factuality in large language models.
\newblock {\em arXiv preprint arXiv:2411.04368}, 2024.

\bibitem{joshi2017triviaqa}
Mandar Joshi, Eunsol Choi, Daniel~S Weld, and Luke Zettlemoyer.
\newblock Triviaqa: A large scale distantly supervised challenge dataset for reading comprehension.
\newblock {\em arXiv preprint arXiv:1705.03551}, 2017.

\bibitem{li2022faithfulness}
Wei Li, Wenhao Wu, Moye Chen, Jiachen Liu, Xinyan Xiao, and Hua Wu.
\newblock Faithfulness in natural language generation: A systematic survey of analysis, evaluation and optimization methods.
\newblock {\em arXiv preprint arXiv:2203.05227}, 2022.

\bibitem{jacovi2020towards}
Alon Jacovi and Yoav Goldberg.
\newblock Towards Faithfully Interpretable NLP Systems: How Should We Define and Evaluate Faithfulness?
\newblock In {\em Proc. ACL}, pages 4198--4205, 2020.

\bibitem{yee2024dissociation}
Evelyn Yee, Alice Li, Chenyu Tang, Yeon~Ho Jung, Ramamohan Paturi, and Leon Bergen.
\newblock Dissociation of faithful and unfaithful reasoning in llms.
\newblock {\em arXiv preprint arXiv:2405.15092}, 2024.

\bibitem{hase2020leakage}
Peter Hase, Shiyue Zhang, Harry Xie, and Mohit Bansal.
\newblock Leakage-Adjusted Simulatability: Can Models Generate Non-Trivial Explanations of Their Behavior in Natural Language?
\newblock In {\em Findings of Proc. EMNLP}, pages 4351--4367, 2020.

\bibitem{zhang2024negative}
Ruiqi Zhang, Licong Lin, Yu~Bai, and Song Mei.
\newblock Negative preference optimization: From catastrophic collapse to effective unlearning.
\newblock In {\em Proc. COLM}, 2024.

\bibitem{pearl2009causality}
Judea Pearl.
\newblock {\em Causality}.
\newblock Cambridge university press, 2009.

\bibitem{hsieh2023distilling}
Cheng-Yu Hsieh, Chun-Liang Li, Chih-kuan Yeh, Hootan Nakhost, Yasuhisa Fujii, Alex Ratner, Ranjay Krishna, Chen-Yu Lee, and Tomas Pfister.
\newblock Distilling Step-by-Step! Outperforming Larger Language Models with Less Training Data and Smaller Model Sizes.
\newblock In {\em Findings of Proc. ACL}, pages 8003--8017, 2023.

\bibitem{mazeika2024harmbench}
Mantas Mazeika, Long Phan, Xuwang Yin, Andy Zou, Zifan Wang, Norman Mu, Elham Sakhaee, Nathaniel Li, Steven Basart, Bo~Li, et~al.
\newblock Harmbench: A standardized evaluation framework for automated red teaming and robust refusal.
\newblock In {\em Proc. ICML}, 2024.

\bibitem{souly2024strongreject}
Alexandra Souly, Qingyuan Lu, Dillon Bowen, Tu~Trinh, Elvis Hsieh, Sana Pandey, Pieter Abbeel, Justin Svegliato, Scott Emmons, Olivia Watkins, et~al.
\newblock A strongreject for empty jailbreaks.
\newblock In {\em Proc. NeurIPS D\&B Track}, 2024.

\bibitem{zeng2024air}
Yi~Zeng, Yu~Yang, Andy Zhou, Jeffrey~Ziwei Tan, Yuheng Tu, Yifan Mai, Kevin Klyman, Minzhou Pan, Ruoxi Jia, Dawn Song, et~al.
\newblock Air-bench 2024: A safety benchmark based on risk categories from regulations and policies.
\newblock {\em arXiv preprint arXiv:2407.17436}, 2024.

\bibitem{hanwildguard}
Seungju Han, Kavel Rao, Allyson Ettinger, Liwei Jiang, Bill~Yuchen Lin, Nathan Lambert, Yejin Choi, and Nouha Dziri.
\newblock WildGuard: Open One-stop Moderation Tools for Safety Risks, Jailbreaks, and Refusals of LLMs.
\newblock In {\em Proc. NeurIPS D\&B Track}, 2024.

\bibitem{zou2023universal}
Andy Zou, Zifan Wang, Nicholas Carlini, Milad Nasr, J~Zico Kolter, and Matt Fredrikson.
\newblock Universal and transferable adversarial attacks on aligned language models.
\newblock {\em arXiv preprint arXiv:2307.15043}, 2023.

\bibitem{chao2023jailbreaking}
Patrick Chao, Alexander Robey, Edgar Dobriban, Hamed Hassani, George~J Pappas, and Eric Wong.
\newblock Jailbreaking black box large language models in twenty queries.
\newblock In {\em Proc. SaTML}, 2025.

\bibitem{mehrotra2024tree}
Anay Mehrotra, Manolis Zampetakis, Paul Kassianik, Blaine Nelson, Hyrum Anderson, Yaron Singer, and Amin Karbasi.
\newblock Tree of attacks: Jailbreaking black-box llms automatically.
\newblock In {\em Proc. NeurIPS}, 2024.

\bibitem{deepmind_gemini}
Google DeepMind.
\newblock Gemini 2.0 Flash Thinking, 2025.

\bibitem{team2025kimi}
Kimi Team, Angang Du, Bofei Gao, Bowei Xing, Changjiu Jiang, Cheng Chen, Cheng Li, Chenjun Xiao, Chenzhuang Du, Chonghua Liao, et~al.
\newblock Kimi k1.5: Scaling reinforcement learning with llms.
\newblock {\em arXiv preprint arXiv:2501.12599}, 2025.

\bibitem{sky_t1_2025}
NovaSky Team.
\newblock Sky-T1: Train your own O1 preview model within \$450.
\newblock https://novasky-ai.github.io/posts/sky-t1, 2025.
\newblock Accessed: 2025-01-09.

\bibitem{qwq32b}
Qwen Team.
\newblock QwQ-32B: Embracing the Power of Reinforcement Learning, March 2025.

\bibitem{skyworkopeno12024}
Skywork o1~Team.
\newblock Skywork-o1 Open Series.
\newblock \url{https://huggingface.co/Skywork}, November 2024.

\bibitem{jiang2024wildteaming}
Liwei Jiang, Kavel Rao, Seungju Han, Allyson Ettinger, Faeze Brahman, Sachin Kumar, Niloofar Mireshghallah, Ximing Lu, Maarten Sap, Yejin Choi, et~al.
\newblock Wildteaming at scale: From in-the-wild jailbreaks to (adversarially) safer language models.
\newblock In {\em Proc. NeurIPS}, 2024.

\bibitem{wan2024cyberseceval}
Shengye Wan, Cyrus Nikolaidis, Daniel Song, David Molnar, James Crnkovich, Jayson Grace, Manish Bhatt, Sahana Chennabasappa, Spencer Whitman, Stephanie Ding, et~al.
\newblock Cyberseceval 3: Advancing the evaluation of cybersecurity risks and capabilities in large language models.
\newblock {\em arXiv preprint arXiv:2408.01605}, 2024.

\bibitem{zhang2024chisafetybench}
Wenjing Zhang, Xuejiao Lei, Zhaoxiang Liu, Meijuan An, Bikun Yang, KaiKai Zhao, Kai Wang, and Shiguo Lian.
\newblock Chisafetybench: A chinese hierarchical safety benchmark for large language models.
\newblock {\em arXiv preprint arXiv:2406.10311}, 2024.

\bibitem{wei2023jailbroken}
Alexander Wei, Nika Haghtalab, and Jacob Steinhardt.
\newblock Jailbroken: How does llm safety training fail?
\newblock In {\em Proc. NeurIPS}, 2023.

\bibitem{yang2025r1}
Yi~Yang, Xiaoxuan He, Hongkun Pan, Xiyan Jiang, Yan Deng, Xingtao Yang, Haoyu Lu, Dacheng Yin, Fengyun Rao, Minfeng Zhu, et~al.
\newblock R1-onevision: Advancing generalized multimodal reasoning through cross-modal formalization.
\newblock {\em arXiv preprint arXiv:2503.10615}, 2025.

\bibitem{peng2025skywork}
Yi~Peng, Xiaokun Wang, Yichen Wei, Jiangbo Pei, Weijie Qiu, Ai~Jian, Yunzhuo Hao, Jiachun Pan, Tianyidan Xie, Li~Ge, et~al.
\newblock Skywork r1v: Pioneering multimodal reasoning with chain-of-thought.
\newblock {\em arXiv preprint arXiv:2504.05599}, 2025.

\bibitem{peng2025lmm}
Yingzhe Peng, Gongrui Zhang, Miaosen Zhang, Zhiyuan You, Jie Liu, Qipeng Zhu, Kai Yang, Xingzhong Xu, Xin Geng, and Xu~Yang.
\newblock Lmm-r1: Empowering 3b lmms with strong reasoning abilities through two-stage rule-based rl.
\newblock {\em arXiv preprint arXiv:2503.07536}, 2025.

\bibitem{gou2024eyes}
Yunhao Gou, Kai Chen, Zhili Liu, Lanqing Hong, Hang Xu, Zhenguo Li, Dit-Yan Yeung, James~T Kwok, and Yu~Zhang.
\newblock Eyes closed, safety on: Protecting multimodal llms via image-to-text transformation.
\newblock In {\em Proc. ECCV}, pages 388--404, 2024.

\bibitem{wang2025we}
Yanbo Wang, Jiyang Guan, Jian Liang, and Ran He.
\newblock Do We Really Need Curated Malicious Data for Safety Alignment in Multi-modal Large Language Models?
\newblock {\em arXiv preprint arXiv:2504.10000}, 2025.

\bibitem{greshake2023more}
Kai Greshake, Sahar Abdelnabi, Shailesh Mishra, Christoph Endres, Thorsten Holz, and Mario Fritz.
\newblock More than you’ve asked for: A comprehensive analysis of novel prompt injection threats to application-integrated large language models.
\newblock {\em arXiv preprint arXiv:2302.12173}, 2023.

\bibitem{xie2023defending}
Yueqi Xie, Jingwei Yi, Jiawei Shao, Justin Curl, Lingjuan Lyu, Qifeng Chen, Xing Xie, and Fangzhao Wu.
\newblock Defending chatgpt against jailbreak attack via self-reminders.
\newblock {\em Nature Machine Intelligence}, 5(12):1486--1496, 2023.

\bibitem{zhang2023defending}
Zhexin Zhang, Junxiao Yang, Pei Ke, Fei Mi, Hongning Wang, and Minlie Huang.
\newblock Defending large language models against jailbreaking attacks through goal prioritization.
\newblock In {\em Proc. ACL}, 2024.

\bibitem{wei2023jailbreak}
Zeming Wei, Yifei Wang, Ang Li, Yichuan Mo, and Yisen Wang.
\newblock Jailbreak and guard aligned language models with only few in-context demonstrations.
\newblock {\em arXiv preprint arXiv:2310.06387}, 2023.

\bibitem{xiong2024defensive}
Chen Xiong, Xiangyu Qi, Pin-Yu Chen, and Tsung-Yi Ho.
\newblock Defensive Prompt Patch: A Robust and Generalizable Defense of Large Language Models against Jailbreak Attacks.
\newblock In {\em Findings of Proc. ACL}, 2025.

\bibitem{zeng2024root}
Xinyi Zeng, Yuying Shang, Jiawei Chen, Jingyuan Zhang, and Yu~Tian.
\newblock Root defence strategies: Ensuring safety of llm at the decoding level.
\newblock {\em arXiv preprint arXiv:2410.06809}, 2024.

\bibitem{xu2024safedecoding}
Zhangchen Xu, Fengqing Jiang, Luyao Niu, Jinyuan Jia, Bill~Yuchen Lin, and Radha Poovendran.
\newblock Safedecoding: Defending against jailbreak attacks via safety-aware decoding.
\newblock {\em arXiv preprint arXiv:2402.08983}, 2024.

\bibitem{banerjee2025safeinfer}
Somnath Banerjee, Sayan Layek, Soham Tripathy, Shanu Kumar, Animesh Mukherjee, and Rima Hazra.
\newblock Safeinfer: Context adaptive decoding time safety alignment for large language models.
\newblock In {\em Proc. AAAI}, pages 27188--27196, 2025.

\bibitem{dong2024safeguarding}
Yi~Dong, Ronghui Mu, Yanghao Zhang, Siqi Sun, Tianle Zhang, Changshun Wu, Gaojie Jin, Yi~Qi, Jinwei Hu, Jie Meng, et~al.
\newblock Safeguarding large language models: A survey.
\newblock {\em arXiv preprint arXiv:2406.02622}, 2024.

\bibitem{inan2023llama}
Hakan Inan, Kartikeya Upasani, Jianfeng Chi, Rashi Rungta, Krithika Iyer, Yuning Mao, Michael Tontchev, Qing Hu, Brian Fuller, Davide Testuggine, et~al.
\newblock Llama guard: Llm-based input-output safeguard for human-ai conversations.
\newblock {\em arXiv preprint arXiv:2312.06674}, 2023.

\bibitem{chi2024llama}
Jianfeng Chi, Ujjwal Karn, Hongyuan Zhan, Eric Smith, Javier Rando, Yiming Zhang, Kate Plawiak, Zacharie~Delpierre Coudert, Kartikeya Upasani, and Mahesh Pasupuleti.
\newblock Llama guard 3 vision: Safeguarding human-ai image understanding conversations.
\newblock {\em arXiv preprint arXiv:2411.10414}, 2024.

\bibitem{llama_3_2}
llama Team.
\newblock Llama 3.2: Revolutionizing edge AI and vision with open, customizable models.
\newblock https://ai.meta.com/blog/llama-3-2-connect-2024-vision-edge-mobile-devices/, 2024.
\newblock Accessed: 2024-09-25.

\bibitem{wang2024comprehensive}
Zhichao Wang, Bin Bi, Shiva~Kumar Pentyala, Kiran Ramnath, Sougata Chaudhuri, Shubham Mehrotra, Xiang-Bo Mao, Sitaram Asur, et~al.
\newblock A comprehensive survey of LLM alignment techniques: RLHF, RLAIF, PPO, DPO and more.
\newblock {\em arXiv preprint arXiv:2407.16216}, 2024.

\bibitem{ouyang2022training}
Long Ouyang, Jeffrey Wu, Xu~Jiang, Diogo Almeida, Carroll Wainwright, Pamela Mishkin, Chong Zhang, Sandhini Agarwal, Katarina Slama, Alex Ray, et~al.
\newblock Training language models to follow instructions with human feedback.
\newblock In {\em Proc. NeurIPS}, 2022.

\bibitem{bai2022training}
Yuntao Bai, Andy Jones, Kamal Ndousse, Amanda Askell, Anna Chen, Nova DasSarma, Dawn Drain, Stanislav Fort, Deep Ganguli, Tom Henighan, et~al.
\newblock Training a helpful and harmless assistant with reinforcement learning from human feedback.
\newblock {\em arXiv preprint arXiv:2204.05862}, 2022.

\bibitem{lee2023rlaif}
Harrison Lee, Samrat Phatale, Hassan Mansoor, Thomas Mesnard, Johan Ferret, Kellie Lu, Colton Bishop, Ethan Hall, Victor Carbune, Abhinav Rastogi, et~al.
\newblock Rlaif vs. rlhf: Scaling reinforcement learning from human feedback with ai feedback.
\newblock In {\em Proc. ICML}, 2024.

\bibitem{ji2024beavertails}
Jiaming Ji, Mickel Liu, Josef Dai, Xuehai Pan, Chi Zhang, Ce~Bian, Boyuan Chen, Ruiyang Sun, Yizhou Wang, and Yaodong Yang.
\newblock Beavertails: Towards improved safety alignment of llm via a human-preference dataset.
\newblock In {\em Proc. NeurIPS}, 2024.

\bibitem{ji2024pku}
Jiaming Ji, Donghai Hong, Borong Zhang, Boyuan Chen, Josef Dai, Boren Zheng, Tianyi Qiu, Boxun Li, and Yaodong Yang.
\newblock PKU-SafeRLHF: Towards Multi-Level Safety Alignment for LLMs with Human Preference.
\newblock In {\em Proc. ACL}, 2025.

\bibitem{zong2024safety}
Yongshuo Zong, Ondrej Bohdal, Tingyang Yu, Yongxin Yang, and Timothy Hospedales.
\newblock Safety fine-tuning at (almost) no cost: A baseline for vision large language models.
\newblock In {\em Proc. ICML}, 2024.

\bibitem{hu2022lora}
Edward~J Hu, Yelong Shen, Phillip Wallis, Zeyuan Allen-Zhu, Yuanzhi Li, Shean Wang, Lu~Wang, Weizhu Chen, et~al.
\newblock Lora: Low-rank adaptation of large language models.
\newblock In {\em Proc. ICLR}, 2022.

\bibitem{chao2024jailbreakbench}
Patrick Chao, Edoardo Debenedetti, Alexander Robey, Maksym Andriushchenko, Francesco Croce, Vikash Sehwag, Edgar Dobriban, Nicolas Flammarion, George~J Pappas, Florian Tramer, et~al.
\newblock Jailbreakbench: An open robustness benchmark for jailbreaking large language models.
\newblock In {\em Proc. NeurIPS}, 2024.

\bibitem{zhang2025stair}
Yichi Zhang, Siyuan Zhang, Yao Huang, Zeyu Xia, Zhengwei Fang, Xiao Yang, Ranjie Duan, Dong Yan, Yinpeng Dong, and Jun Zhu.
\newblock STAIR: Improving Safety Alignment with Introspective Reasoning.
\newblock {\em arXiv preprint arXiv:2502.02384}, 2025.

\bibitem{li2024salad}
Lijun Li, Bowen Dong, Ruohui Wang, Xuhao Hu, Wangmeng Zuo, Dahua Lin, Yu~Qiao, and Jing Shao.
\newblock Salad-bench: A hierarchical and comprehensive safety benchmark for large language models.
\newblock In {\em Findings of Proc. ACL}, 2024.

\bibitem{mukherjee2023orca}
Subhabrata Mukherjee, Arindam Mitra, Ganesh Jawahar, Sahaj Agarwal, Hamid Palangi, and Ahmed Awadallah.
\newblock Orca: Progressive learning from complex explanation traces of gpt-4.
\newblock {\em arXiv preprint arXiv:2306.02707}, 2023.

\bibitem{lin2023toxicchat}
Zi~Lin, Zihan Wang, Yongqi Tong, Yangkun Wang, Yuxin Guo, Yujia Wang, and Jingbo Shang.
\newblock Toxicchat: Unveiling hidden challenges of toxicity detection in real-world user-ai conversation.
\newblock In {\em Findings of Proc. EMNLP}, 2023.

\bibitem{xie2024sorry}
Tinghao Xie, Xiangyu Qi, Yi~Zeng, Yangsibo Huang, Udari~Madhushani Sehwag, Kaixuan Huang, Luxi He, Boyi Wei, Dacheng Li, Ying Sheng, et~al.
\newblock Sorry-bench: Systematically evaluating large language model safety refusal behaviors.
\newblock In {\em Proc. ICLR}, 2025.

\bibitem{rottger2023xstest}
Paul R{\"o}ttger, Hannah~Rose Kirk, Bertie Vidgen, Giuseppe Attanasio, Federico Bianchi, and Dirk Hovy.
\newblock Xstest: A test suite for identifying exaggerated safety behaviours in large language models.
\newblock In {\em Proc. NAACL}, 2024.

\bibitem{luo2024jailbreakv}
Weidi Luo, Siyuan Ma, Xiaogeng Liu, Xiaoyu Guo, and Chaowei Xiao.
\newblock JailBreakV: A Benchmark for Assessing the Robustness of MultiModal Large Language Models against Jailbreak Attacks.
\newblock In {\em Proc. COLM}, 2024.

\bibitem{madaan2023self}
Aman Madaan, Niket Tandon, Prakhar Gupta, Skyler Hallinan, Luyu Gao, Sarah Wiegreffe, Uri Alon, Nouha Dziri, Shrimai Prabhumoye, Yiming Yang, et~al.
\newblock Self-refine: Iterative refinement with self-feedback.
\newblock In {\em Proc. NeurIPS}, 2023.

\bibitem{vidgen2023simplesafetytests}
Bertie Vidgen, Nino Scherrer, Hannah~Rose Kirk, Rebecca Qian, Anand Kannappan, Scott~A Hale, and Paul R{\"o}ttger.
\newblock Simplesafetytests: a test suite for identifying critical safety risks in large language models.
\newblock {\em arXiv preprint arXiv:2311.08370}, 2023.

\bibitem{mantas2023tdc}
Mazeika Mantas, Zou Andy, Mu~Norman, Phan Long, Wang Zifan, Yu~Chunru, Khoja Adam, Jiang Fengqing, O'Gara Aidan, Sakhaee Ellie, Xiang Zhen, Rajabi Arezoo, Hendrycks Dan, Poovendran Radha, Li~Bo, and Forsyth David.
\newblock Tdc 2023 (llm edition): The trojan detection challenge.
\newblock In {\em Proc. NeurIPS Competition Track}, 2023.

\bibitem{tedeschi2024alert}
Simone Tedeschi, Felix Friedrich, Patrick Schramowski, Kristian Kersting, Roberto Navigli, Huu Nguyen, and Bo~Li.
\newblock ALERT: A Comprehensive Benchmark for Assessing Large Language Models' Safety through Red Teaming.
\newblock {\em arXiv preprint arXiv:2404.08676}, 2024.

\bibitem{yu2024rlhf}
Tianyu Yu, Yuan Yao, Haoye Zhang, Taiwen He, Yifeng Han, Ganqu Cui, Jinyi Hu, Zhiyuan Liu, Hai-Tao Zheng, Maosong Sun, et~al.
\newblock Rlhf-v: Towards trustworthy mllms via behavior alignment from fine-grained correctional human feedback.
\newblock In {\em Proc. CVPR}, pages 13807--13816, 2024.

\bibitem{sun2023aligning}
Zhiqing Sun, Sheng Shen, Shengcao Cao, Haotian Liu, Chunyuan Li, Yikang Shen, Chuang Gan, Liang-Yan Gui, Yu-Xiong Wang, Yiming Yang, et~al.
\newblock Aligning large multimodal models with factually augmented rlhf.
\newblock In {\em Findings of Proc. ACL}, 2024.

\bibitem{li2024vlfeedback}
Lei Li, Zhihui Xie, Mukai Li, Shunian Chen, Peiyi Wang, Liang Chen, Yazheng Yang, Benyou Wang, Lingpeng Kong, and Qi~Liu.
\newblock VLFeedback: A Large-Scale AI Feedback Dataset for Large Vision-Language Models Alignment.
\newblock In {\em Proc. EMNLP}, 2024.

\bibitem{ji2025safe}
Jiaming Ji, Xinyu Chen, Rui Pan, Han Zhu, Conghui Zhang, Jiahao Li, Donghai Hong, Boyuan Chen, Jiayi Zhou, Kaile Wang, et~al.
\newblock Safe RLHF-V: Safe Reinforcement Learning from Human Feedback in Multimodal Large Language Models.
\newblock {\em arXiv preprint arXiv:2503.17682}, 2025.

\bibitem{zhang2025mm}
Yi-Fan Zhang, Tao Yu, Haochen Tian, Chaoyou Fu, Peiyan Li, Jianshu Zeng, Wulin Xie, Yang Shi, Huanyu Zhang, Junkang Wu, et~al.
\newblock Mm-rlhf: The next step forward in multimodal llm alignment.
\newblock In {\em Proc. ICML}, 2025.

\bibitem{goodfellow2014explaining}
Ian~J Goodfellow, Jonathon Shlens, and Christian Szegedy.
\newblock Explaining and harnessing adversarial examples.
\newblock {\em arXiv preprint arXiv:1412.6572}, 2014.

\bibitem{lin2023mitigating}
Yong Lin, Hangyu Lin, Wei Xiong, Shizhe Diao, Jianmeng Liu, Jipeng Zhang, Rui Pan, Haoxiang Wang, Wenbin Hu, Hanning Zhang, et~al.
\newblock Mitigating the alignment tax of rlhf.
\newblock In {\em Proc. EMNLP}, 2024.

\bibitem{bojar2014findings}
Ond{\v{r}}ej Bojar, Christian Buck, Christian Federmann, Barry Haddow, Philipp Koehn, Johannes Leveling, Christof Monz, Pavel Pecina, Matt Post, Herve Saint-Amand, et~al.
\newblock Findings of the 2014 workshop on statistical machine translation.
\newblock In {\em Proceedings of the ninth workshop on statistical machine translation}, pages 12--58, 2014.

\bibitem{rajpurkar2018know}
Pranav Rajpurkar, Robin Jia, and Percy Liang.
\newblock Know what you don't know: Unanswerable questions for SQuAD.
\newblock {\em arXiv preprint arXiv:1806.03822}, 2018.

\bibitem{clark2018think}
Peter Clark, Isaac Cowhey, Oren Etzioni, Tushar Khot, Ashish Sabharwal, Carissa Schoenick, and Oyvind Tafjord.
\newblock Think you have solved question answering? try arc, the ai2 reasoning challenge.
\newblock {\em arXiv preprint arXiv:1803.05457}, 2018.

\bibitem{li2022backdoor}
Yiming Li, Yong Jiang, Zhifeng Li, and Shu-Tao Xia.
\newblock Backdoor learning: A survey.
\newblock {\em IEEE Transactions on Neural Networks and Learning Systems}, 35(1):5--22, 2022.

\bibitem{guan2022few}
Jiyang Guan, Zhuozhuo Tu, Ran He, and Dacheng Tao.
\newblock Few-shot backdoor defense using shapley estimation.
\newblock In {\em Proc. CVPR}, pages 13358--13367, 2022.

\bibitem{guan2024backdoor}
Jiyang Guan, Jian Liang, and Ran He.
\newblock Backdoor defense via test-time detecting and repairing.
\newblock In {\em Proc. CVPR}, pages 24564--24573, 2024.

\bibitem{hubinger2024sleeper}
Evan Hubinger, Carson Denison, Jesse Mu, Mike Lambert, Meg Tong, Monte MacDiarmid, Tamera Lanham, Daniel~M Ziegler, Tim Maxwell, Newton Cheng, et~al.
\newblock Sleeper agents: Training deceptive llms that persist through safety training.
\newblock {\em arXiv preprint arXiv:2401.05566}, 2024.

\bibitem{xu2023instructions}
Jiashu Xu, Mingyu~Derek Ma, Fei Wang, Chaowei Xiao, and Muhao Chen.
\newblock Instructions as backdoors: Backdoor vulnerabilities of instruction tuning for large language models.
\newblock In {\em Proc. NAACL}, 2024.

\bibitem{shi2023badgpt}
Jiawen Shi, Yixin Liu, Pan Zhou, and Lichao Sun.
\newblock Badgpt: Exploring security vulnerabilities of chatgpt via backdoor attacks to instructgpt.
\newblock {\em arXiv preprint arXiv:2304.12298}, 2023.

\bibitem{li2024badedit}
Yanzhou Li, Tianlin Li, Kangjie Chen, Jian Zhang, Shangqing Liu, Wenhan Wang, Tianwei Zhang, and Yang Liu.
\newblock Badedit: Backdooring large language models by model editing.
\newblock In {\em Proc. ICLR}, 2024.

\bibitem{wang2023trojan}
Haoran Wang and Kai Shu.
\newblock Trojan activation attack: Red-teaming large language models using activation steering for safety-alignment.
\newblock {\em arXiv preprint arXiv:2311.09433}, 2023.

\bibitem{yan2023backdooring}
Jun Yan, Vikas Yadav, Shiyang Li, Lichang Chen, Zheng Tang, Hai Wang, Vijay Srinivasan, Xiang Ren, and Hongxia Jin.
\newblock Backdooring instruction-tuned large language models with virtual prompt injection.
\newblock In {\em Proc. NAACL}, 2024.

\bibitem{clop2024backdoored}
Cody Clop and Yannick Teglia.
\newblock Backdoored retrievers for prompt injection attacks on retrieval augmented generation of large language models.
\newblock {\em arXiv preprint arXiv:2410.14479}, 2024.

\bibitem{zhao2023prompt}
Shuai Zhao, Jinming Wen, Luu~Anh Tuan, Junbo Zhao, and Jie Fu.
\newblock Prompt as triggers for backdoor attack: Examining the vulnerability in language models.
\newblock In {\em Proc. EMNLP}, 2023.

\bibitem{liu2024formalizing}
Yupei Liu, Yuqi Jia, Runpeng Geng, Jinyuan Jia, and Neil~Zhenqiang Gong.
\newblock Formalizing and benchmarking prompt injection attacks and defenses.
\newblock In {\em Proc. USENIX Security}, pages 1831--1847, 2024.

\bibitem{braiek2025machine}
Houssem~Ben Braiek and Foutse Khomh.
\newblock Machine learning robustness: A primer.
\newblock In {\em Trustworthy AI in Medical Imaging}, pages 37--71. Elsevier, 2025.

\bibitem{wang2021measure}
Xuezhi Wang, Haohan Wang, and Diyi Yang.
\newblock Measure and improve robustness in NLP models: A survey.
\newblock In {\em Proc. NAACL}, 2022.

\bibitem{huang2023large}
Jie Huang, Xinyun Chen, Swaroop Mishra, Huaixiu~Steven Zheng, Adams~Wei Yu, Xinying Song, and Denny Zhou.
\newblock Large language models cannot self-correct reasoning yet.
\newblock In {\em Proc. ICLR}, 2024.

\bibitem{xi2023self}
Zhiheng Xi, Senjie Jin, Yuhao Zhou, Rui Zheng, Songyang Gao, Tao Gui, Qi~Zhang, and Xuanjing Huang.
\newblock Self-polish: Enhance reasoning in large language models via problem refinement.
\newblock In {\em Findings of Proc. EMNLP}, 2023.

\bibitem{yue2024mmmu}
Xiang Yue, Yuansheng Ni, Kai Zhang, Tianyu Zheng, Ruoqi Liu, Ge~Zhang, Samuel Stevens, Dongfu Jiang, Weiming Ren, Yuxuan Sun, et~al.
\newblock Mmmu: A massive multi-discipline multimodal understanding and reasoning benchmark for expert agi.
\newblock In {\em Proc. CVPR}, pages 9556--9567, 2024.

\bibitem{lu2023mathvista}
Pan Lu, Hritik Bansal, Tony Xia, Jiacheng Liu, Chunyuan Li, Hannaneh Hajishirzi, Hao Cheng, Kai-Wei Chang, Michel Galley, and Jianfeng Gao.
\newblock Mathvista: Evaluating mathematical reasoning of foundation models in visual contexts.
\newblock In {\em Proc. ICLR}, 2024.

\bibitem{wang2024charxiv}
Zirui Wang, Mengzhou Xia, Luxi He, Howard Chen, Yitao Liu, Richard Zhu, Kaiqu Liang, Xindi Wu, Haotian Liu, Sadhika Malladi, et~al.
\newblock Charxiv: Charting gaps in realistic chart understanding in multimodal llms.
\newblock In {\em Proc. NeurIPS}, 2024.

\bibitem{zhang2024understanding}
Qingjie Zhang, Han Qiu, Di~Wang, Haoting Qian, Yiming Li, Tianwei Zhang, and Minlie Huang.
\newblock Understanding the Dark Side of LLMs' Intrinsic Self-Correction.
\newblock {\em arXiv preprint arXiv:2412.14959}, 2024.

\bibitem{chen2024not}
Xingyu Chen, Jiahao Xu, Tian Liang, Zhiwei He, Jianhui Pang, Dian Yu, Linfeng Song, Qiuzhi Liu, Mengfei Zhou, Zhuosheng Zhang, et~al.
\newblock Do not think that much for 2+ 3=? on the overthinking of o1-like llms.
\newblock {\em arXiv preprint arXiv:2412.21187}, 2024.

\bibitem{xu2025chain}
Silei Xu, Wenhao Xie, Lingxiao Zhao, and Pengcheng He.
\newblock Chain of draft: Thinking faster by writing less.
\newblock {\em arXiv preprint arXiv:2502.18600}, 2025.

\bibitem{ma2025cot}
Xinyin Ma, Guangnian Wan, Runpeng Yu, Gongfan Fang, and Xinchao Wang.
\newblock CoT-Valve: Length-Compressible Chain-of-Thought Tuning.
\newblock {\em arXiv preprint arXiv:2502.09601}, 2025.

\bibitem{yang2025think}
Junjie Yang, Ke~Lin, and Xing Yu.
\newblock Think when you need: Self-adaptive chain-of-thought learning.
\newblock {\em arXiv preprint arXiv:2504.03234}, 2025.

\bibitem{wang2025adaptive}
Minzheng Wang, Yongbin Li, Haobo Wang, Xinghua Zhang, Nan Xu, Bingli Wu, Fei Huang, Haiyang Yu, and Wenji Mao.
\newblock Adaptive Thinking via Mode Policy Optimization for Social Language Agents.
\newblock {\em arXiv preprint arXiv:2505.02156}, 2025.

\bibitem{xia2025tokenskip}
Heming Xia, Yongqi Li, Chak~Tou Leong, Wenjie Wang, and Wenjie Li.
\newblock Tokenskip: Controllable chain-of-thought compression in llms.
\newblock {\em arXiv preprint arXiv:2502.12067}, 2025.

\bibitem{munkhbat2025self}
Tergel Munkhbat, Namgyu Ho, Seo~Hyun Kim, Yongjin Yang, Yujin Kim, and Se-Young Yun.
\newblock Self-training elicits concise reasoning in large language models.
\newblock {\em arXiv preprint arXiv:2502.20122}, 2025.

\bibitem{yu2024distilling}
Ping Yu, Jing Xu, Jason Weston, and Ilia Kulikov.
\newblock Distilling system 2 into system 1.
\newblock In {\em NeurIPS Workshop on Sys-2 Reasoning}, 2024.

\bibitem{huang2025efficient}
Chengsong Huang, Langlin Huang, Jixuan Leng, Jiacheng Liu, and Jiaxin Huang.
\newblock Efficient test-time scaling via self-calibration.
\newblock {\em arXiv preprint arXiv:2503.00031}, 2025.

\bibitem{wang2025sampling}
Yiming Wang, Pei Zhang, Siyuan Huang, Baosong Yang, Zhuosheng Zhang, Fei Huang, and Rui Wang.
\newblock Sampling-efficient test-time scaling: Self-estimating the best-of-n sampling in early decoding.
\newblock {\em arXiv preprint arXiv:2503.01422}, 2025.

\bibitem{yu2025think}
Zishun Yu, Tengyu Xu, Di~Jin, Karthik~Abinav Sankararaman, Yun He, Wenxuan Zhou, Zhouhao Zeng, Eryk Helenowski, Chen Zhu, Sinong Wang, et~al.
\newblock Think Smarter not Harder: Adaptive Reasoning with Inference Aware Optimization.
\newblock {\em arXiv preprint arXiv:2501.17974}, 2025.

\bibitem{huang2025mitigating}
Yao Huang, Huanran Chen, Shouwei Ruan, Yichi Zhang, Xingxing Wei, and Yinpeng Dong.
\newblock Mitigating Overthinking in Large Reasoning Models via Manifold Steering.
\newblock {\em arXiv preprint arXiv:2505.22411}, 2025.

\bibitem{cyberey2025steering}
Hannah Cyberey and David Evans.
\newblock Steering the CensorShip: Uncovering Representation Vectors for LLM ``Thought" Control.
\newblock {\em arXiv preprint arXiv:2504.17130}, 2025.

\bibitem{huang2024position}
Yue Huang, Lichao Sun, Haoran Wang, Siyuan Wu, Qihui Zhang, Yuan Li, Chujie Gao, Yixin Huang, Wenhan Lyu, Yixuan Zhang, et~al.
\newblock Position: Trustllm: Trustworthiness in large language models.
\newblock In {\em Proc. ICML}, 2024.

\bibitem{li2023survey}
Yingji Li, Mengnan Du, Rui Song, Xin Wang, and Ying Wang.
\newblock A survey on fairness in large language models.
\newblock {\em arXiv preprint arXiv:2308.10149}, 2023.

\bibitem{gallegos2024bias}
Isabel~O Gallegos, Ryan~A Rossi, Joe Barrow, Md~Mehrab Tanjim, Sungchul Kim, Franck Dernoncourt, Tong Yu, Ruiyi Zhang, and Nesreen~K Ahmed.
\newblock Bias and fairness in large language models: A survey.
\newblock {\em Computational Linguistics}, 50(3):1097--1179, 2024.

\bibitem{cantini2025benchmarking}
Riccardo Cantini, Alessio Orsino, Massimo Ruggiero, and Domenico Talia.
\newblock Benchmarking adversarial robustness to bias elicitation in large language models: Scalable automated assessment with llm-as-a-judge.
\newblock {\em arXiv preprint arXiv:2504.07887}, 2025.

\bibitem{wang2024towards}
Yanbo Wang, Jian Liang, and Ran He.
\newblock Towards eliminating hard label constraints in gradient inversion attacks.
\newblock In {\em Proc. ICLR}, 2024.

\bibitem{shokri2017membership}
Reza Shokri, Marco Stronati, Congzheng Song, and Vitaly Shmatikov.
\newblock Membership inference attacks against machine learning models.
\newblock In {\em Proc. S\&P}, pages 3--18, 2017.

\bibitem{zhou2024model}
Zhanke Zhou, Jianing Zhu, Fengfei Yu, Xuan Li, Xiong Peng, Tongliang Liu, and Bo~Han.
\newblock Model inversion attacks: A survey of approaches and countermeasures.
\newblock {\em arXiv preprint arXiv:2411.10023}, 2024.

\bibitem{jiang2025shadow}
Le~Jiang, Liyan Ma, and Guang Yang.
\newblock Shadow defense against gradient inversion attack in federated learning.
\newblock {\em Medical Image Analysis}, page 103673, 2025.

\bibitem{guan2022you}
Jiyang Guan, Jian Liang, and Ran He.
\newblock Are you stealing my model? sample correlation for fingerprinting deep neural networks.
\newblock In {\em Proc. NeurIPS}, 2022.

\bibitem{guan2025sample}
Jiyang Guan, Jian Liang, Yanbo Wang, and Ran He.
\newblock Sample Correlation for Fingerprinting Deep Face Recognition.
\newblock {\em International Journal of Computer Vision}, 133(4):1912--1926, 2025.

\bibitem{jiang2024rag}
Changyue Jiang, Xudong Pan, Geng Hong, Chenfu Bao, and Min Yang.
\newblock Rag-thief: Scalable extraction of private data from retrieval-augmented generation applications with agent-based attacks.
\newblock {\em arXiv preprint arXiv:2411.14110}, 2024.

\bibitem{wang2025silent}
Yuhao Wang, Wenjie Qu, Yanze Jiang, Zichen Liu, Yue Liu, Shengfang Zhai, Yinpeng Dong, and Jiaheng Zhang.
\newblock Silent Leaks: Implicit Knowledge Extraction Attack on RAG Systems through Benign Queries.
\newblock {\em arXiv preprint arXiv:2505.15420}, 2025.

\bibitem{carlini2024stealing}
Nicholas Carlini, Daniel Paleka, Krishnamurthy~Dj Dvijotham, Thomas Steinke, Jonathan Hayase, A~Feder Cooper, Katherine Lee, Matthew Jagielski, Milad Nasr, Arthur Conmy, et~al.
\newblock Stealing part of a production language model.
\newblock In {\em Proc. ICML}, 2024.

\bibitem{yao2024large}
Yuanshun Yao, Xiaojun Xu, and Yang Liu.
\newblock Large language model unlearning.
\newblock In {\em Proc. NeurIPS}, 2024.

\bibitem{maini2024tofu}
Pratyush Maini, Zhili Feng, Avi Schwarzschild, Zachary~C Lipton, and J~Zico Kolter.
\newblock Tofu: A task of fictitious unlearning for llms.
\newblock In {\em Proc. COLM}, 2024.

\bibitem{lynch2024eight}
Aengus Lynch, Phillip Guo, Aidan Ewart, Stephen Casper, and Dylan Hadfield-Menell.
\newblock Eight methods to evaluate robust unlearning in llms.
\newblock {\em arXiv preprint arXiv:2402.16835}, 2024.

\bibitem{hu2024jogging}
Shengyuan Hu, Yiwei Fu, Steven Wu, and Virginia Smith.
\newblock Jogging the memory of unlearned llms through targeted relearning attacks.
\newblock In {\em NeurIPS Workshop on Safe Generative AI}, 2024.

\bibitem{peng2022fingerprinting}
Zirui Peng, Shaofeng Li, Guoxing Chen, Cheng Zhang, Haojin Zhu, and Minhui Xue.
\newblock Fingerprinting deep neural networks globally via universal adversarial perturbations.
\newblock In {\em Proc. CVPR}, pages 13430--13439, 2022.

\bibitem{wang2021fingerprinting}
Si~Wang and Chip-Hong Chang.
\newblock Fingerprinting deep neural networks-a deepfool approach.
\newblock In {\em Proc. ISCAS}, pages 1--5, 2021.

\bibitem{guan2024sample}
Jiyang Guan, Jian Liang, Yanbo Wang, and Ran He.
\newblock Sample Correlation for Fingerprinting Deep Face Recognition.
\newblock {\em International Journal of Computer Vision}, pages 1--15, 2024.

\bibitem{fernandez2023stable}
Pierre Fernandez, Guillaume Couairon, Herv{\'e} J{\'e}gou, Matthijs Douze, and Teddy Furon.
\newblock The stable signature: Rooting watermarks in latent diffusion models.
\newblock In {\em Proc. ICCV}, pages 22466--22477, 2023.

\bibitem{li2021modeldiff}
Yuanchun Li, Ziqi Zhang, Bingyan Liu, Ziyue Yang, and Yunxin Liu.
\newblock ModelDiff: Testing-based DNN similarity comparison for model reuse detection.
\newblock In {\em Proc. ISSTA}, pages 139--151, 2021.

\bibitem{fu2022robust}
Shaopeng Fu, Fengxiang He, Yang Liu, Li~Shen, and Dacheng Tao.
\newblock Robust unlearnable examples: Protecting data against adversarial learning.
\newblock In {\em Proc. ICLR}, 2022.

\bibitem{sandoval2022autoregressive}
Pedro Sandoval-Segura, Vasu Singla, Jonas Geiping, Micah Goldblum, Tom Goldstein, and David Jacobs.
\newblock Autoregressive perturbations for data poisoning.
\newblock In {\em Proc. NeurIPS}, 2022.

\bibitem{huang2021unlearnable}
Hanxun Huang, Xingjun Ma, Sarah~Monazam Erfani, James Bailey, and Yisen Wang.
\newblock Unlearnable examples: Making personal data unexploitable.
\newblock In {\em Proc. ICLR}, 2021.

\bibitem{kirchenbauer2023watermark}
John Kirchenbauer, Jonas Geiping, Yuxin Wen, Jonathan Katz, Ian Miers, and Tom Goldstein.
\newblock A watermark for large language models.
\newblock In {\em Proc. ICML}, pages 17061--17084, 2023.

\bibitem{staab2023beyond}
Robin Staab, Mark Vero, Mislav Balunovi{\'c}, and Martin Vechev.
\newblock Beyond memorization: Violating privacy via inference with large language models.
\newblock In {\em Proc. ICLR}, 2024.

\bibitem{tomekcce2024private}
Batuhan T{\"o}mek{\c{c}}e, Mark Vero, Robin Staab, and Martin Vechev.
\newblock Private Attribute Inference from Images with Vision-Language Models.
\newblock In {\em Proc. NeurIPS}, 2024.

\bibitem{chu2025sft}
Tianzhe Chu, Yuexiang Zhai, Jihan Yang, Shengbang Tong, Saining Xie, Dale Schuurmans, Quoc~V Le, Sergey Levine, and Yi~Ma.
\newblock Sft memorizes, rl generalizes: A comparative study of foundation model post-training.
\newblock {\em arXiv preprint arXiv:2501.17161}, 2025.

\bibitem{chen2025sft}
Hardy Chen, Haoqin Tu, Fali Wang, Hui Liu, Xianfeng Tang, Xinya Du, Yuyin Zhou, and Cihang Xie.
\newblock Sft or rl? an early investigation into training r1-like reasoning large vision-language models.
\newblock {\em arXiv preprint arXiv:2504.11468}, 2025.

\end{thebibliography}
}

\end{document}